\documentclass{article}

\usepackage[eandd, final]{neurips_2026}
\usepackage[utf8]{inputenc} 
\usepackage[T1]{fontenc}    
\usepackage{hyperref}       
\usepackage{url}            
\usepackage{booktabs}       
\usepackage{amsfonts}       
\usepackage{nicefrac}       
\usepackage{microtype}      
\usepackage{xcolor}         
\usepackage{subcaption}
\usepackage{multirow}
\usepackage{tabularx}
\usepackage{pifont}
\usepackage{silence}
\usepackage[utf8]{inputenc}
\usepackage[T1]{fontenc}
\usepackage{tabularx}
\usepackage{longtable}

\usepackage{lipsum}
\usepackage{url}
\usepackage{placeins}
\usepackage{booktabs}
\usepackage{soul}
\usepackage{amsfonts}
\usepackage{nicefrac}
\usepackage{microtype}
\usepackage{wrapfig}
\usepackage{caption}
\usepackage{amssymb}
\usepackage{pifont}

\usepackage{subcaption}
\usepackage{float}
\usepackage{rotating}
\usepackage{etoolbox}
\usepackage{xparse}
\usepackage{grffile}

\usepackage{amsmath}
\usepackage{xspace}
\usepackage{euscript}
\usepackage{enumitem}
\usepackage{graphicx}
\usepackage{mathtools}
\usepackage{tikz}
\usepackage{tablefootnote}
\usepackage[flushleft]{threeparttable}
\usepackage{multirow}
\usepackage[title]{appendix}
\usepackage{arydshln}
\usepackage{array, booktabs}
\usepackage{graphicx}
\usepackage{glossaries}
\usepackage{amsmath,amsthm}
\usepackage{dsfont}
\usepackage{balance}
\usepackage{multicol}
\usepackage{nameref}
\usepackage{pdfpages}
\usepackage{braket}
\usepackage[normalem]{ulem}
\usepackage{bbding}
\usepackage{xfrac}
\usepackage[usestackEOL]{stackengine}
\usepackage{indentfirst}
\usepackage{csquotes}
\usepackage{pgf}
\usepackage{varwidth}
\usepackage{verbatimbox}
\usepackage{verbatim}
\usepackage{bm}
\usepackage{anyfontsize}
\usepackage{nccmath}
\usepackage{makecell}
\usepackage{colortbl}

\usepackage[table,dvipsnames]{xcolor}

\usepackage{microtype}
\usepackage{booktabs} 

\usepackage{hyperref}
\usepackage{amsmath}
\usepackage{amssymb}
\usepackage{mathtools}
\usepackage{amsthm}
\usepackage[capitalise, noabbrev, nameinlink]{cleveref}   
\usepackage{amsfonts,bm}

\usepackage{fancyhdr}
\usepackage{titletoc}
\usepackage{empheq}
\usepackage{soul}

\usepackage[framemethod=TikZ]{mdframed}
\usepackage[textsize=tiny]{todonotes}

\usepackage{thm-restate}

\usepackage[most]{tcolorbox}

\makeatletter
\def\thmt@refnamewithcomma #1#2#3,#4,#5\@nil{%
	\@xa\def\csname\thmt@envname #1utorefname\endcsname{#3}%
	\ifcsname #2refname\endcsname
	\csname #2refname\expandafter\endcsname\expandafter{\thmt@envname}{#3}{#4}%
	\fi}
\makeatother
\Crefname{conjecture}{Conjecture}{Conjectures}
\Crefname{definition}{Definition}{Definitions}
\Crefname{observation}{Observation}{Observations}
\Crefname{assumption}{Assumption}{Assumptions}
\Crefname{axiom}{Axiom}{Axioms}
\Crefname{case}{Case}{Cases}
\Crefname{claim}{Claim}{Claims}
\Crefname{conclusion}{Conclusion}{Conclusions}
\Crefname{condition}{Condition}{Conditions}
\Crefname{criterion}{Criterion}{Criteria}
\Crefname{exercise}{Exercise}{Exercises}
\Crefname{example}{Example}{Examples}
\Crefname{notation}{Notation}{Notations}
\Crefname{problem}{Problem}{Problems}
\Crefname{property}{Property}{Properties}
\Crefname{remark}{Remark}{Remarks}
\Crefname{solution}{Solution}{Solutions}
\Crefname{summary}{Summary}{Summaries}
\Crefname{motivation}{Motivation}{Motivations}
\Crefname{query}{Query}{Queries}
\usepackage{cancel}

\renewcommand{\paragraph}[1]{{\noindent \textbf{#1.}}}

\newcommand*\dbar[1]{\overline{\overline{\lower0.2ex\hbox{$#1$}}}}

\DeclareFontFamily{U}{BOONDOX-calo}{\skewchar\font=45 }
\DeclareFontShape{U}{BOONDOX-calo}{m}{n}{
  <-> s*[1.05] BOONDOX-r-calo}{}
\DeclareFontShape{U}{BOONDOX-calo}{b}{n}{
  <-> s*[1.05] BOONDOX-b-calo}{}
\DeclareMathAlphabet{\mathcalb}{U}{BOONDOX-calo}{m}{n}
\SetMathAlphabet{\mathcalb}{bold}{U}{BOONDOX-calo}{b}{n}
\DeclareMathAlphabet{\mathbcalb}{U}{BOONDOX-calo}{b}{n}

\def\eqref#1{equation~\ref{#1}}

\def\1{\bm{1}}

\DeclareMathAlphabet{\mathsfit}{\encodingdefault}{\sfdefault}{m}{sl}
\SetMathAlphabet{\mathsfit}{bold}{\encodingdefault}{\sfdefault}{bx}{n}

\mdfdefinestyle{MyFrame}{%
    linecolor=black,
    outerlinewidth=0.5pt,
    roundcorner=1pt,
    innertopmargin=2pt, 
    innerbottommargin=2pt, 
    innerrightmargin=7pt,
    innerleftmargin=7pt,
    backgroundcolor=black!0!white}

\mdfdefinestyle{MyFrame2}{%
    linecolor=white,
    outerlinewidth=1pt,
    roundcorner=2pt,
    innertopmargin=10pt,
    innerbottommargin=0.5pt,
    innerrightmargin=10pt,
    innerleftmargin=10pt,
    backgroundcolor=black!3!white}

\mdfdefinestyle{MyFrameEq}{%
    linecolor=white,
    outerlinewidth=0pt,
    roundcorner=0pt,
    innertopmargin=0pt,
    innerbottommargin=0pt,
    innerrightmargin=7pt,
    innerleftmargin=7pt,
    backgroundcolor=black!3!white}

\mdfdefinestyle{FrameAdded}{%
    linecolor=black,
    outerlinewidth=0pt,
    roundcorner=0pt,
    innertopmargin=0pt,
    innerbottommargin=0pt,
    innerrightmargin=0pt,
    innerleftmargin=0pt,
    backgroundcolor=PastelGreen}

\theoremstyle{plain}

\theoremstyle{definition}

\theoremstyle{remark}

\newcommand{\name}{\textsc{SelectBench}\xspace}
\newcommand{\taskA}{\name-Case\xspace}
\newcommand{\taskB}{\name-Safety\xspace}
\newcommand{\taskBsmall}{\name-Safety-Priority\xspace}
\newcommand{\taskBbig}{\name-Safety-All\xspace}
\newcommand{\taskC}{\name-XL\xspace}

\newcommand{\mycaption}[2]{\caption{\textbf{#1} #2}}

\fancypagestyle{toc}{%
\fancyhf{} 
\fancyfoot[C]{}

\fancyfoot[R]{\hyperlink{appendixtoc}{\footnotesize{\texttt{(Back to Table of Contents)}}}}
}

\usepackage{tikz}
\newcommand{\colordot}[1]{\tikz[baseline=-0.6ex]\draw[fill=#1, draw=none] (0,0) circle (0.6ex);}
\newcommand{\colorsquare}[1]{%
  \tikz[baseline=-0.6ex]\draw[fill=#1, draw=none]
  (-0.6ex,-0.6ex) rectangle (0.6ex,0.6ex);}
\newcommand{\colortriangle}[1]{\tikz[baseline=-0.6ex]\draw[fill=#1, draw=none] (0,0.75ex) -- (-0.65ex,-0.4ex) -- (0.65ex,-0.4ex) -- cycle;}
\newcommand{\yes}{\textcolor{green!60!black}{\ding{51}}} 
\newcommand{\no}{\textcolor{red!70!black}{\ding{55}}}    

\newtcolorbox{customabstract}{%
  enhanced,
  breakable,
  colback    = white,
  colframe   = Blue1,
  boxrule    = 1.5pt,
  arc        = 3pt,
  left       = 12pt,
  right      = 12pt,
  top        = 10pt,
  bottom     = 10pt,
  before upper = {%
    \centerline
    {\large\bf Abstract}%
    \vspace{10pt}%
  },
}

\definecolor{Blue1}{HTML}{C7DDDE} 
\definecolor{Blue2}{HTML}{7DBDC2} 
\definecolor{Blue3}{HTML}{007A87} 
\definecolor{Blue4}{HTML}{004751} 

\definecolor{Yellow1}{HTML}{FFF8D4} 
\definecolor{Yellow2}{HTML}{F7C886} 
\definecolor{Yellow3}{HTML}{F8B65A} 
\definecolor{Yellow4}{HTML}{DE9B35} 

\definecolor{Grey1}{HTML}{FBFAF4} 
\definecolor{Grey2}{HTML}{EDEBE2} 
\definecolor{Grey3}{HTML}{CFCDC3} 
\definecolor{Grey4}{HTML}{9D9D9C}
\definecolor{Grey5}{HTML}{4D4D4D}

\definecolor{Pink1}{HTML}{F3CDF0} 
\definecolor{Pink2}{HTML}{E993D0} 
\definecolor{Pink3}{HTML}{CF50A0} 
\definecolor{Pink4}{HTML}{94133F} 

\definecolor{Purple1}{HTML}{D6D4F5} 
\definecolor{Purple2}{HTML}{B4B4ED} 
\definecolor{Purple3}{HTML}{504FB3} 
\definecolor{Purple4}{HTML}{2B2B7F} 

\definecolor{citeblue}{HTML}{007A87} 

\makeatletter
\if@anonymous
    
\else
    \renewenvironment{abstract}%
        {%
          \vspace{-20pt}
          \begin{customabstract}%
        }
        {
          \par%
          \end{customabstract}%
          \vskip 1ex%
    }
\fi
\makeatother

\definecolor{customwhite}{HTML}{FCFBF7}
\definecolor{customturq}{HTML}{1D9D79}
\definecolor{customorange}{HTML}{D96002} 
\definecolor{custombeige}{HTML}{d6c9b1} 
\definecolor{customblue}{HTML}{4D9DE0} 
\definecolor{custompurple}{HTML}{AEADF0}

\definecolor{plotgreen}{RGB}{72, 164, 135}   
\definecolor{plotorange}{RGB}{222, 111, 68}  
\definecolor{plotpurple}{RGB}{111, 130, 173}

\definecolor{myblue}{rgb}{.8, .8, 1}

\definecolor{pastelblue}{RGB}{76,113,175}
\definecolor{pastelgreen}{RGB}{144,238,144}
\definecolor{pastelred}{RGB}{196,78,82}
\definecolor{pastelgrey}{RGB}{230,230,230}
\definecolor{pastelbeige}{RGB}{243,236,221}
\definecolor{pastelpurple}{RGB}{154,139,192}
\definecolor{antiquefuchsia}{rgb}{0.57, 0.36, 0.51}
\definecolor{amethyst}{rgb}{0.6, 0.4, 0.8}

\definecolor{verylightgreen}{HTML}{F6FFF9}
\definecolor{verylightred}{HTML}{FFF4F3}
\definecolor{verylightgray}{HTML}{F4F6F6}
\definecolor{babyblueeyes}{rgb}{0.63, 0.79, 0.95}
\definecolor{lightpink}{rgb}{1.00, 0.714, 0.757}

\definecolor{highlightturq}{HTML}{92CEBD}
\definecolor{highlightpurple}{HTML}{AEADF0}
\definecolor{highlightorange}{HTML}{ECAF80}
\definecolor{highlightgray}{HTML}{E6E6E6}
\definecolor{highlightblue}{HTML}{86CFDA}

\sethlcolor{highlightblue!50!white}

\newtcbox{\hlblue}{
  on line,
  colback=Blue1,
  colframe=Blue1,
  arc=2pt,        
  boxrule=0pt,    
  left=2pt,
  right=2pt,
  top=1pt,
  bottom=1pt
}

\newtcbox{\hlpink}{
  on line,
  colback=Pink1,
  colframe=Pink1,
  arc=2pt,        
  boxrule=0pt,    
  left=2pt,
  right=2pt,
  top=1pt,
  bottom=1pt
}

\newtcbox{\hlyellow}{
  on line,
  colback=Yellow2,
  colframe=Yellow2,
  arc=2pt,        
  boxrule=0pt,    
  left=2pt,
  right=2pt,
  top=1pt,
  bottom=1pt
}
\newtcbox{\hlpurple}{
  on line,
  colback=Purple1,
  colframe=Purple1,
  arc=2pt,        
  boxrule=0pt,    
  left=2pt,
  right=2pt,
  top=1pt,
  bottom=1pt
}
\newtcbox{\hlgrey}{
  on line,
  colback=Grey2,
  colframe=Grey2,
  arc=2pt,        
  boxrule=0pt,    
  left=2pt,
  right=2pt,
  top=1pt,
  bottom=1pt
}

\newtcbox{\hlcustom}[1]{
  on line,
  colback=#1,
  colframe=#1,
  arc=2pt,
  boxrule=0pt,
  left=2pt,
  right=2pt,
  top=1pt,
  bottom=1pt
}

\hypersetup{
	colorlinks=true,       
	linkcolor=Blue3,        
	citecolor=Blue3,        
	filecolor=Pink3,     
	urlcolor=Blue2         
}

\let\originalleft\left
\let\originalright\right
\renewcommand{\left}{\mathopen{}\mathclose\bgroup\originalleft}
\renewcommand{\right}{\aftergroup\egroup\originalright}

\newcommand{\cut}[1]{}

\newcommand{\var}{{\rm I\kern-.3em D}}

\newtheorem*{theorem*}{Theorem}
\newtheorem*{lemma*}{Lemma}
\newtheorem*{proposition*}{Proposition}
\newtheorem*{corollary*}{Corollary}

\newtheorem*{example*}{Example}
\DeclareMathSymbol{\shortminus}{\mathbin}{AMSa}{"39}

\crefname{equation}{Eq.}{Eqs.}
\crefname{section}{Sec.}{Secs.}
\crefname{corollary}{Cor.}{Cors.}
\crefname{proposition}{Prop.}{Props.}
\crefname{appendix}{App.}{Apps.}
\crefname{theorem}{Thm.}{Thms.}
\crefname{figure}{Fig.}{Figs.}
\crefname{tabular}{Tab.}{Tabs.}

\newcommand{\ifbm}[1]
{#1}

\tcbuselibrary{theorems}
\newtcbtheorem[number within=section]{myprop}{Proposition}%
{colback=blue!5,colframe=blue!35!black,fonttitle=\bfseries}{th}

\makeatletter
\let\save@mathaccent\mathaccent
\newcommand*\if@single[3]{%
  \setbox0\hbox{${\mathaccent"0362{#1}}^H$}%
  \setbox2\hbox{${\mathaccent"0362{\kern0pt#1}}^H$}%
  \ifdim\ht0=\ht2 #3\else #2\fi
  }
\newcommand*\rel@kern[1]{\kern#1\dimexpr\macc@kerna}
\newcommand*\widebar[1]{\@ifnextchar^{{\wide@bar{#1}{0}}}{\wide@bar{#1}{1}}}
\newcommand*\wide@bar[2]{\if@single{#1}{\wide@bar@{#1}{#2}{1}}{\wide@bar@{#1}{#2}{2}}}
\newcommand*\wide@bar@[3]{%
  \begingroup
  \def\mathaccent##1##2{%
    \let\mathaccent\save@mathaccent
    \if#32 \let\macc@nucleus\first@char \fi
    \setbox\z@\hbox{$\macc@style{\macc@nucleus}_{}$}%
    \setbox\tw@\hbox{$\macc@style{\macc@nucleus}{}_{}$}%
    \dimen@\wd\tw@
    \advance\dimen@-\wd\z@
    \divide\dimen@ 3
    \@tempdima\wd\tw@
    \advance\@tempdima-\scriptspace
    \divide\@tempdima 10
    \advance\dimen@-\@tempdima
    \ifdim\dimen@>\z@ \dimen@0pt\fi
    \rel@kern{0.6}\kern-\dimen@
    \if#31
      \overline{\rel@kern{-0.6}\kern\dimen@\macc@nucleus\rel@kern{0.4}\kern\dimen@}%
      \advance\dimen@0.4\dimexpr\macc@kerna
      \let\final@kern#2%
      \ifdim\dimen@<\z@ \let\final@kern1\fi
      \if\final@kern1 \kern-\dimen@\fi
    \else
      \overline{\rel@kern{-0.6}\kern\dimen@#1}%
    \fi
  }%
  \macc@depth\@ne
  \let\math@bgroup\@empty \let\math@egroup\macc@set@skewchar
  \mathsurround\z@ \frozen@everymath{\mathgroup\macc@group\relax}%
  \macc@set@skewchar\relax
  \let\mathaccentV\macc@nested@a
  \if#31
    \macc@nested@a\relax111{#1}%
  \else
    \def\gobble@till@marker##1\endmarker{}%
    \futurelet\first@char\gobble@till@marker#1\endmarker
    \ifcat\noexpand\first@char A\else
      \def\first@char{}%
    \fi
    \macc@nested@a\relax111{\first@char}%
  \fi
  \endgroup
}
\makeatother

\usepackage{tikzpeople}
\usetikzlibrary{shapes,decorations,arrows,calc,arrows.meta,fit,positioning}

\tikzset{
    -Latex,auto,node distance =1 cm and 1 cm,semithick,
    state/.style ={ellipse, draw, minimum width = 0.7 cm},
    point/.style = {circle, draw, inner sep=0.04cm,fill,node contents={}},
    bidirected/.style={Latex-Latex,dashed},
    el/.style = {inner sep=2pt, align=left, sloped}
}

\mdfdefinestyle{MyFrame}{%
    linecolor=black,
    outerlinewidth=0.5pt,
    roundcorner=1pt,
    innertopmargin=2pt, 
    innerbottommargin=2pt, 
    innerrightmargin=7pt,
    innerleftmargin=7pt,
    backgroundcolor=black!0!white}

\mdfdefinestyle{MyFrame2}{%
    linecolor=white,
    outerlinewidth=1pt,
    roundcorner=2pt,
    innertopmargin=10pt,
    innerbottommargin=0.5pt,
    innerrightmargin=10pt,
    innerleftmargin=10pt,
    backgroundcolor=black!3!white}

\mdfdefinestyle{MyFrameEq}{%
    linecolor=white,
    outerlinewidth=0pt,
    roundcorner=0pt,
    innertopmargin=0pt,
    innerbottommargin=0pt,
    innerrightmargin=7pt,
    innerleftmargin=7pt,
    backgroundcolor=black!3!white}

\mdfdefinestyle{FrameAdded}{%
    linecolor=black,
    outerlinewidth=0pt,
    roundcorner=0pt,
    innertopmargin=0pt,
    innerbottommargin=0pt,
    innerrightmargin=0pt,
    innerleftmargin=0pt,
    backgroundcolor=PastelGreen}

\tcbset{mybox/.style={colback=olive!3, width =\textwidth, boxrule=0.0pt, top=5pt,bottom=5pt,left=2pt, right=2pt},
  before upper=\setlength{\parindent}
  {0em}\everypar{{\setbox0\lastbox}\everypar{}}
}
\newtcolorbox{mybox}[1][]{mybox,#1}

\tcbset{halfmybox/.style={colback=olive!3, width =0.49\textwidth, boxrule=0.0pt, top=5pt,bottom=5pt,left=2pt, right=2pt},
  before upper=\setlength{\parindent}
  {0em}\everypar{{\setbox0\lastbox}\everypar{}}
}
\newtcolorbox{halfmybox}[1][]{halfmybox,#1}

\tcbuselibrary{skins, breakable}

\newtcolorbox{custombox}[3][]{%
  enhanced,
  colback=#2!30!white,
  colframe=#3,
  boxrule=0pt,
  arc=5pt,
  left=-20pt,
  right=5pt,
  top=6pt,
  bottom=6pt,
  grow to left by=-5mm,
  title={#1},
  fonttitle=\ttfamily\bfseries\color{#3},
  detach title,
  overlay unbroken and first={
    \node[
      rotate=90,
      minimum width=1cm,
      anchor=south,
      font=\bfseries\ttfamily\color{#3}
    ] at (frame.west) {\tcbtitle};
  },
  toprule=0pt,
  rightrule=0pt,
  bottomrule=0pt,
  leftrule=1mm,
}

\newcommand{\zerodisplayskips}{%
\setlength{\abovedisplayskip}{2mm}%
\setlength{\belowdisplayskip}{2mm}%
\setlength{\abovedisplayshortskip}{1mm}%
\setlength{\belowdisplayshortskip}{1mm}}
\appto{\normalsize}{\zerodisplayskips}
\appto{\small}{\zerodisplayskips}
\appto{\footnotesize}{\zerodisplayskips}

\title{On the Selectivity of Generative Models\\in Structure-Based Drug Design}

\author{%
  {\textbf{Ella Miray Rajaonson\textsuperscript{* 1 2},\;}
  \textbf{Jungyoon Lee\textsuperscript{* 3 4}}}\\[3pt]
  {\textbf{William Chau\textsuperscript{ 1},}\;
  \textbf{Alán Aspuru-Guzik\textsuperscript{ 1 2 5 6 7},}\;
  \textbf{Benjamin Sanchez-Lengeling\textsuperscript{ 1 2}}}\\[3pt]
  {\textbf{Dominique Beaini\textsuperscript{ 3 4 8},}\;
  \textbf{Kirill Neklyudov\textsuperscript{ 3 4 9},}\;
  \textbf{Marta Skreta\textsuperscript{ 3 4}}}\\[3pt]
  {\small%
    \textsuperscript{1}University of Toronto\enspace
    \textsuperscript{2}Vector Institute\enspace
    \textsuperscript{3}Université de Montréal\enspace
    \textsuperscript{4}Mila -- Québec AI Institute}\\
  {\small%
    \textsuperscript{5}NVIDIA\enspace
    \textsuperscript{6}Acceleration Consortium\enspace
    \textsuperscript{7}CIFAR\enspace
    \textsuperscript{8}Valence Labs\enspace
    \textsuperscript{9}Institut Courtois}\\[6pt]
  {\small Correspondance to: \texttt{ella.rajaonson@mail.utoronto.ca, jungyoon.lee@mila.quebec},}\\[6pt]
{\small\texttt{marta.skreta@mila.quebec}}\\[6pt]
  {\small\textsuperscript{*}Equal contribution}
}

\begin{document}

\maketitle

\begin{abstract} 
Designing small-molecule drugs that selectively bind to a target protein while avoiding unintended interactions with other proteins is essential for reducing side effects that contribute substantially to clinical attrition. Although structure-based drug design (SBDD) has advanced rapidly with deep learning, most generative models still optimize binding to a single target and disregard selectivity due to limited training data and unknown off-target interactions. Motivated by this gap, we introduce \name, a comprehensive benchmark assessing the selectivity of generative SBDD models. \name provides three curated evaluation tasks reflecting different real-world drug discovery scenarios: literature-based case studies, safety screening panels used in pharmaceutical development, and large-scale structure-based evaluation sets. We further introduce a suite of metrics to quantify selectivity and conduct a systematic evaluation of nine representative SBDD models, including methods explicitly trained for selectivity and widely-used inference-time guidance techniques. Our results show that even selectivity-trained models achieve only modest off-target discrimination and inference-time guidance provides limited gains, underscoring the need for method improvement. \name provides both a characterization of where current generative SBDD stands on selectivity and an open, extensible platform to accelerate progress.\footnote{The benchmark is available at \url{https://github.com/lovetaro/selectbench}}
\end{abstract}

\section{Introduction}
\label{sec:intro}

Decreasing the high attrition rate in the drug discovery and development process is a primary goal of the pharmaceutical industry, as the current success rate of new drug candidates is less than 10\% \cite{sun202290}. One of the main challenges is striking an appropriate balance between drug efficacy and adverse effects as early as possible in order to reduce safety-related attrition~\citep{stevens2006future}. Indeed, analyses of clinical trial data from 2010 to 2017 show that lack of clinical efficacy and unmanageable toxicity are responsible for 30-50\% of current clinical failures \cite{sun202290, fraser2026mapping}. The ability of small molecules to interact with multiple proteins is commonly referred to as \textit{polypharmacology}. Polypharmacology of drug-like small molecules beyond their intended targets is a fact in drug discovery, as it has been estimated that over half of approved small-molecule drugs interact with more than five protein targets \cite{jalencas2013origins}. Polypharmacology can be beneficial for drug efficacy in some cases due to the benefits of targeting multiple signalling pathways \cite{dai2016proteome}. However, unexpected interactions can equally lead to undesired and often severe side effects, and has resulted in the withdrawal of several drugs from the market in the past~\citep{gintant2016evolution}. Directed polypharmacology has recently been promoted as a new paradigm for drug discovery, particularly in oncology \cite{shapiro2019promiscuous}. Controlling the polypharmacology of small-molecules implies appropriate tuning of their binding \textit{selectivity}: we want to rationally select which target(s) they bind and do not bind to.

The field of SBDD, which aims to find small-molecule ligands that bind to specific three-dimensional (3D) sites in proteins with high affinity and selectivity, has seen rapid progress driven by recent deep learning (DL) breakthroughs in protein structure prediction \cite{abramson2024accurate, jumper2021highly}. 

In parallel, a growing body of generative SBDD methods has emerged, leveraging structural information of protein–ligand complexes to propose novel drug candidates \cite{guan2023decompdiff, huang2025molform, xie2024pocketflow, luo20213d, peng2022pocket2mol, ragoza2022generating, ziv2024molsnapper}. However, most of these approaches focus on optimizing small molecules binding affinity against a single target, and do not consider the crucial selectivity aspect of the drug development process. 
Although recent efforts have begun to incorporate selectivity into their modeling objectives \cite{choi2025controllable, zou2025structure, zhang2025speclig}, a systematic evaluation of how well generative SBDD models achieve selective binding remains to be conducted and is crucial to understand the real-world utility of their outputs.

\begin{figure}[ht]
    \centering
    \vspace{-5pt}
    \includegraphics[width=0.9\linewidth]{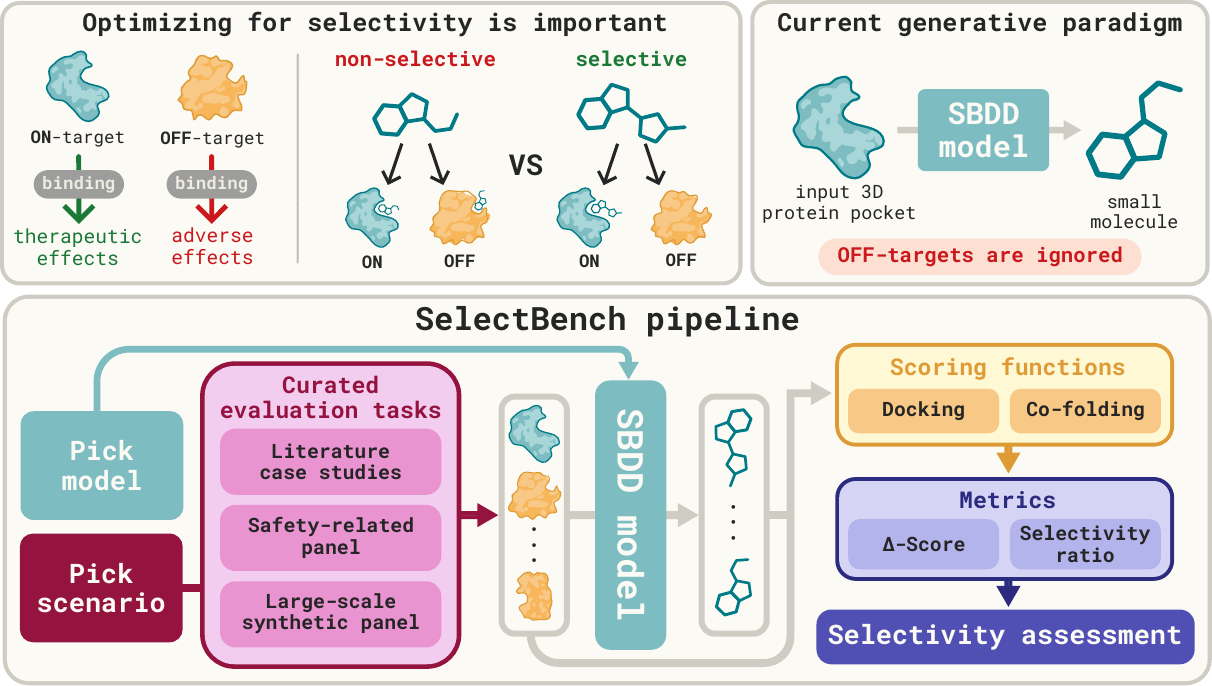}
   \mycaption{Off-target interactions are a major consideration in real-world drug design (top left) and often not accounted for in current generative SBDD (top right).}{\name provides tasks, metrics, and baselines to assess selectivity in generative SBDD (bottom).}
    \label{fig:abstract-fig}
    \vspace{-10pt}
\end{figure}

Here, we introduce \name: the first comprehensive evaluation of the selectivity of SBDD generative models (See \cref{fig:abstract-fig}). 
\name aims to address the following research questions:

\begin{custombox}[\shortstack{Research\\Questions}]{Grey2}{Grey4}
\begin{itemize}
    \item How do we align \textit{in silico} assessment of generative SBDD models with the selectivity requirements in real-world drug discovery scenarios?
    \item Can state-of-the-art SBDD generative models design selective molecules, despite not necessarily being trained to do so? Can inference-time techniques improve this?
\end{itemize}
\end{custombox}

To facilitate the investigation of these questions, we provide the following contributions in this work:

\begin{custombox}[Contributions]{Blue1}{Blue3}
\begin{itemize}[label=\textcolor{Blue3}{$\triangleright$}]
     \item \textcolor{Blue3}{\textbf{Task curation:}} We design and curate three new evaluation tasks to assess the selectivity of small molecules. These tasks cover a range of design scenarios, including real-world drug discovery use cases. We provide an easy-to-use, open framework for users to enrich \name and design their own selectivity case studies.
    \item \textcolor{Blue3}{\textbf{Establishing metrics:}} We provide a suite of metrics to quantify molecular selectivity.
    \item \textcolor{Blue3}{\textbf{Establishing baselines:}} We benchmark nine representative SBDD generative models to establish reference performance levels, as well as two inference-time steering techniques, providing a comparative framework for future works.
    \vspace{-5pt}
    
\end{itemize}
\end{custombox}

\section{Related works}

\paragraph{Selectivity in drug design} Beyond potency, selectivity is a fundamental requirement in drug discovery, both in the traditional ``one drug, one target'' paradigm and in polypharmacological settings~\citep{huggins2012rational}. It plays a key role in reducing off-target toxicity, caused by interactions with unintended proteins, as well as on-target toxicity, which can arise when the intended protein is inhibited in healthy contexts~\citep{rudmann2013target, albanese2020structure}. Designing for selectivity, however, is substantially more challenging than optimizing affinity for a single target. Prior work has identified several structure-based principles that can improve selectivity, including binding site shape complementarity, electrostatic complementarity, and conformational plasticity~\citep{huggins2012rational}. In addition, allosteric modulation (binding to a site different from the active site) has been an alternative strategy to achieve more selective control.

\paragraph{SBDD generative models}
Deep learning approaches for SBDD span several modeling classes, including autoregressive, flow-, and diffusion-based methods. Autoregressive models generate molecules atom by atom, predicting their types and positions conditioned on a protein pocket~\citep{luo2017scoring, peng2022pocket2mol}. More recently, flow- and diffusion-based methods have come to dominate the SBDD landscape~\citep{guan2023targetdiff, guan2023decompdiff, schneuing2024structure}, although they typically focus on single-target generation. There has been growing interest in incorporating binding specificity into generative models. Prior work explores this through contrastive objectives over targets~\citep{gao2024rethinking}, physics-informed energy-guided formulations that promote stable binding~\citep{choi2024pidiff}, or implicit priors that bias generation toward pocket-specific chemistry~\citep{zhang2025speclig}. 

\paragraph{Evaluation metrics in generative modeling for SBDD} 
Evaluation in generative SBDD is largely centered on single-target objectives. Common metrics include docking scores (e.g., Vina) as empirical estimates of binding affinity~\citep{eberhardt2021autodock}, alongside ligand-centric measures of drug-likeness (QED)~\citep{bickerton2012quantifying}, synthesizability (SA score)~\citep{ertl2009estimation, papidocha2026elephant}, and adherence to medicinal chemistry heuristics (e.g. Lipinski’s Rule of Five~\citep{lipinski2004lead}). More recent work has expanded evaluation to include the quality of generated 3D binding poses~\citep{harris2023benchmarking, buttenschoen2024posebusters}. However, only a limited number of studies explicitly consider selectivity, typically via docking-based comparisons across targets~\citep{gao2024rethinking, zhang2025speclig}.

\paragraph{Evaluation tasks for selective SBDD generative models} 
Evaluation of selectivity in SBDD generative models remains limited. Prior work typically relies on single case studies~\citep{schneuing2024structure, zou2025structure}, subsets of CrossDocked2020~\citep{gao2024rethinking, francoeur2020three, luo20213d}, or randomly selected off-targets from the Protein Data Bank (PDB) with low sequence similarity~\citep{zhang2025speclig}. While case studies lack statistical robustness, random target selection often lacks biological relevance. Although recent efforts have explored biologically grounded test sets (e.g., kinase panels)~\citep{choi2025controllable}, such resources are not publicly available. These limitations highlight the need for standardized, biologically relevant benchmarks for evaluating selectivity in generative SBDD models.

\section{Evaluation tasks}
\label{sec:datasets}

\subsection{Overview}

\begin{figure}[h]
    \centering
    \vspace{-5pt}
    \includegraphics[width=\linewidth]{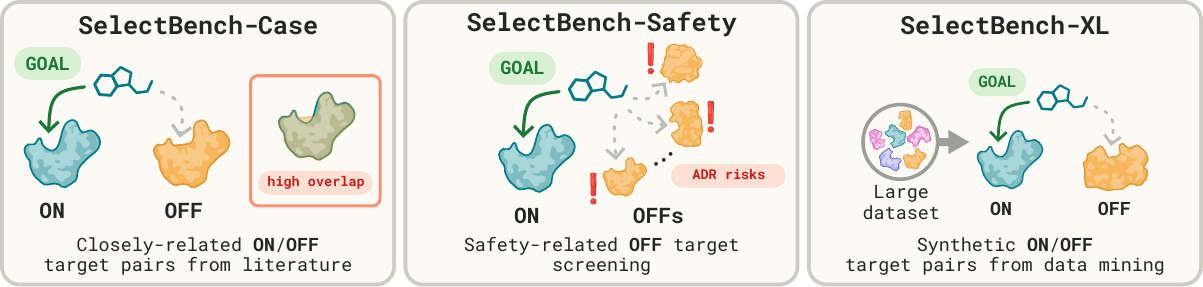}
    \mycaption{\name task summary.}{\taskA focuses on case studies, \taskB models safety screening, and \taskC enables large-scale evaluation.}
    \label{fig:dataset-summary}
    \vspace{-10pt}
\end{figure}

We propose three evaluation tasks: two manually-curated and based on literature-backed drug discovery use-cases, and a third to provide a larger-scale evaluation setting for structure-based small-molecule selectivity (see \cref{fig:dataset-summary} for a graphical summary and \cref{app:licenses} for licenses).

\begin{enumerate}
    \item \textbf{\taskA}: Five selectivity tasks based on on-/off-target pairs from the same protein family or subfamily, drawn from established literature case studies.
     \item \textbf{\taskB}: A pharmaceutical safety panel of 85 off-targets linked to adverse drug reactions (ADRs) based on secondary pharmacology assay recommendations, forming 5 selectivity tasks with one on-target and up to 85 off-targets.
    \item \textbf{\taskC}: A large-scale benchmark from CrossDocked2020 with 9,900 synthetic on-/off-target pairs, grouped into difficulty categories based on structural similarity.
\end{enumerate}

To ensure accessibility and standardization for DL applications, we provide Croissant files~\citep{akhtar2024croissant} describing dataset metadata, structure, and semantics. We describe each task in detail below; additional curation details are provided in \cref{app:curation-details}.

\subsection{\taskA: Selectivity assessment from literature-based case studies}

We identified five protein pocket pairs from the literature to study small-molecule selectivity. The pairs were chosen based on application domain and prior use as benchmarks in computational studies. The case studies are summarized in Table~\ref{tab:case-studies-summary}, with additional details for each pair in Appendix~\ref{app:case-studies-details}.

\begin{table}[htbp]
    \centering
    \vspace{-13pt}
    \mycaption{Summary of case studies in \taskA.}{}
    \label{tab:case-studies-summary}
    \resizebox{\textwidth}{!}{%
    \begin{tabular}{l l l l l}
        \toprule
        \textbf{On-Target Gene} & \textbf{On-Target Application} & \textbf{Off-Target Gene} & \textbf{Off-Target Effect} & \textbf{Computational study} \\
        \midrule
        PARP1 & Cancer \cite{ngoi2021targeting} & PARP2 & Hematological toxicity \cite{ngoi2021targeting} & CMD-GEN \cite{zou2025structure} \\
        BMP2K & Antiviral \cite{Garcia-Carceles2022-pr} & TTK & Hematological toxicity, neutropenia \cite{Aquino-Acevedo2024-el}& DiffSBDD \cite{schneuing2024structure} \\
        CDK9 & Cancer \cite{Mo2024-gq}& CDK2 & Hematological toxicity, GI issues \cite{House2025-cn}& Alchemical free-energy \cite{albanese2020structure} \\
        MAPK1 & Cancer\cite{Rebecca2014-jm}& CDK2 & Hematological toxicity, GI issues \cite{House2025-cn}& Alchemical free-energy \cite{albanese2020structure} \\
        DRD2 & Neuropsychiatric \cite{Zhang2024-ws}& HRH1 & Hypertension, neutropenia \cite{Elmore2025-ie}& Docking \cite{weiss2018selectivity} \\
        \bottomrule
    \end{tabular}
    }
    \vspace{-13pt}
\end{table}

\subsection{\taskB: Targeted selectivity for improving safety-related drug attrition}

Secondary pharmacology evaluates off-target effects unrelated to a compound's intended therapeutic action. Brennan et al. introduced Brennan-77, a panel linking drug targets to adverse events, which has informed widely used \textit{in vitro} safety screening assays in pharmaceutical research and development~\citep{brennan2024state, bowes2012reducing, lynch2017potential, albert2025comprehensive}. We constructed an \textit{in silico} analogue of such a safety panel by curating protein structures for targets in Brennan-77 (\cref{app:safety-details}). Taking into account the assay endpoints required for a comprehensive screening, we curated multiple structures in the case where both antagonistic and agonistic binding modes need to be investigated, resulting in 85 total off-target structures. Of the 85 structures, 22 correspond to the ``priority'' (high-risk) panel of 17 targets, while the remaining 63 cover the other targets in Brennan-77. We denote the 22-target subset as \taskBsmall\ and the full set as \taskBbig.

This panel enables evaluation of selectivity with respect to safety-related off-targets for any user-defined on-target protein of interest (POI). To standardize benchmarking, we selected five commonly used POIs from prior ML work: \textbf{PARP1} (Poly [ADP-ribose] polymerase-1), \textbf{FA7} (Coagulation factor VII), \textbf{5HT1B} (5-hydroxytryptamine receptor 1B), \textbf{BRAF} (Serine/threonine-protein kinase B-raf), and \textbf{JAK2} (Tyrosine-protein kinase JAK2) \cite{yang2021hit, lee2023exploring}. This set enables direct comparison with existing benchmarks while introducing selectivity as an additional challenge. To improve reproducibility, we provide a fully specified and open-sourced protein structure preparation pipeline, addressing inconsistencies in prior work (e.g., faulty 5HT1B PDBQT structures). Finally, to avoid overlap between on- and off-targets, 5HT1B is excluded from the Brennan-77 panel in our benchmark.

\subsection{\taskC: Scaling up structure-based small-molecule selectivity assessment}

Most SBDD generative models are trained on a curated CrossDocked2020 dataset~\citep{francoeur2020three, luo20213d}, which contains 100{,}000 protein--ligand pairs for training and 100 proteins for testing under a 30\% sequence-identity split. We built a large-scale selectivity benchmark from this test set by considering all non-identical protein pocket pairs and computing their structural similarity using ProBiS~\citep{konc2010probis}. Based on the resulting $\mathrm{Z}$-scores (shown in \cref{fig:probis_sb_safety}), we stratified pairs into easy ($\mathrm{Z} < 2$, $n=9834$), medium ($2 \leq \mathrm{Z} < 3$, $n=38$), and hard ($\mathrm{Z} \geq 3$, $n=28$) categories, with higher similarity corresponding to more challenging selectivity cases.

\section{Evaluation Pipeline}
\label{sec:eval-criteria}

\subsection{Overview}
\label{sec:eval-protocol}
We begin with an overview of the \name evaluation protocol. For a given generative SBDD model, we evaluate performance across all target pairs described in \cref{sec:datasets}. For each pair, the model generates 100 valid candidate molecules conditioned on the on-target 3D protein pocket (see \cref{app:generation_protocol} for more details). Inference-time guidance and selectivity-aware methods (e.g., CMD-GEN) use off-target pocket as additional conditioning input.

All generated molecules are then scored against both the on- and off-target protein pockets using the scoring functions described in \cref{sec:scoring-funcs}. These scores are aggregated into selectivity-focused evaluation metrics outlined in \cref{sec:metrics}. The set of evaluated baselines is listed in \cref{sec:model-summary}, and the overall pipeline is illustrated in \cref{fig:abstract-fig}.

\subsection{Selectivity metrics}
\label{sec:metrics}

We define our selectivity metrics in terms of a generic scoring function $f$, where $s = f(\text{ligand}, \text{pocket})$ denotes the predicted binding score for a given ligand--pocket pair. Concrete choices of scoring functions are described in \cref{sec:scoring-funcs}.

\paragraph{Delta-score $\Delta$ (per molecule)}
We define the delta-score as the difference between on- and off-target scores \citep{gao2024rethinking} from a given scoring function, with the sign chosen such that higher values indicate better selectivity:

\begin{equation*}
\Delta^{(i)} \;=\;
\begin{cases}
s_{\text{off}}^{(i)} - s_{\text{on}}^{(i)}, & \text{for scores where lower is better (e.g., docking energies)} \\
s_{\text{on}}^{(i)} - s_{\text{off}}^{(i)}, & \text{for scores where higher is better (e.g., predicted binding probabilities).}
\end{cases}
\end{equation*}

Under this convention, $\Delta^{(i)} > 0$ indicates that the ligand binds more strongly to the on-target than the off-target, reflecting selectivity toward the intended pocket. However, small differences may not correspond to meaningful selectivity; we therefore also report a stricter margin criterion $\Delta^{(i)} > \tau$. Choices of $\tau$ are explained for each scoring function. We report the mean on-target score for a population of ligands against a pocket as $\bar{s}_{\text{on}}$.

\subsubsection{Scenario I: one on-target vs.\ one off-target}

For \taskA and \taskC, each evaluation case consists of a single on-target and a single off-target. For each method, we consider the set of $N_{\text{valid}}$ generated ligands that are valid and successfully docked against both pockets. We report the following set-level statistics:

\begin{itemize}
    \item \textbf{Mean $\Delta$-score} \citep{zhang2025speclig}, $\bar{\Delta} = \frac{1}{N_{\text{valid}}}\sum_{i=1}^{N_{\text{valid}}} \Delta^{(i)}$: average selectivity gap.

    \item \textbf{Top-$k$ $\Delta$-score}, $\bar{\Delta}_{\text{top-}k}$: mean of the $k$ largest (most selective) $\Delta^{(i)}$ values, evaluating the best-performing candidates.

    \item \textbf{Selectivity rate} \citep{zhang2025speclig}, $\mathrm{SR}_{\tau} = \frac{1}{N_{\text{valid}}} \sum_{i=1}^{N_{\text{valid}}} \mathbf{1}\!\left( \Delta^{(i)} > \tau \right)$: fraction of ligands exhibiting any preference for the on-target by a margin of $\tau$.

    \item \textbf{Threshold-based selectivity rate}, $\mathrm{SR}_{\tau_{\text{on}}, \;\tau_{\text{off}}} = \frac{1}{N_{\text{valid}}} \sum_{i=1}^{N_{\text{valid}}} \mathbf{1}\!\left( s_{\text{on}}^{(i)} > \tau_{\text{on}} \;\wedge\; s_{\text{off}}^{(i)} < \tau_{\text{off}} \right)$: fraction of ligands that achieve high-confidence binding to the on-target while avoiding the off-target, based on absolute score thresholds, if known.
\end{itemize}

\subsubsection{Scenario II: one on-target vs.\ many off-targets}

The \taskB setting evaluates each ligand against a panel of $T$ anti-targets $\{t_1,\dots,t_T\}$ representing common off-target liabilities (we use $T{=}22$ for the priority panel and $T{=}84$ for the full panel). For the $i$-th ligand, we define the delta-score with respect to anti-target $t_j$ as $\Delta^{(i)}_{t_j} = s_{t_j}^{(i)} - s_{\text{on}}^{(i)}$, and the panel mean as $\bar{s}_{\text{off}}^{(i)} = \frac{1}{T}\sum_{j=1}^T s_{t_j}^{(i)}$. We then aggregate as follows:

\begin{itemize}
\item \textbf{Mean $\Delta$-score} \citep{zhang2025speclig}, $\bar{\Delta} = \frac{1}{N_{\text{valid}}}\sum_{i=1}^{N_{\text{valid}}} \big(\bar{s}_{\text{off}}^{(i)} - s_{\text{on}}^{(i)}\big)$: average selectivity gap against entire panel.

\item \textbf{Worst $\Delta$-score per molecule}, $\bar{\Delta}_{\text{worst-mol}} = \frac{1}{N_{\text{valid}}}\sum_{i=1}^{N_{\text{valid}}} \Delta_{\text{worst-mol}}^{(i)}$, 
where $\Delta_{\text{worst-mol}}^{(i)} = \min_j \Delta_{t_j}^{(i)}$: least selective (worst-case) anti-target for ligand $i$. This penalizes any single panel failure and is particularly relevant for safety screening.

\item \textbf{Average selectivity rate}, $\mathrm{SR}_{\tau}^{\mathrm{avg}} = \frac{1}{N_{\text{valid}} \cdot T} \sum_{i=1}^{N_{\text{valid}}} \sum_{j=1}^T \mathbf{1}\!\left( \Delta^{(i)}_{t_j} > \tau \right)$: fraction of ligands exhibiting preference for the on-target against the anti-target $t_j$, averaged across all anti-targets. This setting is tolerant to occasional misses.

\item \textbf{Worst selectivity rate per molecule}, $\mathrm{SR}_{\tau}^{\mathrm{worst-mol}}$: fraction of ligands satisfying $\Delta_{\text{worst-mol}}^{(i)} > \tau$; i.e., ligands that outperform \emph{all} anti-targets by at least margin $\tau$. 

\item \textbf{Top-20\% ratio} \citep{zhang2025speclig}, $\mathrm{R}_{20}$: fraction of ligands whose on-target score ranks among the top 20\% of all $T{+}1$ docking scores (on-target plus panel). This metric is invariant to the absolute scale and complements threshold-based rates.
\end{itemize}

\subsection{Scoring functions}
\label{sec:scoring-funcs}

\subsubsection{Docking-based evaluation}
\label{sec:scoring-func-docking}

For each generated ligand, we perform rigid molecular docking against both the intended (on-target) protein and a set of competing (off-target) proteins using AutoDock Vina \cite{eberhardt2021autodock}. Given a protein structure and a defined binding site, molecular docking first predicts the most energetically favorable binding pose of a ligand in complex with that protein. This corresponds to the set of 3D coordinates of each atom in the ligand relative to the target. The molecular docking scoring function estimates the associated binding energy of this pose, returning a score which reflects how strongly the ligand is predicted to bind to the target~\citep{paggi2024art}. Lower (more negative) values indicate stronger predicted binding. By comparing scores across on- and off-target proteins, we assess whether a ligand preferentially binds to the intended target. Additional details regarding docking parameters are in \cref{app:docking-details}. We report metrics for margins $\tau \in \{0, 1\}$ kcal/mol. We select $\tau = 1$ kcal/mol based on the typical accuracy limits of state-of-the-art binding free energy prediction methods \cite{albanese2020structure, aldeghi2017predictions}.

\paragraph{Z-score normalization} 
One key challenge with docking scores is that they are not calibrated across different proteins due to differences in the pocket structure and chemistry, so scores may not be directly comparable between targets~\citep{liu2024docking}. To remedy this, we normalize binding scores using Z-score normalization~\citep{kim2019improving, luo2017scoring}. For each target pocket, we compute the mean $\mu_{\text{ref}}$ and standard deviation $\sigma_{\text{ref}}$ of docking scores over a reference set of 128 drug-like ligands from ~\citep{comajuncosa2024comprehensive}. Given a generated ligand $i$ with its binding score $s^{(i)}$, we define its Z-score as

\begin{equation*}
\mathrm{Z}^{(i)} = \frac{s^{(i)} - \mu_{\text{ref}}}{\sigma_{\text{ref}}}.
\end{equation*}

This metric quantifies how the predicted binding strength of a generated ligand compares to the reference distribution for a given pocket, enabling better comparisons across targets. We report metrics for margins $\tau \in \{0, 1\}$. Setting $\tau = 1$ indicates whether the molecule is one standard deviation more favourable for the on-target than for the off-target, relative to typical molecules for each target. We denote metrics computed from Z-normalized scores with the superscript $\mathrm{z}$ (e.g., $\bar{\Delta}^{\mathrm{z}}$, $\mathrm{SR}_{\tau=1}^{\mathrm{z}}$). Positive control experiments that demonstrate the predictive power of docking-based Z-score normalized metrics can be found in \cref{app:docking-positive-control}.

\subsubsection{Predicted binding probability with Boltz-2}
\label{sec:scoring-func-boltz}

In addition to docking, we evaluate binding using Boltz-2~\citep{passaro2025boltz}, a learned co-folding model that predicts protein--ligand structure and associated binding metrics from the protein sequence and ligand representation. For each ligand--target pair, Boltz-2 produces a predicted binding probability and an affinity estimate. Note that we constrain co-folding to occur with the predefined protein binding pocket. We use the predicted binding probability as our primary score, as prior work has shown that affinity predictions exhibit weaker correlation with experimental binding measurements~\citep{bret2026assessing}. 

\paragraph{Ranking} One challenge with Boltz-2 predicted binding probabilities is that they are not calibrated across targets and may be influenced by target-specific biases. Thus, we primarily rely on ranking-based evaluations, which was shown to have good discriminative power~\citep{bret2026assessing}. For each target, we convert predicted binding probabilities to percentile-normalized ranks computed over all evaluated ligands for a given target. We indicate metrics computed from ranks with the superscript $\mathrm{rank}$ (e.g., $\mathrm{SR}^{\mathrm{rank}}_{\tau_{\mathrm{on}}=0.6, \;\tau_{\mathrm{off}}=0.4}$ is the fraction of ligands whose percentile ranks are in the top 60\% for the on-target and bottom 40\% for the off-target).

\section{Results}
We apply the evaluation protocol described in Section \ref{sec:eval-protocol} to all models in \cref{sec:model-summary} across the \name tasks. Following the research questions outlined in \cref{sec:intro}, we evaluate selective molecular design across tasks by analyzing (i) standard SBDD generative models, (ii) selectivity-focused models, and (iii) inference-time guidance. Results are presented through case studies (\cref{sec:cs_result}), safety analyses (\cref{sec:safety_result}), and SB-XL experiments (\cref{sec:xl_result}).

\subsection{Model baseline selection}
\label{sec:model-summary}

We benchmark a diverse set of SBDD models spanning multiple modeling types, including diffusion/flow-based methods, autoregressive models, and methods that incorporate selectivity. A summary of all baselines is provided in \cref{tab:case_study}; detailed descriptions are provided in \cref{app:models}, and additional RL-based methods are reported in \cref{app:cs_result_app}. We also evaluate two inference-time guidance strategies on compatible models: classifier-free guidance (CFG) and Feynman-Kac Correctors (FKC). Further descriptions on these guidance methods are provided in \cref{app:guidance}.

\begin{table}[h!]
\vspace{-10pt}
\centering
\mycaption{Summary of baseline models evaluated in \name.}{}
\label{tab:case_study}

\begin{subtable}[t]{0.9\textwidth}
\centering
\resizebox{\textwidth}{!}{%
\begin{tabular}{lcccc}

\toprule
Method name & \hlblue{Diffusion/Flow Matching} & \hlyellow{Auto-regressive} & \hlpink{Selectivity-optimized} & \hlpurple{CFG/FKC guidance} \\
\midrule

\colordot{Grey2}\ DiffSBDD       & \yes & \no & \no & \yes \\
\colordot{Blue1}\ DecompDiff     & \yes & \no & \no & \yes \\
\colordot{Blue2}\ TargetDiff     & \yes & \no & \no & \yes \\
\colordot{Blue3}\ MolForm        & \yes & \no & \no & \no \\
\colordot{Blue4}\ MolSnapper     & \yes & \no & \no & \no \\
\colordot{Yellow2}\ Pocket2Mol   & \no  & \yes & \no & \no \\
\colordot{Yellow4}\ PocketFlow   & \no  & \yes & \no & \no \\
\colordot{Pink2}\ CMD-GEN        & \yes & \no & \yes & \no \\
\colordot{Pink3}\ SpecLig        & \yes & \no & \yes & \no \\

\bottomrule
\end{tabular}
}
\end{subtable}
\vspace{-10pt}
\end{table}

\subsection{Performances on \taskA}
\label{sec:cs_result}

\begin{figure}[h]
    \centering
    \vspace{-5pt}
    \includegraphics[width=\linewidth]{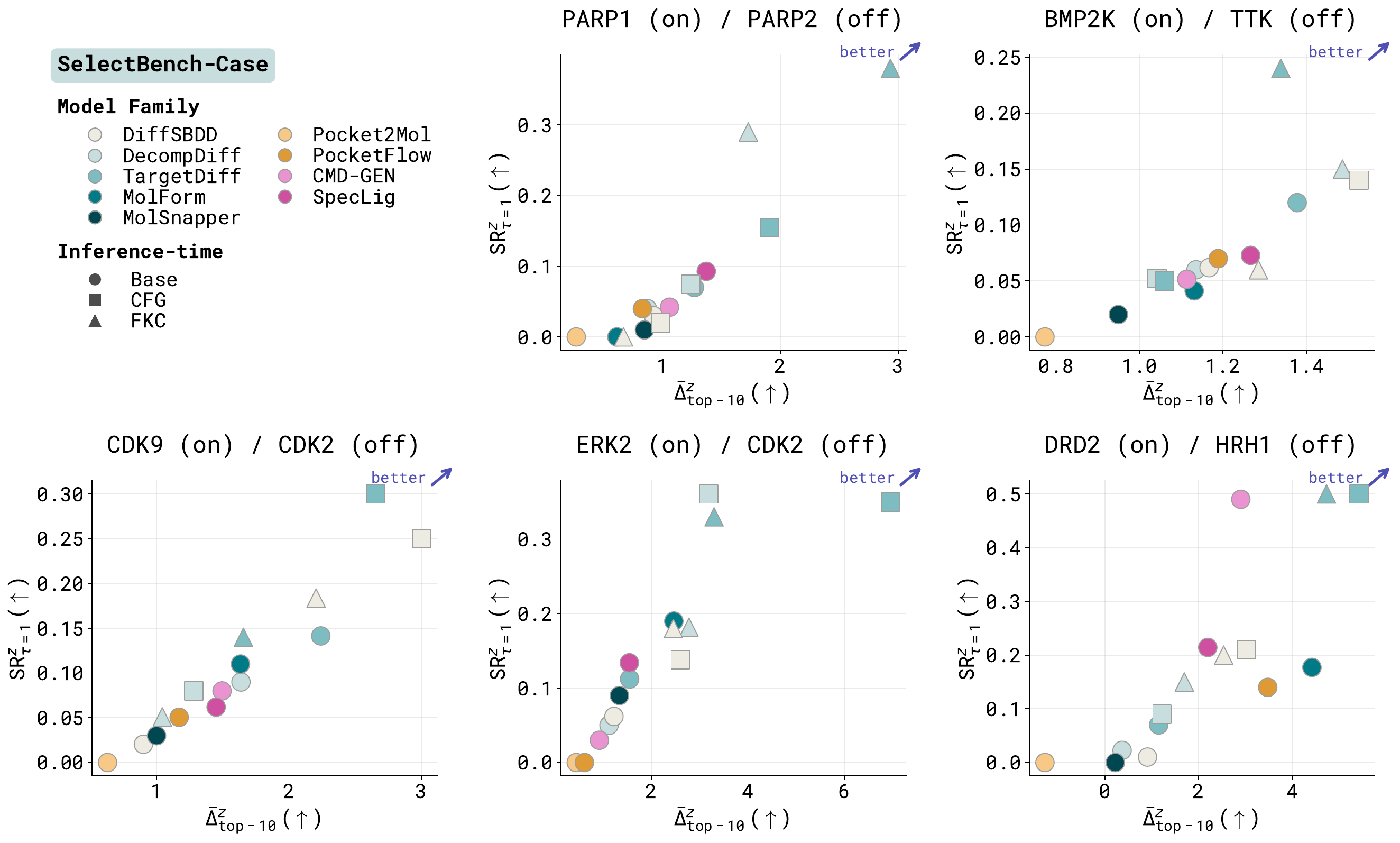}
    \mycaption{Baseline performances on \taskA}{The mean performance is reported across $\bar{\Delta}^{\mathrm{z}}_{\text{top-}10}$ ($\uparrow$) and $\mathrm{SR}^{\mathrm{z}}_{\tau=1}$ ($\uparrow$) for each baseline on each on-/off-target pair in \taskA. The color of the marker indicates the model type, and its shape indicates if inference-time methods were additionally employed. Data points in the top-right region indicate better performance.}
    \label{fig:result-plot-case}
\end{figure}

We visualize performance across two key metrics: $\bar{\Delta}^{\mathrm{z}}_{\text{top-}10}$ and $\mathrm{SR}^{\mathrm{z}}_{\tau=1}$ in \cref{fig:result-plot-case}. Most methods cluster in the lower-left region of the plot, far from the desired high-performance regime (top-right). This suggests that standard SBDD models rarely yield consistent selectivity, though highly selective candidates occasionally appear among the top-10. Across base models, the auto-regressive approaches (cf. \colordot{Yellow2} Pocket2Mol, \colordot{Yellow4} PocketFlow) tend to underperform relative to others. A possible explanation for this lower performance is the generation of consistently smaller molecules as approximated by molecular weight values reported in \cref{app:xtra_metrics_cs}, which makes binding to multiple pockets more probable.
Methods that are explicitly designed for selectivity (cf. \colordot{Pink2} CMD-GEN, \colordot{Pink3} SpecLig) show mixed results: while they sometimes show improvement over standard baselines, the performance is highly variable across different targets, and on average remains comparable to models that do not incorporate mechanisms to enhance selectivity.
The lack of consistent gains across cases highlights the importance of extensive evaluation when assessing selectivity claims.
Finally, inference-time guidance applied to 3 diffusion-based standard baselines generally leads to modest improvements, while no clear advantage emerges between FKC (\colortriangle{Grey5}) and CFG (\colorsquare{Grey5}). Despite these gains, $\mathrm{SR}^{\mathrm{z}}_{\tau=1}$ remain low (rarely exceeding 50\%), indicating that guidance improves outcomes but does not fully solve the underlying difficulty of reliably generating selective molecules. We report additional docking-based metrics and Boltz-2 metrics in \cref{app:cs_result_app}.

\subsection{Performances on \taskBsmall}
\label{sec:safety_result}

\begin{figure}[h!]
    \centering
    \includegraphics[width=\linewidth]{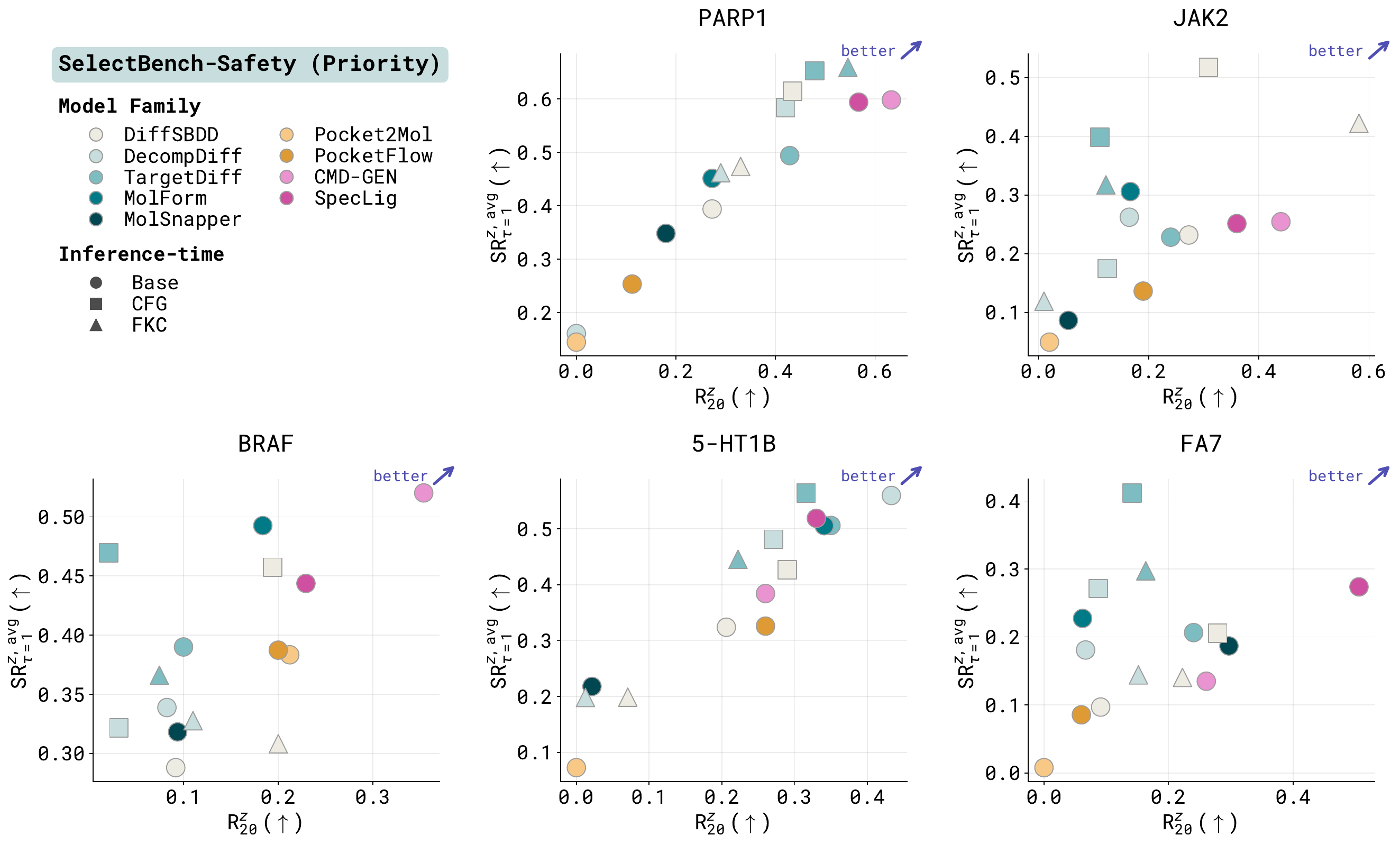}
    \mycaption{Baseline performances on \taskBsmall.}{The mean performance is reported across $\mathrm{R}_{20}^{\mathrm{z}}\,(\uparrow)$  and $\mathrm{SR}_{\tau=1}^{\text{z, avg}}\,(\uparrow)$ for each baseline on each on-target POI/priority off-targets case in \taskBsmall. The color of the marker indicates the model type, and its shape indicates if inference-time methods were additionally employed. Data points in the top-right region indicate better performance.}
    \label{fig:result-plot-safety-priority}
\end{figure}

For the safety panel, we show performance across two metrics in \cref{fig:result-plot-safety-priority} ($\mathrm{R}_{20}^{\mathrm{z}}$  and $\mathrm{SR}_{\tau=1}^{\text{z, avg}}$) for  each POI against the set of 22 priority safety targets in \taskBsmall. We report additional docking metrics in \cref{app:safety_result_priority_app}, as well as results on \taskBbig in \cref{app:safety_result_all}. 
Compared to \taskA, $\mathrm{SR}_{\tau=1}^{\text{z, avg}}$ values are substantially higher, with several settings exceeding 50\% (e.g., PARP1, BRAF, 5-HT1B). This is partly because \taskA consists of homologous targets with similar binding sites, while safety panels in \taskB often contain structurally dissimilar off-targets (see \cref{app:probis-sim}), making selectivity easier to satisfy. Note that $\mathrm{SR}_{\tau=1}^{\text{z, avg}}$ does not require selectivity against all off-targets simultaneously; under this stricter criterion, success rates remain near zero for all methods (\cref{app:safety_result_priority_app}). However, we note that factors beyond structural similarity also affect task difficulty \cite{eguida2022estimating}.
Similarly to \cref{sec:cs_result}, lower-left clustering pattern persists: standard SBDD models rarely reach the top-right regime, and autoregressive approaches tend to trail diffusion-based ones across nearly every evaluation case. However, selectivity-focused methods (cf. \colordot{Pink2} CMD-GEN, \colordot{Pink3} SpecLig) tend to perform better in this setting, suggesting improved robustness across broader targets, while within-subtype selectivity remains challenging (cf. \cref{sec:cs_result}). These results highlight the value of evaluating methods across multiple selectivity scenarios within \name. Inference-time guidance yields gains over their base counterparts on most POIs, but interpretation here is more nuanced: guidance is applied against only the single hardest off-target, while the metric averages over the full safety panel, so pushing samples away from one off-target can have mixed effects on the others. As a result, trends are less consistent than the pair-based guidance setting though this is expected by construction. Within this caveat, CFG (\colorsquare{Grey5}) more often reaches the top-right region than FKC (\colortriangle{Grey5}) (PARP1, 5HT1B, FA7). Despite these gains, $\mathrm{R}_{20}^{\mathrm{z}}$ values typically stay below 0.4, indicating that reliable safety-aware generation remains an open problem. 

\subsection{Performances on \taskC}
\label{sec:xl_result}

\begin{figure}[ht]
    \centering
    \includegraphics[width=\linewidth]{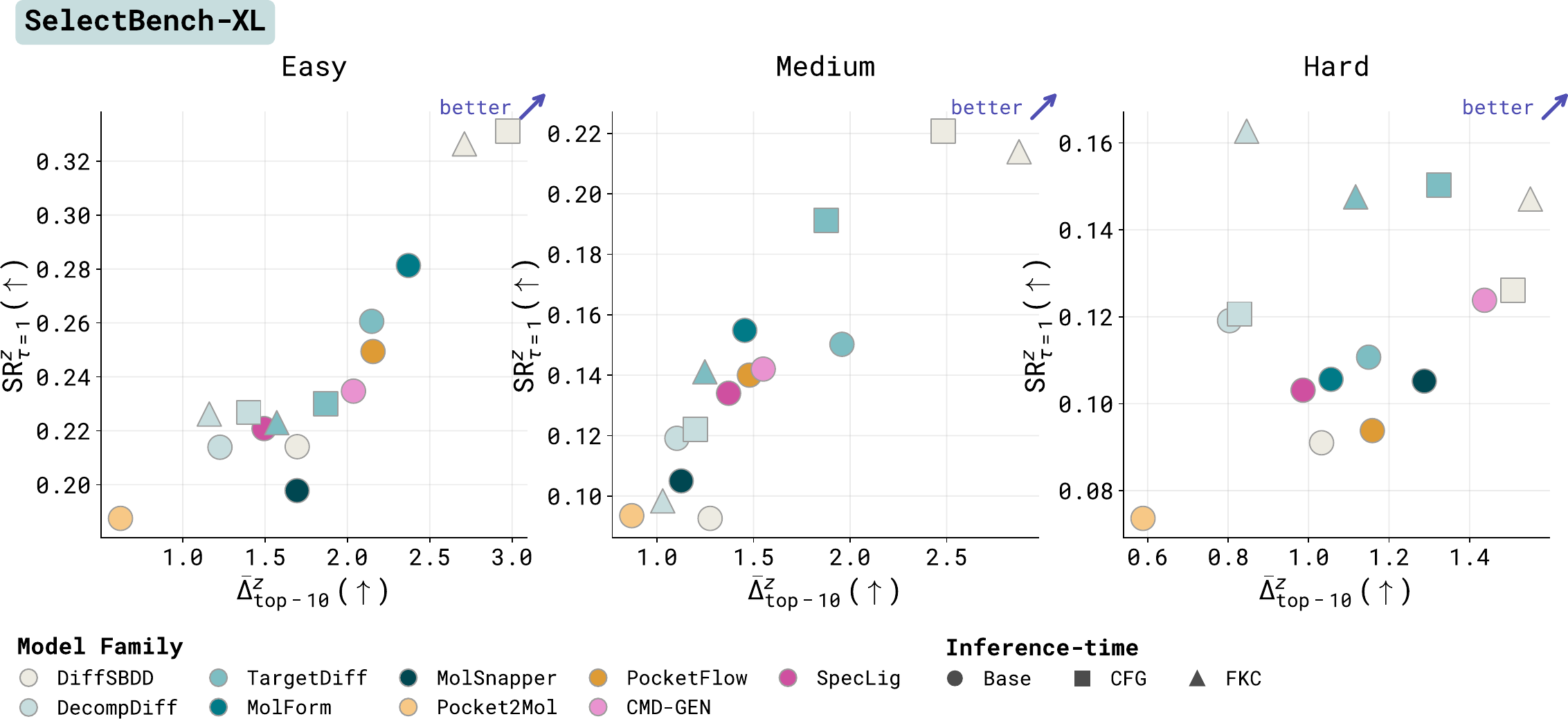}
    \mycaption{Baseline performances on \taskC}{The mean performance is reported across $\bar{\Delta}^{\mathrm{z}}_{\text{top-}10}$ ($\uparrow$) and $\mathrm{SR}^{\mathrm{z}}_{\tau=1}$ ($\uparrow$) for each baseline on each on-/off-target pairs in \taskC by difficulty. The color of the marker indicates the model type, and its shape indicates if inference-time methods were additionally employed. Data points in the top-right region indicate better performance.}
    \label{fig:result-plot-xl}
\end{figure}

We evaluate across three difficulty regimes (easy, medium, hard)\footnote{The hard and medium tasks have 28 and 38 pairs, respectively. While the benchmark contains 9834 pairs in the easy task, for ease of computing, we evaluate total of 100 pairs, which consist of five non-overlapping subsets of 20 pairs.
} and report performance across two metrics: $\bar{\Delta}^{\mathrm{z}}_{\text{top-}10}$ and $\mathrm{SR}^{\mathrm{z}}_{\tau=1}$ in \cref{fig:result-plot-xl}. Additional metrics are reported in \cref{app:xl_result_app}. Selectivity-focused methods are largely on-par with standard SBDD baselines across all regimes, with no consistent improvement in either $\bar{\Delta}^{\mathrm{z}}_{\text{top-}10}$ and $\mathrm{SR}^{\mathrm{z}}_{\tau=1}$. This echoes the \taskA finding: when off-targets are structurally close to the on-target, current explicit selectivity design fails to deliver a reliable advantage. The effects of inference-time guidance are clearer in the medium and hard regimes, with less consistent gains on easy. This is expected as easy targets are already separable, leaving less headroom for guidance to exploit. Across regimes, $\mathrm{SR}^{\mathrm{z}}_{\tau=1}$ decreases monotonically from easy to hard, indicating that the difficulty stratification captures genuine variation in the underlying selectivity challenge rather than noise, supporting \taskC as a controlled testbed for studying how methods scale with selectivity difficulty.

\subsection{Beyond selectivity: multi-objective assessment of generated molecules}
\label{sec:beyond_selectivity_result}

\begin{figure}[ht]
    \centering
    \includegraphics[width=\linewidth]{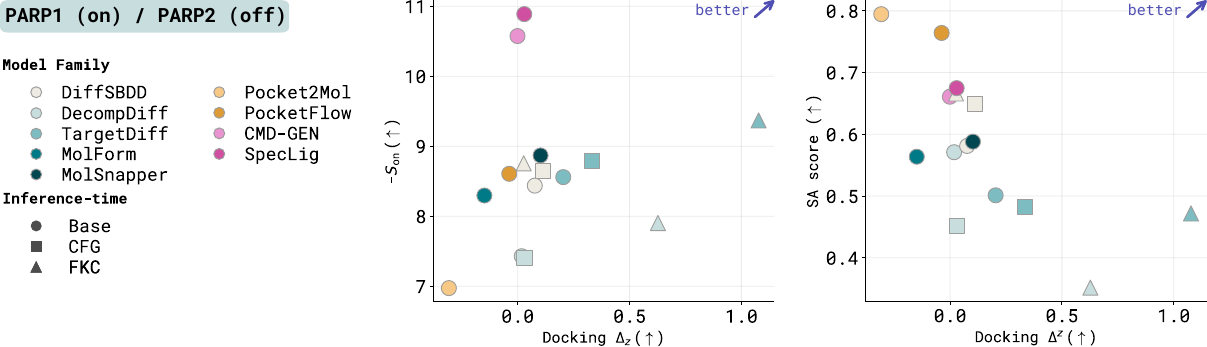}
    \mycaption{Multi-objective metrics exploration on one case study of \taskA}{The mean performance is reported across $\Delta^{\mathrm{z}}$ ($\uparrow$), $- s_{\mathrm{on}}$ ($\uparrow$) and $\mathrm{SA}$ ($\uparrow$) for each baseline on the PARP1/PARP2 case study The color of the marker indicates the model type, and its shape indicates if inference-time methods were additionally employed. Data points in the top-right region indicate better performance. The potency/selectivity trend in can be captured by introducing the potency-and-selectivity metric $\mathrm{PS} = - s_{\mathrm{on}} * \Delta^{\mathrm{z}}$ ($\uparrow$). We report $\mathrm{PS}$ across all other evaluation tasks in \name in \cref{app:beyond_selectivity_metrics}. Additional metrics assessing the quality of generated molecules beyond selectivity and potency for each task are reported in \cref{app:xtra_metrics}.}
    \label{fig:result-plot-parp-beyond-sel}
\end{figure}

It is crucial to consider the multi-objective nature of drug discovery, as gains in selectivity should not come at the expense of other desirable molecular properties, such as potency or synthesizability. We investigate the interplay among these three objectives on the \taskA PARP task in \cref{fig:result-plot-parp-beyond-sel}. We observe that selectivity-focused methods (cf. \colordot{Pink2} CMD-GEN, \colordot{Pink3} SpecLig) tend to produce the most potent molecules (on-target binders), whereas auto-regressive approaches (cf. \colordot{Yellow2} Pocket2Mol, \colordot{Yellow4} PocketFlow) tend to produce the most synthesizable molecules, although at the cost of selectivity. Applying selectivity-focused guidance helps improve potency of molecules but does not improve their synthesizability, highlighting a need for developing improved multi-objective guidance methods that can jointly optimize selectivity, potency, and synthesizability (among other properties).

\section{Conclusion}
We introduce \name, a benchmark that characterizes the state of generative SBDD on selectivity and provides an open platform to accelerate progress. It comprises three curated tasks reflecting distinct drug discovery scenarios, along with baselines spanning widely-used SBDD models, selectivity-trained models, and inference-time guidance methods. Our results show that diffusion and flow models generally outperform autoregressive approaches in selectivity. While selectivity-aware methods improve performance in one-to-many settings on average (cf. \taskB), they offer limited gains in high-homology regimes where targets are structurally similar (cf. \taskA, \taskC). Inference-time guidance further enhances diffusion models, but robust selectivity in all settings remains challenging, especially when considering multi-objective generation with other properties like potency and synthesizability. Overall, \name highlights substantial room for improvement toward reliable selective molecular generation. 

\section{Limitations}

Accelerating SBDD with deep learning requires generative models to better capture its inherently multi-objective nature. While \name provides a foundation for systematically evaluating selectivity in target-conditioned molecular generation, several limitations remain. First, current tasks do not account for allosteric effects, which are important for realistic drug discovery scenarios. Second, expanding synthetic tasks to cover a larger portion of CrossDocked2020 would improve scale and diversity. Third, \name can be extended to incorporate additional scoring functions as their accuracy and computational efficiency improve. We note that commonly used proxies such as docking scores and predicted affinities can deviate from real-life experimental measurements; however, they are currently the most practical tools for large-scale screening.
Finally, we note that ligand quality extends beyond the objectives considered here. Other important factors include the quality and physical plausibility of the 3D binding pose, and other drug factors~\citep{buttenschoen2024posebusters, harris2023benchmarking, lipinski2004lead}. Extending multi-objective evaluation of generated molecules beyond selectivity, potency, and synthesizability therefore represents an important direction for future work.

\section*{Acknowledgements}
\label{sec:acknowledgements}
\looseness=-1

The authors would like to thank Chris Radoux for insightful discussions. The research was enabled in part by computational resources provided by the Digital Research Alliance of Canada (\url{https://alliancecan.ca}), Mila (\url{https://mila.quebec}), the SciNet HPC Consortium (\url{https://scinethpc.ca/}) and was partially sponsored by Google through the Google \& Mila projects program. E.M.R. was supported by the Vector Institute. A.A.-G. thanks Anders~G.~Fr{\o}seth for his generous support. A.A.-G. also acknowledges the generous support of Natural Resources Canada and the Canada 150 Research Chairs program. This research is part of the University of Toronto’s Acceleration Consortium, which receives funding from the CFREF-2022-00042 Canada First Research Excellence Fund. This work was supported by the Defense Advanced Research Projects Agency (DARPA) under Agreement No. HR0011262E022. In addition, K.N. was supported by IVADO and Institut Courtois, and M.S. was supported by the IVADO 2025 Postdoctoral Research Funding Program. This project was undertaken thanks to funding from IVADO and the Canada First Research Excellence Fund.

\bibliography{lib}

\newpage
\appendix
\renewcommand{\thefigure}{A\arabic{figure}}  
\renewcommand{\thetable}{A\arabic{table}}    
\setcounter{figure}{0}  
\setcounter{table}{0}   

\clearpage
\hypertarget{appendixtoc}{} 

\startcontents[appendix]
\printcontents[appendix]{}{1}{\section*{Appendix}}
\thispagestyle{plain}   

\pagestyle{toc}
\clearpage
\section{Reproducibility details}

\subsection{Evaluation tasks licenses}
\label{app:licenses}

We list the different licenses of the datasets curated in \name for each evaluation task below:

\begin{itemize}
    \item \taskA: CC BY 4.0 
    \item \taskB: CC BY 4.0
    \item \taskC: CC BY 4.0
\end{itemize}

To create the datasets in \name, we used existing datasets for which the licenses are listed below:

\begin{itemize}
    \item RCSB PDB \cite{berman2000protein}: CC0 1.0
    \item UniProt \cite{bateman2024uniprot}: CC BY 4.0
    \item CrossDocked2020 (Luo test set) \cite{francoeur2020three, luo20213d}: MIT
    \item MOOD POIs \cite{lee2023exploring}: MIT 
\end{itemize}

\subsection{Compute resources}
All model inferences were conducted on a single NVIDIA L40S GPU. Molecular docking was run on CPU, with 8 cores and 32 GB of RAM allocated per docking job.

\section{Additional information on \name tasks}

\subsection{Task curation pipeline}
\label{app:curation-details}

The following points highlight task design choices we made:
\begin{itemize}
    \item \textbf{Protein structure selection} We conducted literature-based manual curation for \taskA and \taskB protein structure selection, prioritizing quality of information over size. For \taskA, PDB IDs were retrieved according to original publications. For \taskB, relevant gene names were first obtained from Brennan et al. \cite{brennan2024state} and converted to UniProt IDs (UPIDs) \cite{bateman2024uniprot}. The PDB was searched for each target according to its UPID. The structures were selected based on holo-ness of their orthosteric binding site, ensuring that no mutations in the binding site were present. For the on-target POIs, PDB IDs were retrieved based on original publications. For \taskC, PDB structures of the CrossDocked test set were retrieved.

    \item \textbf{Binding site selection} For \taskA, \taskC and the on-target POIs of \taskB, we maintained binding site definitions from previous publications. For off-targets of \taskB, we defined a unique orthosteric binding site per protein target, leaving allosteric binding site consideration for future work. When necessary, and if available in the PDB, we considered both agonistic and antagonistic-bound protein target structures. 

    \item \textbf{File preparation} Target and ligand file preparation constitute a crucial step in SBDD computational pipelines \cite{paggi2024art, bender2021practical}. We kept protein chains present within a $10\mathring{\text{A}}$ box around the ligand centroid, remove water molecules and used \texttt{Meeko} for molecular docking receptor preparation \cite{santos2025meeko}. Ligands were prepared for molecular docking from SMILES strings on-the-fly using \texttt{RDKit} and \texttt{Meeko} \cite{santos2025meeko}. To ensure reproducibility and standardization of further community efforts, the file preparation code is openly released alongside each task.
\end{itemize}

\subsection{\taskA pair details}
\label{app:case-studies-details}

\paragraph{PARP1 (on) / PARP2 (off)} Poly [ADP-ribose] polymerase 1 (PARP1) emerges as a crucial target in cancer therapy \cite{zou2025structure}. All current FDA approved PARP inhibitors non-selectively bind and inhibit both PARP1 and PARP2 \citep{pilie2019parp}. The selective inhibition of PARP1 alone has been shown to induce cell death in homologous recombinant-deficient cancer cells, while the inhibition of PARP2 is associated with hematological toxicity \citep{ngoi2021targeting}. Therefore, there is an urgent need for the development of selective inhibitors targeting PARP1. This pair of polymerases was studied in CMD-GEN, a computational framework for inhibitor design \cite{zou2025structure}.

\paragraph{BMP2K (on) / TTK (off)} Inhibition of some proteins of the numb-associated kinase (NAK) family has broad spectrum antiviral activities \cite{Garcia-Carceles2022-pr}. The design of highly selective kinase inhibitors remains a challenge as closely related kinases share significant structural homology \citep{karaman2008quantitative}. Schneuing et al. previously investigated designing selective kinase inhibitor against BMP2 inducible kinase (BMP2K) while simultaneously minimizing off-target activity against Monopolar Spindle Kinase 1 (TTK), a structurally similar kinase \citep{schneuing2024structure}. Inhibition of TTK is associated with effects such as hematological toxicity and neuropenia \cite{Aquino-Acevedo2024-el}.

\paragraph{CDK9 (on) / CDK2 (off)} CDK9 as a therapeutic target has demonstrated potential in both hematological malignancies and solid tumours \cite{Mo2024-gq}. As mentioned previously, achieving selective inhibition of kinases is a challenging task. Albanese et al. benchmarked alchemical free-energy methods for compound selectivity prediction \cite{albanese2020structure} on the very similar binding sites of Cyclin-dependent kinase 2 (CDK2) and CDK9. Inhibition of CDK2 is associated with effects such as hematological toxicity and gastrointestinal issues \cite{House2025-cn}.

\paragraph{ERK2 (on) / CDK2 (off)} Albanese et al. also considered the slightly less related study of selective inhibition of Mitogen-activated protein kinase 1 (MAPK1), also known as ERK2, compared to CDK2 \cite{albanese2020structure}. Just like CDK9, ERK2 has demonstrated potential in treating cancer \cite{Rebecca2014-jm}. While three kinases (CDK2, CDK9 and ERK2) belong to the CMGC family, CDK2 and CDK9 belong to the CDK (cyclin-dependent kinase) subfamily while ERK2 is part of the nearby MAPK (mitogen-activated protein kinases) subfamily. Multiple-sequence alignment reveals that CDK2 and ERK2 share 52\% sequence identity.

\paragraph{DRD2 (on) / HRH1 (off)} The dopamine receptor is a crucial target for treatment of neuropsychiatric disorders. We simplify the dual antagonist of serotonin 5-HT2A (HTR2A) and dopamine D2 (DRD2) with histamine H1 (HRH1) as off-target campaign of Weiss et al. \cite{weiss2018selectivity}, taking for objective to have DRD2 as the only on-target and HRH1 as off-target. Inhibition of HRH1 is associated with hypertension and neutropenia \cite{Elmore2025-ie}.

\subsection{\taskB design choices}
\label{app:safety-details}

The following points highlight task design choices we made:
\begin{itemize}
    \item We prioritize for \textit{holo}-ness -- ligand-bound protein conformation -- in structure over resolution, and limit our curation to orthosteric sites -- primary, conserved binding location.
    \item Out of the 77 targets in the Brennan-77 panel, 4 targets have no \textit{holo} structure in the PDB (COX1, TACR3, CCKBR, STK35), we therefore exclude these targets from our analysis.
\end{itemize}

\subsection{\taskC Easy pairs random sampling}
\label{app:easy-pairs}
The hard and medium tasks have 28 and 38 pairs, respectively.

The easy task contains 9,834 pairs; for tractability, we evaluate total of 100 pairs, which consist of five non-overlapping subsets of 20 easy pairs each. Each subset uses 20 distinct on-target pockets, no pocket is the on-target in more than one subset, and together the subsets cover all 100 pockets exactly once. The subsets are released with the benchmark. Example of the randomly selected 20 pairs are listed in \cref{tab:easy_pairs}. 

\begin{table}[ht]
\centering
\setlength{\tabcolsep}{4.5pt}
\caption{Easy pairs sampled from \taskC}
\label{tab:easy_pairs}
\begin{tabular}{ccc|ccc|ccc}
\hline
\textbf{ID} & \textbf{On-target} & \textbf{Off-target} & \textbf{ID} & \textbf{On-target} & \textbf{Off-target} & \textbf{ID} & \textbf{On-target} & \textbf{Off-target}\\
\hline
717  & 4tqr\_A & 14gs\_A & 3498 & 1r1h\_A & 2rma\_A & 2308 & 1coy\_A & 2pqw\_A \\
823  & 1h0i\_A & 14gs\_A & 4773 & 5mgl\_A & 4keu\_A & 2508 & 2jjg\_A & 1djy\_A \\
981  & 4rlu\_A & 2v3r\_A & 4886 & 4q8b\_B & 3ej8\_A & 2515 & 2rma\_A & 2jjg\_A \\
2167 & 4zfa\_A & 3g51\_A & 5895 & 1k9t\_A & 4p6p\_A & 2978 & 2pqw\_A & 3pnm\_A \\
2219 & 4kcq\_A & 3g51\_A & 6354 & 4lfu\_A & 3l3n\_A & 3311 & 2pc8\_A & 5bur\_A \\
8550 & 1ai4\_A & 3hy9\_B  & 8989 & 4gvd\_A & 5mgl\_A & 9741 & 3af2\_A & 3u9f\_C \\
9744 & 3u9f\_C & 3li4\_A & 9870 & 1jn2\_P & 3li4\_A & & & \\
\hline
\end{tabular}
\end{table}

\section{Additional information on metrics}
\subsection{Docking set-up}
\label{app:docking-details}

All rigid molecular docking calculations were conducted using Autodock Vina \cite{eberhardt2021autodock}. The number of poses was set to 5. The exhaustiveness parameter was set to the default parameter of 20 for all evaluations conducted on \taskA,  set to 10 for all evaluations conducted on \taskC, and set to 1 for all evaluations conducted on \taskB. Docking boxes were centered on the centroid of the original ligand, their size in the x-, y-, z-axis were chosen to be 20 $\mathring{\text{A}}$ for \taskA, \taskC and for the off-targets in \taskB. Original x-, y-, z-axis docking boxes sizes were retained for on-target POIs in \taskB.

\subsection{Docking positive control experiments}
\label{app:docking-positive-control}

To ensure our docking-based scoring function can reliably identify selectivity, we conducted positive control experiments on two of the case studies included in \taskA by recovering experimentally validated selectivity trend for ligands in original literature with our docking-based scoring pipeline. The case studies were selected based on abundance of experimental data availability (eliminating BMPK/TTK) and prioritization of harder case studies on the basis of structural similarity (cf. \cref{app:probis-sim}). The SMILES of the ligand were extracted from the molecular graph images found in the original literature using a LLM agent and manually checked using a SMILES visualizer. They are made available in our GitHub repository for reproducibility.

\paragraph{PARP1 (on) / PARP2 (off)} We docked the Z2, Z5, Z6, Y1, Y2, Y4, Y5 and Y6 ligands as well as the ligand depicted in Figure 8d of the CMD-GEN paper \cite{zou2025structure} against the PARP1 (on) and PARP2 (off) structures in \taskA. We report our results in \cref{tab:pos-control-parp}.

{\footnotesize
\setlength{\tabcolsep}{4.75pt}
\renewcommand{\arraystretch}{1.00}
\begin{longtable}{cccccccc}
\mycaption{PARP1/PARP2 positive control evaluation}{We report: the mean on-target Vina score $\bar{s}_{\text{on}}$ ($\downarrow$); $\bar{\Delta}$ over selected molecules and the selectivity ratio $\mathrm{SR}_{\tau}$ for $\tau \in \{0, 1\}$ kcal/mol. Additionally, we report the z-score normalized analogues of the corresponding raw metrics, standardized against a reference score distribution: $\bar{s}_{\text{on}}^{z}$, $\bar{\Delta}^{z}$, $\mathrm{SR}^{z}_{\tau=0}$ and $\mathrm{SR}^{z}_{\tau=1}$.}
\label{tab:pos-control-parp} \\

\toprule
$\bar{s}_{\text{on}}$ 
& $\bar{\Delta}$ 
& $\mathrm{SR}_{\tau=0}$ 
& $\mathrm{SR}_{\tau=1}$ 
& $\bar{s}_{\text{on}}^{z}$ 
& $\bar{\Delta}^{z}$ 
& $\mathrm{SR}^{z}_{\tau=0}$
& $\mathrm{SR}^{z}_{\tau=1}$ \\
\midrule
$-11.008 \pm 0.969$
& $-0.033 \pm 0.502$
& $0.556$
& $0.000$
& $-1.631 \pm 1.009$
& $0.207 \pm 0.540$
& $0.667$
& $0.111$ \\
\bottomrule

\end{longtable}}

We see that on average, $\bar{\Delta}^{z}$ recovers the selectivity trend in the correct direction while $\bar{\Delta}$ fails to do so, highlighting the importance of Z-score normalization. $\mathrm{SR}^{z}_{\tau=0}$ shows a selectivity rate of 66.7\%, showing that docking is able to recover a large proportion of selective molecules (although not all – this is expected as docking remains a proxy scoring function). The selectivity rate $\mathrm{SR}^{z}_{\tau=1}$ drops significantly when picking a stricter threshold $\tau$, suggesting that the positive signal of docking is sensitive to $\tau$. This surfaces a trade-off between having a stricter threshold that could yield false negatives and a looser threshold that could yield false positives. We consider the former option to be better suited to challenge future model development, and therefore focus on $\mathrm{SR}^{z}_{\tau=1}$ in our analysis in the main text.

\paragraph{CDK9 (on) / CDK2 (off)} We picked 4 selective (12 a,c,e,f) and 4 non-selective ligands (12 h,l,o,t) from the paper of Albanese et al. \cite{albanese2020structure} and docked them against the CDK9 (on) and CDK2 (off) structures in \taskA. We report our results in \cref{tab:pos-control-cdk9}.

{\footnotesize
\setlength{\tabcolsep}{0.5pt}
\renewcommand{\arraystretch}{1.00}
\begin{longtable}{lcccccccc}
\mycaption{CDK9/CDK2 positive control evaluation}{We report: the mean on-target Vina score $\bar{s}_{\text{on}}$ ($\downarrow$); $\bar{\Delta}$ over selected molecules and the selectivity ratio $\mathrm{SR}_{\tau}$ for $\tau \in \{0, 1\}$ kcal/mol. Additionally, we report the z-score normalized analogues of the corresponding raw metrics, standardized against a reference score distribution: $\bar{s}_{\text{on}}^{z}$, $\bar{\Delta}^{z}$, $\mathrm{SR}^{z}_{\tau=0}$ and $\mathrm{SR}^{z}_{\tau=1}$.}
\label{tab:pos-control-cdk9} \\

\toprule
Ligands
& $\bar{s}_{\text{on}}$
& $\bar{\Delta}$
& $\mathrm{SR}_{\tau=0}$
& $\mathrm{SR}_{\tau=1}$
& $\bar{s}_{\text{on}}^{z}$
& $\bar{\Delta}^{z}$
& $\mathrm{SR}^{z}_{\tau=0}$ 
& $\mathrm{SR}^{z}_{\tau=1}$ \\
\midrule
CDK9-selective
& $-8.050 \pm 0.121$
& $-0.691 \pm 0.365$
& $0.000$
& $0.000$
& $-0.117 \pm 0.166$
& $0.172 \pm 0.439$
& $0.750$
& $0.000$ \\
CDK9 non-selective
& $-7.822 \pm 0.741$
& $-1.074 \pm 0.594$
& $0.000$
& $0.000$
& $0.195 \pm 1.014$
& $-0.323 \pm 0.648$
& $0.250$
& $0.000$ \\
\bottomrule
\end{longtable}}

We see that $\bar{\Delta}$ still fails to recover selectivity trends, while successfully recovers both positive and negative selectivity trends in the correct direction. Importantly, we observe that $\mathrm{SR}^{z}_{\tau=0}$ is high for CDK9-selective molecules (75\%) and substantially lower for molecules known to be non-selective (25\%), demonstrating the desired behavior of a benchmark designed to quantify selectivity. We also observe a selectivity rate drop when picking a stricter threshold of $\tau = 1$

Overall, these results obtained with experimentally validated ligands on two cases studies from \taskA highlight that our normalized metrics obtained via docking score can recover meaningful selectivity trends but that limits also arise, as expected and stated in our limitations section.

\section{Baseline descriptions}
\label{app:models}

\subsection{Standard SBDD models}
We evaluate four diffusion, two auto-regressive and one flow-matching based SBDD methods that generate 3D molecules conditioned on a protein binding pocket.

\textbf{TargetDiff} \citep{guan2023targetdiff} is an SE(3)-equivariant diffusion model that jointly diffuses over atom positions (continuous) and atom types (discrete) within a binding pocket, learning to denoise both modalities simultaneously to produce chemically plausible molecules.       

\textbf{DecompDiff} \citep{guan2023decompdiff} extends this framework by incorporating decomposed molecular priors based on Murcko scaffold decomposition, factoring the generation process into arms and scaffold substructures to better capture the compositional nature of drug-like molecules.

\textbf{DiffSBDD} \citep{schneuing2024structure} is an equivariant diffusion model that supports both \textit{de novo} generation conditioned on a pocket and scaffold-constrained generation via an inpainting procedure, where parts of a known ligand can be fixed while the model generates the remainder.

\textbf{MolSnapper} \citep{ziv2024molsnapper} builds upon MolDiff, an unconditional molecular diffusion model that does not take pocket features as network input. Instead, pocket conditioning is applied externally during the sampling process through pharmacophore-based position masking and gradient-based steric clash penalties, steering the unconditional generation toward pocket-compatible molecules.

\textbf{MolForm} \citep{huang2025molform} replaces the discrete diffusion framework with continuous normalizing flows via flow matching, jointly generating atom positions and types along learned vector field trajectories. This formulation offers straighter sampling paths and potentially more efficient generation compared to diffusion-based counterparts.

\textbf{Pocket2Mol} \citep{peng2022pocket2mol} employs an E($n$)-equivariant graph neural network to autoregressively place atoms one at a time, predicting each atom's element, position, and bond connectivity conditioned on both the pocket and previously placed atoms.

\textbf{PocketFlow} \citep{xie2024pocketflow} is a knowledge-guided autoregressive model that decomposes generation into sequential prediction of atom types, positions, and bond orders, achieving 100\% chemical validity by construction through its constrained decoding procedure. The method has been validated through wet-lab experiments.

\subsection{Selectivity-trained models}
We additionally compare against two methods explicitly designed to generate ligands with target selectivity.

\textbf{CMD-GEN} \citep{zou2025structure} is a multi-stage pipeline consisting of three components: (1)DiffPhar, a diffusion model that samples 3D pharmacophore point clouds for both on-target and off-target pockets; (2) a Gaussian mixture model clustering step that reduces these point clouds to key pharmacophoric features; and (3) GCPG, a pharmacophore-constrained 2D molecule generator. Selectivity is achieved by aligning the on- and off-target pharmacophore representations via ICP and Kabsch alignment, then retaining only those pharmacophoric features unique to the on-target pocket.

\textbf{SpecLig} \citep{zhang2025speclig} operates in a learned latent space and employs energy-guided latent diffusion, where physics-based binding energy terms serve as guidance signals to steer the generative process toward molecules with preferential affinity for the desired target. The method supports both small-molecule and peptide generation through a fragment-based decoding scheme.

\subsection{Inference-time guidance methods}

\label{app:guidance}
Structure-based diffusion models learn to generate molecules conditioned on a single protein pocket. A natural question is whether we can steer generation at inference time to produce molecules that are \emph{selective}---binding strongly to an on-target while avoiding an off-target. We investigate two principled approaches derived from the Feynman-Kac Correctors (FKC)~\citep{skreta2025feynmankac}.

\paragraph{Setup}
Let $q_t^{(1)}(x)$ and $q_t^{(2)}(x)$ denote the noised marginal densities of a pretrained diffusion model conditioned on the on-target and
off-target protein pockets, respectively. Both share the same model weights---only the protein context differs. The standard reverse-time stochastic differential equation (SDE) for a single pocket $i$ is
\begin{equation}
  \mathrm{d}x_t = \big[-f_t(x_t) + \sigma_t^2 \nabla \log q_t^{(i)}(x_t)\big]\mathrm{d}t + \sigma_t \mathrm{d}\bar{W}_t,
  \label{eq:reverse-sde}
\end{equation}
where $f_t(x) = -\tfrac{1}{2}\beta(t)x$ is the linear drift and $\sigma_t = \sqrt{\beta(t)}$ for a variance preserving (VP) SDE noise schedule.

\paragraph{Selective target distribution}
We define the selective target as a weighted product of the two marginals (Proposition~D.5 of~\citep{skreta2025feynmankac}):
\begin{equation}
  p_t(x) \propto q_t^{(1)}(x)^{\beta_1} \cdot q_t^{(2)}(x)^{\beta_2}, \quad \beta_1 > 0, \; \beta_2 < 0.
  \label{eq:selective-target}
\end{equation}
The positive exponent $\beta_1$ amplifies the on-target density, while the negative exponent $\beta_2$ suppresses the off-target density. The score of this target distribution decomposes as $\nabla \log p_t(x) = \beta_1 \nabla \log q_t^{(1)}(x) + \beta_2 \nabla \log q_t^{(2)}(x)$.

In what follows, we describe two methods for approximately sampling from~\eqref{eq:selective-target}: a simple classifier-free guidance (CFG) approach and a more principled FKC approach with importance-weight corrections.

\subsubsection{CFG selective guidance}
Classifier-free guidance~\citep{ho2022classifier} is a widely used heuristic for steering diffusion model outputs by combining predictions from two models. In the standard image generation setting, CFG interpolates between an unconditional and a conditional model using a guidance weight $\beta$. This corresponds to attempting to sample from the geometric average $p_t^{\text{geo}}(x) \propto q_t^{(1)}(x)^{\beta_1} q_t^{(2)}(x)^\beta_2$ (Proposition~3.1 of \citep{skreta2025feynmankac}), though the standard SDE simulation does not exactly match the prescribed intermediate marginals.

We adapt this idea for selectivity. At each reverse diffusion step, the model predicts the clean position $\hat{x}_0$ — and, where the model uses a discrete atom-type channel, the predicted-clean-type log-probabilities $\log\hat{p}_0(a)$ — conditioned on each pocket independently.\footnote{TargetDiff~\citep{guan2023targetdiff} and DecompDiff~\citep{guan2023decompdiff} use a dual-process diffusion with a categorical atom-type channel; DiffSBDD~\citep{schneuing2024structure} encodes atom types as soft one-hot vectors noised jointly with continuous positions, so for it Eq.~\ref{eq:cfg-atomtype} is absorbed into Eq.~\ref{eq:cfg-x0}.} We combine the two pockets' predictions in real space and in log-probability space respectively:
\begin{equation}
\hat{x}_0^{\text{guided}} \;=\; \beta_1\,\hat{x}_0^{(1)} \;-\; \beta_2\,\hat{x}_0^{(2)},
\label{eq:cfg-x0}
\end{equation}

\begin{equation}                 
\log\hat{p}_0^{\text{guided}}(a) \;\propto\; \beta_1\,\log\hat{p}_0^{(1)}(a) \;-\;
\beta_2\,\log\hat{p}_0^{(2)}(a).
\label{eq:cfg-atomtype}
\end{equation}

The guided position is then propagated by one step of the reverse-time SDE, and the atom-type channel by the multinomial posterior on $\log\hat{p}_0^{\text{guided}}$. This requires nothing beyond the per-pocket forward passes and a linear combination of their outputs, but provides no correction for the mismatch between the guided trajectory and the true selective-target marginals.

\subsubsection{FKC selective guidance}
Rather than simply combining predictions heuristically, we can follow the \emph{weighted SDE} that accounts for the discrepancy between the simulation trajectory and the desired target through importance weights. Using the general weighted product result (Proposition~D.5 of~\citep{skreta2025feynmankac}), the selective target~\eqref{eq:selective-target} admits the following weighted SDE:
\begin{align}
  \mathrm{d}x_t &= \left[-f_t(x_t) + \sigma_t^2 \left(\beta_1 \nabla \log q_t^{(1)}(x_t) + \beta_2 \nabla \log
q_t^{(2)}(x_t)\right)\right] \mathrm{d}t + \sigma_t \,\mathrm{d}\bar{W}_t, \label{eq:fkc-drift} \\[4pt]
  \mathrm{d}w_t &= \left[
      \frac{\sigma_t^2}{2} \left\| \beta_1 \, s_t^{(1)} + \beta_2 \, s_t^{(2)} \right\|^2
      - \frac{\sigma_t^2}{2} \left( \beta_1 \left\| s_t^{(1)} \right\|^2 + \beta_2 \left\| s_t^{(2)} \right\|^2 \right)
  \right] \mathrm{d}t, \label{eq:fkc-weight}
\end{align}
where $s_t^{(i)} = \nabla \log q_t^{(i)}(x_t)$ denotes the score function for pocket $i$, and we have dropped the constant divergence term $(\beta_1 + \beta_2 - 1)\langle \nabla, f_t(x_t) \rangle$ which cancels upon reweighting for linear drifts.

\paragraph{Sequential Monte Carlo resampling}
To translate the importance weights into improved samples, we apply Sequential Monte Carlo (SMC) resampling over a batch of $K$ particles. At each diffusion step within an active interval $[t_{\min}, t_{\max}]$, particles are resampled proportional to $\exp(w_t^{(k)})$ using systematic resampling~\citep{douc2005comparison}. This selectively duplicates particles with high weights (those achieving good selectivity) and eliminates low-weight particles, progressively steering the population toward the selective target~\eqref{eq:selective-target}.
In practice, we activate resampling only during the later diffusion steps (e.g., the final 60\% of steps), as early steps operate at high noise levels where the score signals are weak.

\paragraph{Choice of exponents}
The exponents $\beta_1$ and $\beta_2$ control the strength of on-target attraction and off-target repulsion, respectively. A symmetric choice $\beta_2 = -\beta_1$ applies equal strength to both, but we find that aggressive negative guidance (e.g., $\beta_1 = 1, \beta_2 = -1$) can push atom positions into unrealistic configurations, degrading molecular validity. In practice, we choose asymmetric exponents ($\beta_1 = 2.5$, $\beta_2 = -0.5$ and $\beta_1 = 7.0$, $\beta_2 = -2.0$) which produce more chemically valid molecules while still conferring measurable selectivity.

\subsection{RL-based finetuning methods}
\textbf{REINVENT} \citep{olivecrona2017,loeffler2024reinvent4} is a ligand-based
autoregressive SMILES generator fine-tuned by reinforcement learning against a
user-defined scoring function. A prior trained on ChEMBL is adapted through an
augmented-likelihood objective, $\log P_{\text{aug}}(x) = \log P_{\text{prior}}(x) + \sigma\,S(x)$, which keeps the agent anchored to drug-like chemical space while up-weighting high-scoring molecules. Because the target enters only through the scalar reward, arbitrary objectives --- including off-target docking scores --- can be optimized directly.

\textbf{ACEGEN} \citep{bou2024acegen} is a RL library for generative chemistry built on TorchRL, providing the same SMILES-policy family under a range of RL algorithms and network backbones. It follows the REINVENT formulation with a broader and more reproducible optimization setup.

For both methods, sampled molecules are docked into the on- and off-target pockets and the reward combines the two scores, so that the selectivity gap $\Delta$ enters the fitness function explicitly.

\section{Generation protocol}
\label{app:generation_protocol}
\subsection{Ligand generation and filtering} For every (method, target) pair across all three evaluation tracks (\taskA, \taskB, \taskC), we run each baseline's published sampling code at its default hyperparameters, except where noted below, and accumulate batches until 100 valid, connected molecules are collected. Validity is enforced by a uniform wrapper-level filter applied identically to every method: a molecule is retained if it (i) loads without error under RDKit's default sanitization (\texttt{Chem.SanitizeMol}) and (ii) yields a single-component canonical SMILES. The only validity criterion that varies across methods is each method's own internal reconstruction logic, which we leave unmodified.

We do not pre-screen for force-field (UFF/MMFF) parameter coverage nor for docking success. Molecules that pass the validity filter are sent to AutoDock Vina, and rows where Meeko/RDKit ligand preparation raises an error are reported as missing scores rather than re-generated, imputed, or back-filled. This choice preserves each method's emitted molecule set without further selection, at the cost of an uneven pre-docking failure rate across methods, which we surface below.

\subsection{Empirical pre-docking failure rates} Table~\ref{tab:prep_failure} aggregates per-method failure rates across all targets in each track. We accept the resulting
cross-method asymmetry: a method-aware filter would deviate from each method's published reconstruction behaviour and effectively introduce method-specific selection.

\begin{table}[h]
  \centering
  \mycaption{Pre-docking failure rates by method and evaluation track.}{Failures occur when a wrapper-validated molecule is rejected by Meeko/RDKit ligand preparation before AutoDock Vina runs. Aggregated as $1 - (N_\text{docked} / N_\text{generated})$ over all (method, target) cells in each track. \taskB totals are computed over the full 85-pair panel.}
  \label{tab:prep_failure}
  \begin{tabular}{lccc}
      \toprule
      Method      & \taskA & \taskB & \taskC \\
      \midrule
      DecompDiff  & 2.4\% & 3.5\% & 3.1\% \\
      DiffSBDD    & 1.6\% & 1.4\% & 5.2\% \\
      TargetDiff  & 0.6\% & 0.7\% & 2.3\% \\
      MolSnapper  & 2.5\% & 5.4\% & 6.7\% \\
      MolForm     & 0.6\% & 2.2\% & 9.2\% \\
      Pocket2Mol  & 0.6\% & 0.4\% & 0.4\% \\
      PocketFlow  & 0.2\% & 0.2\% & 0.4\% \\
      CMDGEN      & 0.2\% & 1.4\% & 0.3\% \\
      SpecLig     & 2.2\% & 2.6\% & 2.1\% \\
      \bottomrule
  \end{tabular}
\end{table}

\paragraph{DecompDiff configuration} DecompDiff's released sampler applies two hand-crafted geometric drifts during sampling: one constraining the nearest arm--scaffold distance to $[1.2, 1.9]$~\AA, and one penalizing ligand--protein body intersection. For conditional generations on \texttt{PARP1} and \texttt{DRD2} specifically, the published drifts collapse the sampler to near-zero usable yield (13/100 and 4/100, respectively). With the drifts disabled, these recover to 100/100 and 88/100 dockable molecules respectively, which are therefore evaluated in our study. 
For the rest of \taskA, and for \taskB and \taskC, the published configuration produces sufficient yield and we leave it unmodified. We additionally disable these custom guidance where evaluating our DecompDiff+CFG and DecompDiff+FKC selectivity-guided variants, where they would otherwise interact with the on/off-target guidance signal we are measuring.

\section{Full suite of performance metrics}
\subsection{Performance on \taskA}
\label{app:cs_result_app}
{\tiny
\setlength{\tabcolsep}{3.0pt}
\renewcommand{\arraystretch}{1.00}
\begin{longtable}{lcccccccccc}
\mycaption{\taskA  evaluation.} {For each method, we report: the mean on-target Vina score $\bar{s}_{\text{on}}$ ($\downarrow$); $\bar{\Delta}$ over valid molecules;  the mean of the $k$ largest $\Delta^{(i)}$ ($k = 5, 10$) $\bar{\Delta}_{\text{top-}k}$; the selectivity ratio $\mathrm{SR}_{\tau}$ for $\tau \in \{0, 1\}$ kcal/mol. Additionally, we report the z-score normalized analogues of the corresponding raw metrics, standardized against a reference score distribution: $\bar{s}_{\text{on}}^{z}$, $\bar{\Delta}^{z}$, $\mathrm{SR}^{z}_{\tau=0}$, $\mathrm{SR}^{z}_{\tau=1}$. The $\tau=1$ cutoff in z-space corresponds to $\Delta^{z} \geq 1$ (one reference-$\sigma$ in the off-target's own scale).}
\label{tab:case_study} \\

\toprule
Method & 
\makecell{$\bar{s}_{\text{on}}$\\$(\downarrow)$} & \makecell{$\bar{\Delta}$\\$(\uparrow)$} & \makecell{$\bar{\Delta}_{\text{top-}5}$\\$(\uparrow)$} & \makecell{$\bar{\Delta}_{\text{top-}10}$\\$(\uparrow)$} & 
\makecell{$\mathrm{SR}_{\tau=0}$\\$(\uparrow)$} & \makecell{$\mathrm{SR}_{\tau=1}$\\$(\uparrow)$}  & \makecell{$\bar{s}_{\text{on}}^{z}$\\$(\downarrow)$} & 
\makecell{$\bar{\Delta}^{z}$\\$(\uparrow)$} & \makecell{$\mathrm{SR}^{z}_{\tau=0}$\\$(\uparrow)$} & \makecell{$\mathrm{SR}^{z}_{\tau=1}$\\$(\uparrow)$} \\
\midrule
\endfirsthead

\toprule
Method & 
\makecell{$\bar{s}_{\text{on}}$\\$(\downarrow)$} & \makecell{$\bar{\Delta}$\\$(\uparrow)$} & \makecell{$\bar{\Delta}_{\text{top-}5}$\\$(\uparrow)$} & \makecell{$\bar{\Delta}_{\text{top-}10}$\\$(\uparrow)$} & 
\makecell{$\mathrm{SR}_{\tau=0}$\\$(\uparrow)$} & \makecell{$\mathrm{SR}_{\tau=1}$\\$(\uparrow)$}  & \makecell{$\bar{s}_{\text{on}}^{z}$\\$(\downarrow)$} & 
\makecell{$\bar{\Delta}^{z}$\\$(\uparrow)$} & \makecell{$\mathrm{SR}^{z}_{\tau=0}$\\$(\uparrow)$} & \makecell{$\mathrm{SR}^{z}_{\tau=1}$\\$(\uparrow)$} \\
\midrule
\endhead

\midrule
\multicolumn{11}{r}{Continued on next page}
\endfoot

\bottomrule
\endlastfoot

\multicolumn{11}{l}{\hlcustom{Blue1}{\texttt{Case: PARP1 (on) / PARP2 (off)}}} \\
\midrule
\multicolumn{11}{l}{\textit{Standard SBDD models}} \\
\midrule
DecompDiff & $-7.43_{\scriptscriptstyle\pm0.57}$ & $-0.03_{\scriptscriptstyle\pm0.42}$ & $1.05$ & $0.84$ & $0.41$ & $0.02$ & $2.15_{\scriptscriptstyle\pm0.60}$ & $0.02_{\scriptscriptstyle\pm0.41}$ & $0.46$ & $0.04$ \\
DiffSBDD & $-8.44_{\scriptscriptstyle\pm1.67}$ & $-0.05_{\scriptscriptstyle\pm0.43}$ & $0.90$ & $0.77$ & $0.46$ & $0.01$ & $1.09_{\scriptscriptstyle\pm1.76}$ & $0.08_{\scriptscriptstyle\pm0.43}$ & $0.57$ & $0.03$ \\
TargetDiff & $-8.56_{\scriptscriptstyle\pm1.22}$ & $0.07_{\scriptscriptstyle\pm0.54}$ & $1.52$ & $1.16$ & $0.55$ & $0.05$ & $0.96_{\scriptscriptstyle\pm1.28}$ & $0.20_{\scriptscriptstyle\pm0.53}$ & $0.67$ & $0.07$ \\
MolForm & $-8.30_{\scriptscriptstyle\pm1.23}$ & $-0.27_{\scriptscriptstyle\pm0.43}$ & $0.59$ & $0.45$ & $0.29$ & $0.00$ & $1.24_{\scriptscriptstyle\pm1.29}$ & $-0.15_{\scriptscriptstyle\pm0.44}$ & $0.37$ & $0.00$ \\
MolSnapper & $-8.87_{\scriptscriptstyle\pm1.57}$ & $-0.05_{\scriptscriptstyle\pm0.44}$ & $0.68$ & $0.61$ & $0.49$ & $0.00$ & $0.63_{\scriptscriptstyle\pm1.65}$ & $0.10_{\scriptscriptstyle\pm0.45}$ & $0.65$ & $0.01$ \\
Pocket2Mol & $-6.97_{\scriptscriptstyle\pm0.71}$ & $-0.33_{\scriptscriptstyle\pm0.33}$ & $0.39$ & $0.30$ & $0.16$ & $0.00$ & $2.63_{\scriptscriptstyle\pm0.75}$ & $-0.31_{\scriptscriptstyle\pm0.31}$ & $0.17$ & $0.00$ \\
\bottomrule

\end{longtable}}

We additionally report RL-based fine-tuning method results. They find molecules that bind the on-target tightly, matching or beating every structure-based generator in all five cases and falling behind only the two models trained for selectivity. What does not carry over is the selectivity itself. Even though the off-target score is part of the reward, the normalized gap $\bar{\Delta}^{z}$ stays close to zero everywhere, which suggests that the positive raw gaps on DRD2/HRH1 and BMP2K/TTK come mostly from the off-target simply being easier to score well against, not from the model having learned to tell the two pockets apart. The pattern seems to suggest they converge on a narrow set of high-affinity molecules and rarely venture outside it, so they produce few of the outliers that a selectivity screen actually cares about. 

Complementary to our docking-based evaluation, we report a selectivity assessment of all methods in \cref{sec:model-summary} using Boltz2 as scoring function in \cref{tab:case_study_boltz2}. A major difference introduced by that change is that higher values of $s_{on}$ are positively perceived ($\uparrow$). Additional assessment of selectivity is conducted in terms of threshold-based SR over predicted binding probabilities ($\mathrm{SR}_{\tau_{\mathrm{on}}=0.6, \;\tau_{\mathrm{off}}=0.4}$) and percentile-normalized ranks ($\mathrm{SR}^{\mathrm{rank}}_{\tau_{\mathrm{on}}=0.6, \;\tau_{\mathrm{off}}=0.4}$).

{\tiny
\setlength{\tabcolsep}{3.0pt}
\renewcommand{\arraystretch}{1.00}
\begin{longtable}{lccccccc}
\mycaption{\taskA  Boltz-2 evaluation.}{For each method, we report: the mean on-target predicted probability score $\bar{s}_{\text{on}}$ ($\uparrow$); $\bar{\Delta}$ over valid molecules;  the mean of the $k$ largest $\Delta^{(i)}$ ($k = 5, 10$) $\bar{\Delta}_{\text{top-}k}$; the selectivity ratio $\mathrm{SR}^{\mathrm{rank}}_{\tau = 0}$ over percentile-normalized ranks as well as the threshold-based SR over predicted binding probabilities $\mathrm{SR}_{\tau_{\mathrm{on}}=0.6, \;\tau_{\mathrm{off}}=0.4}$ and percentile-normalized ranks $\mathrm{SR}^{\mathrm{rank}}_{\tau_{\mathrm{on}}=0.6, \;\tau_{\mathrm{off}}=0.4}$}
\label{tab:case_study_boltz2} \\
\toprule
Method &
\makecell{$\bar{s}_{\text{on}}$\\$(\uparrow)$} &
\makecell{$\bar{\Delta}$\\$(\uparrow)$} &
\makecell{$\bar{\Delta}_{\text{top-}5}$\\$(\uparrow)$} &
\makecell{$\bar{\Delta}_{\text{top-}10}$\\$(\uparrow)$} &
\makecell{$\mathrm{SR}^{\mathrm{rank}}_{\tau=0}$\\$(\uparrow)$} &
\makecell{$\mathrm{SR}_{s_{\text{on}}>0.6, s_{\text{off}}<0.4}$\\$(\uparrow)$} &
\makecell{$\mathrm{SR}^{\mathrm{rank}}_{p_{\text{on}}>0.6, p_{\text{off}}<0.4}$\\$(\uparrow)$}   \\
\midrule
\endfirsthead
\toprule
Method &
\makecell{$\bar{s}_{\text{on}}$\\$(\uparrow)$} &
\makecell{$\bar{\Delta}$\\$(\uparrow)$} &
\makecell{$\bar{\Delta}_{\text{top-}5}$\\$(\uparrow)$} &
\makecell{$\bar{\Delta}_{\text{top-}10}$\\$(\uparrow)$} &
\makecell{$\mathrm{SR}^{\mathrm{rank}}_{\tau=0}$\\$(\uparrow)$} &
\makecell{$\mathrm{SR}_{s_{\text{on}}>0.6, s_{\text{off}}<0.4}$\\$(\uparrow)$} &
\makecell{$\mathrm{SR}^{\mathrm{rank}}_{p_{\text{on}}>0.6, p_{\text{off}}<0.4}$\\$(\uparrow)$}   \\
\midrule
\endhead
\multicolumn{8}{r}{Continued on next page} \\
\endfoot
\bottomrule
\endlastfoot
\multicolumn{8}{l}{\hlcustom{Blue1}{\texttt{Case: PARP1 (on) / PARP2 (off)}}} \\
\midrule
\multicolumn{8}{l}{\textit{Standard SBDD models}} \\
\midrule
DecompDiff & $0.396_{\pm 0.126}$ & $0.085_{\pm 0.083}$ & $0.257$ & $0.241$ & $0.490$ & $0.010$ & $0.080$ \\
DiffSBDD & $0.421_{\pm 0.214}$ & $0.050_{\pm 0.094}$ & $0.300$ & $0.253$ & $0.444$ & $0.010$ & $0.020$ \\
TargetDiff & $0.372_{\pm 0.158}$ & $0.069_{\pm 0.088}$ & $0.284$ & $0.241$ & $0.500$ & $0.010$ & $0.030$ \\
MolForm & $0.323_{\pm 0.162}$ & $0.038_{\pm 0.070}$ & $0.174$ & $0.163$ & $0.500$ & $0.000$ & $0.010$ \\
MolSnapper & $0.327_{\pm 0.125}$ & $-0.004_{\pm 0.082}$ & $0.161$ & $0.138$ & $0.450$ & $0.000$ & $0.030$ \\
Pocket2Mol & $0.330_{\pm 0.110}$ & $-0.019_{\pm 0.095}$ & $0.126$ & $0.108$ & $0.520$ & $0.000$ & $0.010$ \\
\end{longtable}}

\subsection{Performance on \taskBsmall}
\label{app:safety_result_priority_app}

{\tiny
\setlength{\tabcolsep}{0.5pt}
\renewcommand{\arraystretch}{1.00}
\begin{longtable}{lcccccccccc}
\mycaption{\taskBsmall evaluation}{For each method, we report: the mean on-target Vina score $\bar{s}_{\text{on}}$ ($\downarrow$); the worst per-molecule $\bar{\Delta}_{\text{worst-mol}}$ averaged across $T=22$ off-targets (least Selectivity-trained models); the fraction of molecules $\mathrm{SR}_{\tau=0}^{\text{worst-mol}}$ / $\mathrm{SR}_{\tau=1}^{\text{worst-mol}}$ whose worst per-target $\Delta$ is $\geq 0$ / $\geq 1$, respectively; the fraction $\mathrm{R}_{20}$ of ligands whose on-target ranks in the top-20\% across all 23 targets. Additionally, we report the z-score normalized analogues of the corresponding raw metrics, standardized against a reference score distribution: $\bar{s}_{\text{on}}^{z}$, $\bar{\Delta}^{z}$,$\mathrm{SR}_{\tau=0}^{z,\text{avg}}$, $\mathrm{SR}_{\tau=1}^{z,\text{avg}}$, $\mathrm{R}_{20}^{z}$. $\mathrm{SR}_{\tau}^{z,\text{avg}}$ is the panel-average of per-target z-SRs. $\tau=1$ in z-space corresponds to one reference-$\sigma$ margin in the off-target's own scale}

\label{tab:case_study} \\

\toprule
Method &
\makecell{$\bar{s}_{\text{on}}$\\$(\downarrow)$} &
\makecell{$\bar{\Delta}_{\text{worst-mol}}$\\$(\uparrow)$} &
\makecell{$\mathrm{SR}_{\tau=0}^{\text{worst-mol}}$\\$(\uparrow)$} &
\makecell{$\mathrm{SR}_{\tau=1}^{\text{worst-mol}}$\\$(\uparrow)$} &
\makecell{$\mathrm{R}_{20}$\\$(\uparrow)$} &
\makecell{$\bar{s}_{\text{on}}^{z}$\\$(\downarrow)$} &
\makecell{$\bar{\Delta}^{z}$\\$(\uparrow)$} &
\makecell{$\mathrm{SR}_{\tau=0}^{z,\text{avg}}$\\$(\uparrow)$} &
\makecell{$\mathrm{SR}_{\tau=1}^{z,\text{avg}}$\\$(\uparrow)$} &
\makecell{$\mathrm{R}_{20}^{\mathrm{z}}$\\$(\uparrow)$} \\
\midrule
\endfirsthead

\toprule
Method &
\makecell{$\bar{s}_{\text{on}}\,(\downarrow)$} &
\makecell{$\bar{\Delta}_{\text{worst-mol}}\,(\uparrow)$} &
\makecell{$\mathrm{SR}_{\tau=0}^{\text{worst-mol}}\,(\uparrow)$} &
\makecell{$\mathrm{SR}_{\tau=1}^{\text{worst-mol}}\,(\uparrow)$} &
\makecell{$\mathrm{R}_{20}\,(\uparrow)$} &
\makecell{$\bar{s}_{\text{on}}^{z}\,(\downarrow)$} &
\makecell{$\bar{\Delta}^{z}\,(\uparrow)$} &
\makecell{$\mathrm{SR}_{\tau=0}^{z,\text{avg}}\,(\uparrow)$} &
\makecell{$\mathrm{SR}_{\tau=1}^{z,\text{avg}}\,(\uparrow)$} &
\makecell{$\mathrm{R}_{20}^{z}\,(\uparrow)$} \\
\midrule
\endhead

\midrule
\multicolumn{11}{r}{Continued on next page}
\endfoot

\bottomrule
\endlastfoot

\multicolumn{11}{l}{\hlcustom{Blue1}{\texttt{POI: 5HT1B (priority)}}} \\
\midrule
\multicolumn{11}{l}{\textit{Standard SBDD models}} \\
\midrule
DecompDiff &
$-6.73_{\scriptscriptstyle\pm1.69}$ &
$-1.16_{\scriptscriptstyle\pm1.71}$ &
$0.24$ &
$0.02$ &
$0.64$ &
$2.45_{\scriptscriptstyle\pm1.75}$ &
$0.74_{\scriptscriptstyle\pm1.85}$ &
$0.68$ &
$0.44$ &
$0.43$ \\
DiffSBDD &
$-6.83_{\scriptscriptstyle\pm1.76}$ &
$-1.32_{\scriptscriptstyle\pm1.05}$ &
$0.07$ &
$0.00$ &
$0.38$ &
$2.34_{\scriptscriptstyle\pm1.82}$ &
$-0.08_{\scriptscriptstyle\pm1.18}$ &
$0.51$ &
$0.21$ &
$0.21$ \\
TargetDiff &
$-7.21_{\scriptscriptstyle\pm1.53}$ &
$-1.20_{\scriptscriptstyle\pm1.42}$ &
$0.19$ &
$0.04$ &
$0.50$ &
$1.95_{\scriptscriptstyle\pm1.59}$ &
$1.04_{\scriptscriptstyle\pm1.77}$ &
$0.65$ &
$0.41$ &
$0.35$ \\
MolForm &
$-7.00_{\scriptscriptstyle\pm2.64}$ &
$-2.00_{\scriptscriptstyle\pm2.27}$ &
$0.09$ &
$0.02$ &
$0.45$ &
$2.17_{\scriptscriptstyle\pm2.73}$ &
$0.72_{\scriptscriptstyle\pm2.01}$ &
$0.63$ &
$0.40$ &
$0.34$ \\
MolSnapper &
$-7.04_{\scriptscriptstyle\pm1.13}$ &
$-1.43_{\scriptscriptstyle\pm1.00}$ &
$0.03$ &
$0.00$ &
$0.17$ &
$2.13_{\scriptscriptstyle\pm1.17}$ &
$-0.41_{\scriptscriptstyle\pm0.94}$ &
$0.41$ &
$0.10$ &
$0.02$ \\
Pocket2Mol &
$-6.33_{\scriptscriptstyle\pm1.37}$ &
$-1.26_{\scriptscriptstyle\pm0.54}$ &
$0.00$ &
$0.00$ &
$0.01$ &
$2.86_{\scriptscriptstyle\pm1.42}$ &
$-0.76_{\scriptscriptstyle\pm0.48}$ &
$0.31$ &
$0.01$ &
$0.00$ \\
\bottomrule
\end{longtable}}

\clearpage
\subsection{Performance on \taskBbig}
\label{app:safety_result_all}

To facilitate analysis, we visualize performance across two metrics: $R_{20z}$ and Avg $\text{SR}^{1\text{kcal}}_z$ in \cref{fig:result-plot-safety-priority}, evaluated against each POI and 84 safety targets (HTR1B is excluded, cf. \cref{fig:probis_sb_safety}); further metrics are reported in \cref{tab:safety_per_target_all}. Overall, similar trends as in \cref{sec:safety_result} and \cref{app:safety_result_priority_app} can be observed.

\begin{figure}[h]
    \centering
    \includegraphics[width=\linewidth]{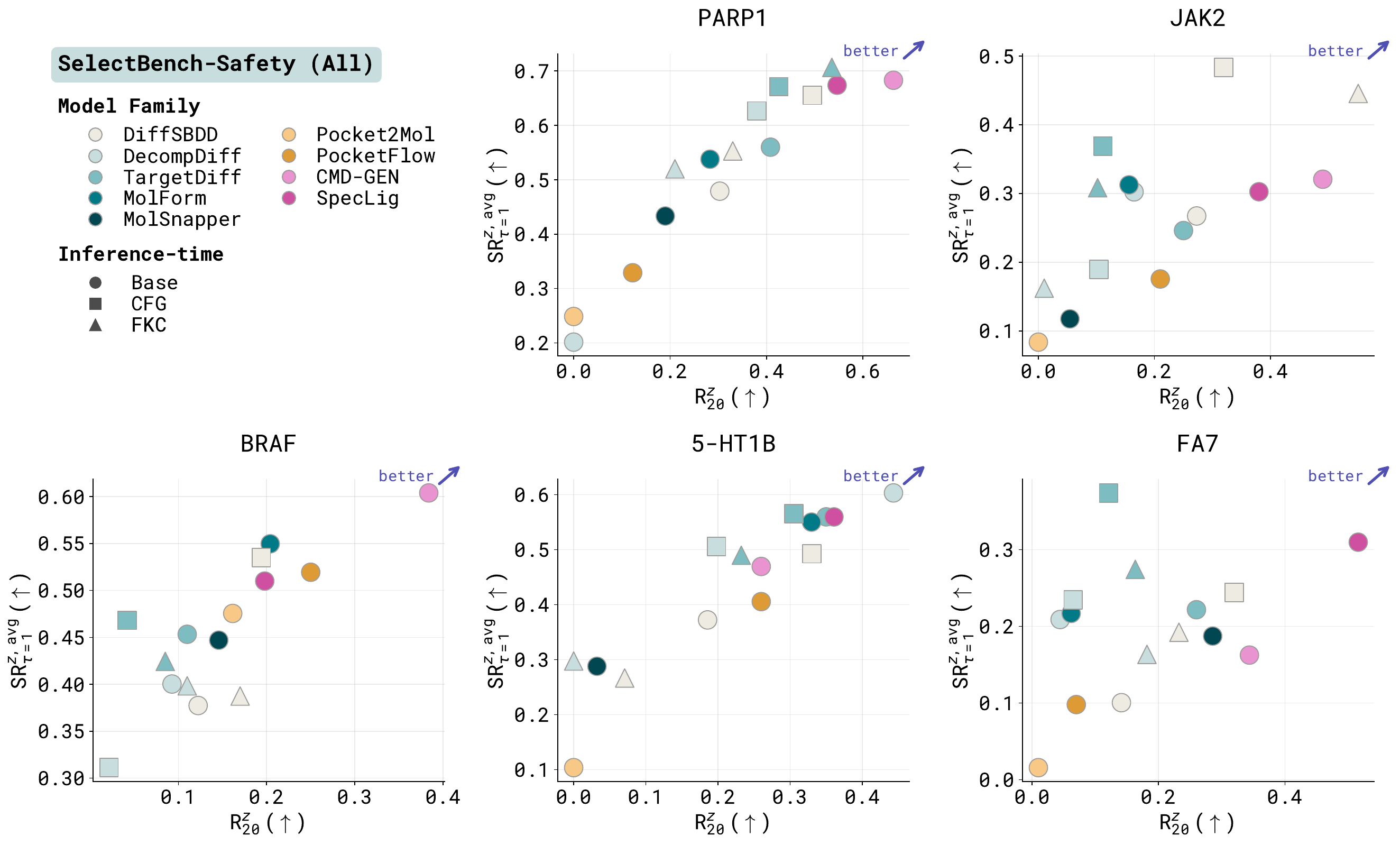}
    \mycaption{Baseline performances on \taskBbig}{The mean performance is reported across $\mathrm{R}_{20}^{\mathrm{z}}\,(\uparrow)$  and $\mathrm{SR}_{\tau=1}^{\text{avg}}\,(\uparrow)$ for each baseline on each on-target POI/all off-targets case in \taskBbig. The color of the marker indicates the model type, and its shape indicates if inference-time methods were additionally employed. Data points in the top-right region indicate better performance.}
    \label{fig:result-plot-safety-all}
\end{figure}

{\tiny
\setlength{\tabcolsep}{0.5pt}
\renewcommand{\arraystretch}{1.00}
\begin{longtable}{lcccccccccc}
\mycaption{\taskBbig evaluation.}{For each method, we report: the mean on-target Vina score $\bar{s}_{\text{on}}$ ($\downarrow$); the worst per-molecule $\bar{\Delta}_{\text{worst-mol}}$ averaged across $T=22$ off-targets (least Selectivity-trained models); the fraction of molecules $\mathrm{SR}_{\tau=0}^{\text{worst-mol}}$ / $\mathrm{SR}_{\tau=1}^{\text{worst-mol}}$ whose worst per-target $\Delta$ is $\geq 0$ / $\geq 1$, respectively; the fraction $\mathrm{R}_{20}$ of ligands whose on-target ranks in the top-20\% across all 23 targets. Additionally, we report the z-score normalized analogues of the corresponding raw metrics, standardized against a reference score distribution: $\bar{s}_{\text{on}}^{z}$, $\bar{\Delta}^{z}$,$\mathrm{SR}_{\tau=0}^{z,\text{avg}}$, $\mathrm{SR}_{\tau=1}^{z,\text{avg}}$, $\mathrm{R}_{20}^{z}$. $\mathrm{SR}_{\tau}^{z,\text{avg}}$ is the panel-average of per-target z-SRs. $\tau=1$ in z-space corresponds to one reference-$\sigma$ margin in the off-target's own scale}
\label{tab:safety_per_target_all} \\

\toprule
Method &
\makecell{$\bar{s}_{\text{on}}$\\$(\downarrow)$} &
\makecell{$\bar{\Delta}_{\text{worst-mol}}$\\$(\uparrow)$} &
\makecell{$\mathrm{SR}_{\tau=0}^{\text{worst-mol}}$\\$(\uparrow)$} &
\makecell{$\mathrm{SR}_{\tau=1}^{\text{worst-mol}}$\\$(\uparrow)$} &
\makecell{$\mathrm{R}_{20}$\\$(\uparrow)$} &
\makecell{$\bar{s}_{\text{on}}^{z}$\\$(\downarrow)$} &
\makecell{$\bar{\Delta}^{z}$\\$(\uparrow)$} &
\makecell{$\mathrm{SR}_{\tau=0}^{z,\text{avg}}$\\$(\uparrow)$} &
\makecell{$\mathrm{SR}_{\tau=1}^{z,\text{avg}}$\\$(\uparrow)$} &
\makecell{$\mathrm{R}_{20}^{\mathrm{z}}$\\$(\uparrow)$} \\
\midrule
\endfirsthead

\toprule
Method &
\makecell{$\bar{s}_{\text{on}}\,(\downarrow)$} &
\makecell{$\bar{\Delta}_{\text{worst-mol}}\,(\uparrow)$} &
\makecell{$\mathrm{SR}_{\tau=0}^{\text{worst-mol}}\,(\uparrow)$} &
\makecell{$\mathrm{SR}_{\tau=1}^{\text{worst-mol}}\,(\uparrow)$} &
\makecell{$\mathrm{R}_{20}\,(\uparrow)$} &
\makecell{$\bar{s}_{\text{on}}^{z}\,(\downarrow)$} &
\makecell{$\bar{\Delta}^{z}\,(\uparrow)$} &
\makecell{$\mathrm{SR}_{\tau=0}^{z,\text{avg}}\,(\uparrow)$} &
\makecell{$\mathrm{SR}_{\tau=1}^{z,\text{avg}}\,(\uparrow)$} &
\makecell{$\mathrm{R}_{20}^{z}\,(\uparrow)$} \\
\midrule
\endhead

\midrule
\multicolumn{11}{r}{Continued on next page}
\endfoot

\bottomrule
\endlastfoot

\multicolumn{11}{l}{\hlcustom{Blue1}{\texttt{POI: 5HT1B}}} \\
\midrule
\multicolumn{11}{l}{\textit{Standard SBDD models}} \\
\midrule
DecompDiff &
$-6.73_{\scriptscriptstyle\pm1.69}$ &
$-1.60_{\scriptscriptstyle\pm1.68}$ &
$0.11$ &
$0.01$ &
$0.70$ &
$2.45_{\scriptscriptstyle\pm1.75}$ &
$0.61_{\scriptscriptstyle\pm1.79}$ &
$0.66$ &
$0.41$ &
$0.44$ \\
DiffSBDD &
$-6.83_{\scriptscriptstyle\pm1.76}$ &
$-1.57_{\scriptscriptstyle\pm0.99}$ &
$0.00$ &
$0.00$ &
$0.44$ &
$2.34_{\scriptscriptstyle\pm1.82}$ &
$-0.13_{\scriptscriptstyle\pm1.12}$ &
$0.50$ &
$0.19$ &
$0.19$ \\
TargetDiff &
$-7.21_{\scriptscriptstyle\pm1.53}$ &
$-1.54_{\scriptscriptstyle\pm1.32}$ &
$0.06$ &
$0.01$ &
$0.62$ &
$1.95_{\scriptscriptstyle\pm1.59}$ &
$0.79_{\scriptscriptstyle\pm1.56}$ &
$0.64$ &
$0.37$ &
$0.35$ \\
MolForm &
$-7.00_{\scriptscriptstyle\pm2.64}$ &
$-2.35_{\scriptscriptstyle\pm2.20}$ &
$0.02$ &
$0.00$ &
$0.55$ &
$2.17_{\scriptscriptstyle\pm2.73}$ &
$0.56_{\scriptscriptstyle\pm1.92}$ &
$0.61$ &
$0.38$ &
$0.33$ \\
MolSnapper &
$-7.04_{\scriptscriptstyle\pm1.13}$ &
$-1.53_{\scriptscriptstyle\pm1.03}$ &
$0.00$ &
$0.00$ &
$0.43$ &
$2.13_{\scriptscriptstyle\pm1.17}$ &
$-0.39_{\scriptscriptstyle\pm0.89}$ &
$0.44$ &
$0.10$ &
$0.03$ \\
Pocket2Mol &
$-6.33_{\scriptscriptstyle\pm1.37}$ &
$-1.42_{\scriptscriptstyle\pm0.56}$ &
$0.00$ &
$0.00$ &
$0.08$ &
$2.86_{\scriptscriptstyle\pm1.42}$ &
$-0.79_{\scriptscriptstyle\pm0.46}$ &
$0.30$ &
$0.01$ &
$0.00$ \\
\bottomrule
\end{longtable}}

\subsection{Performance on \taskC}
\label{app:xl_result_app}

{\tiny
\setlength{\tabcolsep}{3.0pt}
\renewcommand{\arraystretch}{1.00}
\begin{longtable}{lcccccccccc}
\mycaption{\taskC evaluation.} { Metrics are defined as in the case-study table and computed per on-/off-target pair. Values are means over all pairs in the tier; where shown, $\pm$ is the standard deviation across molecules within a pair, averaged over pairs. $\bar{s}_{\text{on}}^{z}$, $\bar{\Delta}^{z}$, $\mathrm{SR}^{z}_{\tau=0}$, $\mathrm{SR}^{z}_{\tau=1}$: z-score variants computed against per-pocket reference distributions; $\tau=1$ in z-space corresponds to one reference-$\sigma$ margin.}
\label{tab:sb_xl_all} \\

\toprule
Method & 
\makecell{$\bar{s}_{\text{on}}$\\$(\downarrow)$} & \makecell{$\bar{\Delta}$\\$(\uparrow)$} & \makecell{$\bar{\Delta}_{\text{top-}5}$\\$(\uparrow)$} & \makecell{$\bar{\Delta}_{\text{top-}10}$\\$(\uparrow)$} & 
\makecell{$\mathrm{SR}_{\tau=0}$\\$(\uparrow)$} & \makecell{$\mathrm{SR}_{\tau=1}$\\$(\uparrow)$}  & \makecell{$\bar{s}_{\text{on}}^{z}$\\$(\downarrow)$} & 
\makecell{$\bar{\Delta}^{z}$\\$(\uparrow)$} & \makecell{$\mathrm{SR}^{z}_{\tau=0}$\\$(\uparrow)$} & \makecell{$\mathrm{SR}^{z}_{\tau=1}$\\$(\uparrow)$} \\
\midrule
\endfirsthead

\toprule
Method & 
\makecell{$\bar{s}_{\text{on}}$\\$(\downarrow)$} & \makecell{$\bar{\Delta}$\\$(\uparrow)$} & \makecell{$\bar{\Delta}_{\text{top-}5}$\\$(\uparrow)$} & \makecell{$\bar{\Delta}_{\text{top-}10}$\\$(\uparrow)$} & 
\makecell{$\mathrm{SR}_{\tau=0}$\\$(\uparrow)$} & \makecell{$\mathrm{SR}_{\tau=1}$\\$(\uparrow)$}  & \makecell{$\bar{s}_{\text{on}}^{z}$\\$(\downarrow)$} & 
\makecell{$\bar{\Delta}^{z}$\\$(\uparrow)$} & \makecell{$\mathrm{SR}^{z}_{\tau=0}$\\$(\uparrow)$} & \makecell{$\mathrm{SR}^{z}_{\tau=1}$\\$(\uparrow)$} \\
\midrule
\endhead

\midrule
\multicolumn{11}{r}{Continued on next page}
\endfoot

\bottomrule
\endlastfoot

\multicolumn{11}{l}{\hlcustom{Blue1}{\texttt{Hard}}} \\
\midrule
\multicolumn{11}{l}{\textit{Standard SBDD models}} \\
\midrule
DecompDiff & $-5.48_{\scriptscriptstyle\pm0.66}$ & $0.10_{\scriptscriptstyle\pm0.40}$ & $0.85$ & $0.74$ & $0.51$ & $0.09$ & $3.39_{\scriptscriptstyle\pm0.76}$ & $0.07_{\scriptscriptstyle\pm0.45}$ & $0.53$ & $0.12$ \\
DiffSBDD & $-5.98_{\scriptscriptstyle\pm1.16}$ & $0.08_{\scriptscriptstyle\pm0.46}$ & $1.12$ & $0.92$ & $0.54$ & $0.06$ & $2.76_{\scriptscriptstyle\pm1.37}$ & $0.07_{\scriptscriptstyle\pm0.53}$ & $0.53$ & $0.09$ \\
TargetDiff & $-6.42_{\scriptscriptstyle\pm1.15}$ & $0.12_{\scriptscriptstyle\pm0.57}$ & $1.27$ & $1.08$ & $0.55$ & $0.11$ & $2.24_{\scriptscriptstyle\pm1.32}$ & $0.10_{\scriptscriptstyle\pm0.62}$ & $0.55$ & $0.11$ \\
MolForm & $-6.75_{\scriptscriptstyle\pm1.30}$ & $0.10_{\scriptscriptstyle\pm0.52}$ & $1.13$ & $0.97$ & $0.53$ & $0.10$ & $1.89_{\scriptscriptstyle\pm1.51}$ & $0.10_{\scriptscriptstyle\pm0.58}$ & $0.53$ & $0.11$ \\
MolSnapper & $-7.75_{\scriptscriptstyle\pm1.14}$ & $0.08_{\scriptscriptstyle\pm0.58}$ & $1.33$ & $1.11$ & $0.53$ & $0.11$ & $0.62_{\scriptscriptstyle\pm1.35}$ & $0.10_{\scriptscriptstyle\pm0.68}$ & $0.53$ & $0.11$ \\
Pocket2Mol & $-5.83_{\scriptscriptstyle\pm0.75}$ & $0.04_{\scriptscriptstyle\pm0.28}$ & $0.61$ & $0.53$ & $0.52$ & $0.02$ & $2.99_{\scriptscriptstyle\pm0.87}$ & $0.01_{\scriptscriptstyle\pm0.33}$ & $0.50$ & $0.07$ \\
\bottomrule
\end{longtable}}

\section{Multi-objective metric for potency and selectivity}
\label{app:beyond_selectivity_metrics}

To capture the potency/selectivity trend investigated in \cref{sec:beyond_selectivity_result} we can introduce the potency-and-selectivity metric $\mathrm{PS} = - s_{\mathrm{on}} \cdot \Delta^{\mathrm{z}}$ ($\uparrow$), provided a negative $s_{\mathrm{on}}$.

\begin{table}[h]
\centering
\small
\mycaption{\taskA\ potency and selectivity evaluation.}{For each method, we report $\mathrm{PS}$; higher values indicate methods yielding jointly potent and selective molecules.}
\begin{tabular}{lccccc}
\toprule
Method & \hlcustom{Blue1}{\textbf{PARP1}} & \hlcustom{Pink1}{\textbf{BMP2K}} & \hlcustom{Yellow2}{\textbf{CDK9}} & \hlcustom{Purple1}{\textbf{ERK2}} & \hlcustom{Blue2}{\textbf{DRD2}} \\
\midrule
\multicolumn{6}{l}{\textit{Standard SBDD models}} \\
\midrule
DecompDiff & $0.14$ & $-3.63$ & $0.49$ & $0.74$ & $-8.27$ \\
DiffSBDD & $0.65$ & $1.13$ & $-0.05$ & $-0.16$ & $-6.87$ \\
TargetDiff & $1.74$ & $-0.16$ & $2.10$ & $0.50$ & $-4.11$ \\
MolForm & $-1.23$ & $-2.76$ & $1.04$ & $3.09$ & $-0.28$ \\
MolSnapper & $0.91$ & $0.82$ & $0.53$ & $1.25$ & $-8.65$ \\
Pocket2Mol & $-2.14$ & $1.11$ & $0.33$ & $-1.20$ & $-9.37$ \\
PocketFlow & $-0.32$ & $0.25$ & $0.52$ & $-3.12$ & $-1.45$ \\
\midrule
\multicolumn{6}{l}{\textit{Selectivity-trained models}} \\
\midrule
CMD-GEN & $-0.01$ & $1.50$ & $-0.14$ & $-3.30$ & $9.33$ \\
SpecLig & $0.32$ & $-0.68$ & $-0.66$ & $0.85$ & $0.89$ \\
\midrule
\multicolumn{6}{l}{\textit{Inference-time guided methods}} \\
\midrule
DecompDiff + CFG & $0.22$ & $-1.55$ & $1.87$ & $6.38$ & $-3.56$ \\
DecompDiff + FKC & $4.96$ & $1.52$ & $0.58$ & $4.35$ & $-0.86$ \\
DiffSBDD + CFG & $0.98$ & $-1.20$ & $3.56$ & $2.86$ & $1.60$ \\
DiffSBDD + FKC & $0.24$ & $-1.56$ & $2.58$ & $3.76$ & $-0.93$ \\
TargetDiff + CFG & $2.93$ & $-0.88$ & $4.25$ & $8.76$ & $7.75$ \\
TargetDiff + FKC & $10.09$ & $2.62$ & $1.63$ & $5.37$ & $5.11$ \\
\bottomrule
\end{tabular}
\end{table}

\begin{table}[h]
\centering
\small
\setlength{\tabcolsep}{6.0pt}
\renewcommand{\arraystretch}{1.05}
\mycaption{\taskB\ potency and selectivity evaluation.}{For each method, we report $\mathrm{PS}$; higher values indicate methods yielding jointly potent and selective molecules.}
\label{tab:safety_study}
\begin{tabular}{lccccc}
\toprule
Method & \hlcustom{Blue1}{\textbf{PARP1}} & \hlcustom{Pink1}{\textbf{JAK2}} & \hlcustom{Yellow2}{\textbf{BRAF}} & \hlcustom{Purple1}{\textbf{5-HT1B}} & \hlcustom{Blue2}{\textbf{FA7}} \\
\midrule
\multicolumn{6}{l}{\textit{Standard SBDD models}} \\
\midrule
DecompDiff & $-4.23$ & $0.91$ & $-1.86$ & $4.09$ & $-1.79$ \\
DiffSBDD & $2.81$ & $2.03$ & $-1.17$ & $-0.88$ & $-1.73$ \\
TargetDiff & $6.09$ & $1.01$ & $1.20$ & $5.70$ & $1.78$ \\
MolForm & $8.36$ & $3.25$ & $7.25$ & $3.89$ & $-1.41$ \\
MolSnapper & $1.87$ & $-1.45$ & $0.57$ & $-2.77$ & $3.32$ \\
Pocket2Mol & $-2.02$ & $-1.18$ & $1.16$ & $-5.02$ & $-3.86$ \\
PocketFlow & $0.23$ & $0.35$ & $5.04$ & $0.69$ & $-0.83$ \\
\midrule
\multicolumn{6}{l}{\textit{Selectivity-trained models}} \\
\midrule
CMD-GEN & $11.23$ & $7.84$ & $9.02$ & $-0.41$ & $3.82$ \\
SpecLig & $15.13$ & $5.05$ & $4.64$ & $5.60$ & $11.21$ \\
\midrule
\multicolumn{6}{l}{\textit{Inference-time guided methods}} \\
\midrule
DecompDiff + CFG & $4.38$ & $-0.46$ & $-3.87$ & $1.14$ & $0.54$ \\
DecompDiff + FKC & $2.48$ & $-0.80$ & $-1.30$ & $-5.11$ & $0.36$ \\
DiffSBDD + CFG & $15.72$ & $16.60$ & $7.07$ & $2.55$ & $5.02$ \\
DiffSBDD + FKC & $6.24$ & $13.27$ & $-0.16$ & $-5.51$ & $2.73$ \\
TargetDiff + CFG & $10.55$ & $-0.38$ & $1.36$ & $3.07$ & $4.03$ \\
TargetDiff + FKC & $10.66$ & $-1.68$ & $-1.16$ & $0.10$ & $1.31$ \\
\bottomrule
\end{tabular}
\end{table}

\begin{table}[h]
\centering
\small
\mycaption{\taskC\ potency and selectivity evaluation.}{For each method, we report $\mathrm{PS}$; higher values indicate methods yielding jointly potent and selective molecules.}
\begin{tabular}{lccc}
\toprule
Method & \hlcustom{Blue1}{\textbf{Easy}} & \hlcustom{Pink1}{\textbf{Medium}} & \hlcustom{Yellow2}{\textbf{Hard}} \\
\midrule
\multicolumn{4}{l}{\textit{Standard SBDD models}} \\
\midrule
DecompDiff & $1.98$ & $0.59$ & $0.38$ \\
DiffSBDD & $0.36$ & $0.48$ & $0.42$ \\
TargetDiff & $0.19$ & $1.14$ & $0.64$ \\
MolForm & $0.50$ & $1.20$ & $0.68$ \\
MolSnapper & $0.86$ & $0.07$ & $0.78$ \\
Pocket2Mol & $0.39$ & $0.27$ & $0.06$ \\
PocketFlow & $1.07$ & $0.86$ & $0.36$ \\
\midrule
\multicolumn{4}{l}{\textit{Selectivity-trained models}} \\
\midrule
CMD-GEN & $0.71$ & $0.35$ & $0.09$ \\
SpecLig & $-0.06$ & $0.70$ & $0.26$ \\
\midrule
\multicolumn{4}{l}{\textit{Inference-time guided methods}} \\
\midrule
DecompDiff + CFG & $1.31$ & $0.19$ & $0.58$ \\
DecompDiff + FKC & $1.28$ & $0.61$ & $1.44$ \\
DiffSBDD + CFG & $2.55$ & $4.76$ & $1.76$ \\
DiffSBDD + FKC & $2.40$ & $5.14$ & $1.86$ \\
TargetDiff + CFG & $1.34$ & $0.42$ & $0.89$ \\
TargetDiff + FKC & $1.38$ & $-0.07$ & $1.99$ \\
\bottomrule
\end{tabular}
\end{table}

\section{Additional molecular generation metrics}
\label{app:xtra_metrics}

In complement to selectivity metrics, we report the quality of generated molecules by each method using the following widely used metrics:

\begin{itemize}
    \item \textbf{QED}: the quantitative estimation of drug-likeness measures several desirable molecular properties of drug-like molecules \cite{bickerton2012quantifying}
    \item \textbf{SA}: the synthetic accessibility score estimates the difficulty of drug synthesis \cite{ertl2009estimation}. We adopt the normalized between 0 and 1 version of the score, with higher values indicating easier synthesis.
    \item \textbf{LogP}: the octanol-water partition coefficient is a measure of hydrophobicity \cite{wildman1999prediction}. In general logP values should be between -0.4 and 5.6 to be good drug candidates.
    \item \textbf{Lipinski}: measures how many rules in the Lipinski rule of five are satisfied by the generated molecules \cite{lipinski23pj}.
    \item \textbf{Diversity}: is estimated as the average pairwise dissimilarity ($1 - \text{Tanimoto similarity}$) between molecular fingerprints of all generated molecules for each on-target pocket \cite{schneuing2024structure}.
\end{itemize}

In addition, we report the average molecular weight (MW, in g/mol) of molecules generated by each method. 

\subsection{Molecular generation metrics on \taskA}
\label{app:xtra_metrics_cs}
{
\footnotesize
\renewcommand{\arraystretch}{1.00}
\begin{longtable}{lcccccc}
\mycaption{\taskB evaluation on additional molecular generation metrics.}{}
\label{tab:safety_xtra_metrics} \\

\toprule
Method & QED$\,(\uparrow)$ & SA$\,(\uparrow)$ & LogP & Lipinski$\,(\uparrow)$ & MW & Diversity$\,(\uparrow)$ \\
\midrule
\endfirsthead

\toprule
Method & QED$\,(\uparrow)$ & SA$\,(\uparrow)$ & LogP & Lipinski$\,(\uparrow)$ & MW & Diversity$\,(\uparrow)$ \\
\midrule
\endhead

\midrule
\multicolumn{7}{r}{Continued on next page}
\endfoot

\bottomrule
\endlastfoot

\multicolumn{7}{l}{\hlcustom{Blue1}{\texttt{PARP1 (on) / PARP2 (off)}}} \\
\midrule
\multicolumn{7}{l}{\textit{Standard SBDD models}} \\
\midrule
DecompDiff & $0.22_{\scriptscriptstyle\pm0.12}$ & $0.57_{\scriptscriptstyle\pm0.07}$ & $1.75_{\scriptscriptstyle\pm2.01}$ & $3.48_{\scriptscriptstyle\pm0.74}$ & $412.72_{\scriptscriptstyle\pm18.46}$ & $0.799$ \\
DiffSBDD & $0.39_{\scriptscriptstyle\pm0.21}$ & $0.58_{\scriptscriptstyle\pm0.12}$ & $1.29_{\scriptscriptstyle\pm1.94}$ & $4.16_{\scriptscriptstyle\pm1.20}$ & $415.26_{\scriptscriptstyle\pm134.94}$ & $0.677$ \\
TargetDiff & $0.25_{\scriptscriptstyle\pm0.17}$ & $0.50_{\scriptscriptstyle\pm0.09}$ & $2.05_{\scriptscriptstyle\pm1.93}$ & $3.42_{\scriptscriptstyle\pm1.39}$ & $495.64_{\scriptscriptstyle\pm94.75}$ & $0.667$ \\
MolForm & $0.23_{\scriptscriptstyle\pm0.10}$ & $0.56_{\scriptscriptstyle\pm0.11}$ & $9.00_{\scriptscriptstyle\pm1.43}$ & $2.98_{\scriptscriptstyle\pm0.73}$ & $483.83_{\scriptscriptstyle\pm100.48}$ & $0.728$ \\
MolSnapper & $0.41_{\scriptscriptstyle\pm0.15}$ & $0.59_{\scriptscriptstyle\pm0.17}$ & $0.19_{\scriptscriptstyle\pm1.28}$ & $4.85_{\scriptscriptstyle\pm0.36}$ & $314.78_{\scriptscriptstyle\pm86.56}$ & $0.746$ \\
Pocket2Mol & $0.66_{\scriptscriptstyle\pm0.10}$ & $0.79_{\scriptscriptstyle\pm0.09}$ & $1.01_{\scriptscriptstyle\pm1.10}$ & $5.00_{\scriptscriptstyle\pm0.00}$ & $196.58_{\scriptscriptstyle\pm20.67}$ & $0.830$ \\
PocketFlow & $0.44_{\scriptscriptstyle\pm0.25}$ & $0.76_{\scriptscriptstyle\pm0.12}$ & $4.04_{\scriptscriptstyle\pm2.26}$ & $4.07_{\scriptscriptstyle\pm1.14}$ & $354.28_{\scriptscriptstyle\pm166.92}$ & $0.815$ \\
\midrule
\multicolumn{7}{l}{\textit{Selectivity-trained models}} \\
\midrule
CMD-GEN & $0.63_{\scriptscriptstyle\pm0.10}$ & $0.66_{\scriptscriptstyle\pm0.05}$ & $3.91_{\scriptscriptstyle\pm0.48}$ & $4.98_{\scriptscriptstyle\pm0.14}$ & $402.90_{\scriptscriptstyle\pm18.59}$ & $0.604$ \\
SpecLig & $0.35_{\scriptscriptstyle\pm0.14}$ & $0.67_{\scriptscriptstyle\pm0.10}$ & $3.98_{\scriptscriptstyle\pm2.46}$ & $3.71_{\scriptscriptstyle\pm0.95}$ & $524.30_{\scriptscriptstyle\pm50.51}$ & $0.554$ \\
\midrule
\multicolumn{7}{l}{\textit{Inference-time guided methods}} \\
\midrule
DecompDiff + CFG & $0.29_{\scriptscriptstyle\pm0.13}$ & $0.45_{\scriptscriptstyle\pm0.08}$ & $1.31_{\scriptscriptstyle\pm2.11}$ & $3.27_{\scriptscriptstyle\pm1.26}$ & $525.31_{\scriptscriptstyle\pm20.40}$ & $0.574$ \\
DecompDiff + FKC & $0.22_{\scriptscriptstyle\pm0.09}$ & $0.35_{\scriptscriptstyle\pm0.07}$ & $-0.88_{\scriptscriptstyle\pm2.40}$ & $2.75_{\scriptscriptstyle\pm1.07}$ & $533.29_{\scriptscriptstyle\pm20.00}$ & $0.452$ \\
DiffSBDD + CFG & $0.55_{\scriptscriptstyle\pm0.21}$ & $0.65_{\scriptscriptstyle\pm0.11}$ & $1.89_{\scriptscriptstyle\pm1.76}$ & $4.74_{\scriptscriptstyle\pm0.59}$ & $361.30_{\scriptscriptstyle\pm85.94}$ & $0.733$ \\
DiffSBDD + FKC & $0.60_{\scriptscriptstyle\pm0.17}$ & $0.67_{\scriptscriptstyle\pm0.11}$ & $1.69_{\scriptscriptstyle\pm1.62}$ & $4.89_{\scriptscriptstyle\pm0.34}$ & $351.95_{\scriptscriptstyle\pm49.79}$ & $0.738$ \\
TargetDiff + CFG & $0.29_{\scriptscriptstyle\pm0.15}$ & $0.48_{\scriptscriptstyle\pm0.09}$ & $2.66_{\scriptscriptstyle\pm1.87}$ & $3.52_{\scriptscriptstyle\pm1.15}$ & $524.93_{\scriptscriptstyle\pm79.13}$ & $0.653$ \\
TargetDiff + FKC & $0.22_{\scriptscriptstyle\pm0.11}$ & $0.47_{\scriptscriptstyle\pm0.10}$ & $2.80_{\scriptscriptstyle\pm2.19}$ & $3.26_{\scriptscriptstyle\pm1.04}$ & $568.56_{\scriptscriptstyle\pm93.30}$ & $0.640$ \\
\midrule
\multicolumn{7}{l}{\hlcustom{Pink1}{\texttt{BMP2K (on) / TTK (off)}}} \\
\midrule
\multicolumn{7}{l}{\textit{Standard SBDD models}} \\
\midrule
DecompDiff & $0.34_{\scriptscriptstyle\pm0.14}$ & $0.49_{\scriptscriptstyle\pm0.10}$ & $2.58_{\scriptscriptstyle\pm1.88}$ & $4.30_{\scriptscriptstyle\pm0.84}$ & $491.86_{\scriptscriptstyle\pm15.28}$ & $0.573$ \\
DiffSBDD & $0.60_{\scriptscriptstyle\pm0.17}$ & $0.62_{\scriptscriptstyle\pm0.11}$ & $1.85_{\scriptscriptstyle\pm1.53}$ & $4.92_{\scriptscriptstyle\pm0.31}$ & $313.03_{\scriptscriptstyle\pm91.49}$ & $0.707$ \\
TargetDiff & $0.25_{\scriptscriptstyle\pm0.17}$ & $0.56_{\scriptscriptstyle\pm0.10}$ & $3.61_{\scriptscriptstyle\pm1.92}$ & $3.64_{\scriptscriptstyle\pm1.07}$ & $485.87_{\scriptscriptstyle\pm109.08}$ & $0.701$ \\
MolForm & $0.25_{\scriptscriptstyle\pm0.13}$ & $0.53_{\scriptscriptstyle\pm0.10}$ & $8.47_{\scriptscriptstyle\pm1.61}$ & $3.35_{\scriptscriptstyle\pm0.81}$ & $472.56_{\scriptscriptstyle\pm88.89}$ & $0.669$ \\
MolSnapper & $0.44_{\scriptscriptstyle\pm0.17}$ & $0.71_{\scriptscriptstyle\pm0.12}$ & $0.63_{\scriptscriptstyle\pm1.71}$ & $4.77_{\scriptscriptstyle\pm0.47}$ & $307.38_{\scriptscriptstyle\pm81.85}$ & $0.761$ \\
Pocket2Mol & $0.70_{\scriptscriptstyle\pm0.11}$ & $0.83_{\scriptscriptstyle\pm0.09}$ & $2.28_{\scriptscriptstyle\pm1.08}$ & $5.00_{\scriptscriptstyle\pm0.00}$ & $214.90_{\scriptscriptstyle\pm23.87}$ & $0.806$ \\
PocketFlow & $0.45_{\scriptscriptstyle\pm0.18}$ & $0.81_{\scriptscriptstyle\pm0.12}$ & $5.30_{\scriptscriptstyle\pm2.26}$ & $4.24_{\scriptscriptstyle\pm0.84}$ & $333.94_{\scriptscriptstyle\pm147.59}$ & $0.795$ \\
\midrule
\multicolumn{7}{l}{\textit{Selectivity-trained models}} \\
\midrule
CMD-GEN & $0.45_{\scriptscriptstyle\pm0.14}$ & $0.71_{\scriptscriptstyle\pm0.06}$ & $4.05_{\scriptscriptstyle\pm0.56}$ & $4.94_{\scriptscriptstyle\pm0.28}$ & $398.48_{\scriptscriptstyle\pm13.43}$ & $0.744$ \\
SpecLig & $0.17_{\scriptscriptstyle\pm0.07}$ & $0.68_{\scriptscriptstyle\pm0.10}$ & $6.02_{\scriptscriptstyle\pm2.34}$ & $2.57_{\scriptscriptstyle\pm0.80}$ & $630.10_{\scriptscriptstyle\pm71.06}$ & $0.540$ \\
\midrule
\multicolumn{7}{l}{\textit{Inference-time guided methods}} \\
\midrule
DecompDiff + CFG & $0.42_{\scriptscriptstyle\pm0.17}$ & $0.55_{\scriptscriptstyle\pm0.11}$ & $1.80_{\scriptscriptstyle\pm2.15}$ & $4.80_{\scriptscriptstyle\pm0.49}$ & $426.47_{\scriptscriptstyle\pm17.84}$ & $0.605$ \\
DecompDiff + FKC & $0.42_{\scriptscriptstyle\pm0.10}$ & $0.63_{\scriptscriptstyle\pm0.12}$ & $3.13_{\scriptscriptstyle\pm1.65}$ & $4.96_{\scriptscriptstyle\pm0.20}$ & $420.70_{\scriptscriptstyle\pm16.07}$ & $0.632$ \\
DiffSBDD + CFG & $0.57_{\scriptscriptstyle\pm0.19}$ & $0.70_{\scriptscriptstyle\pm0.11}$ & $3.34_{\scriptscriptstyle\pm1.82}$ & $4.54_{\scriptscriptstyle\pm0.68}$ & $383.59_{\scriptscriptstyle\pm107.70}$ & $0.624$ \\
DiffSBDD + FKC & $0.51_{\scriptscriptstyle\pm0.15}$ & $0.74_{\scriptscriptstyle\pm0.14}$ & $3.16_{\scriptscriptstyle\pm1.99}$ & $4.64_{\scriptscriptstyle\pm0.67}$ & $367.46_{\scriptscriptstyle\pm153.43}$ & $0.694$ \\
TargetDiff + CFG & $0.50_{\scriptscriptstyle\pm0.17}$ & $0.56_{\scriptscriptstyle\pm0.07}$ & $3.45_{\scriptscriptstyle\pm1.38}$ & $4.75_{\scriptscriptstyle\pm0.54}$ & $414.01_{\scriptscriptstyle\pm49.02}$ & $0.655$ \\
TargetDiff + FKC & $0.45_{\scriptscriptstyle\pm0.15}$ & $0.57_{\scriptscriptstyle\pm0.06}$ & $2.33_{\scriptscriptstyle\pm1.31}$ & $4.82_{\scriptscriptstyle\pm0.55}$ & $410.38_{\scriptscriptstyle\pm44.32}$ & $0.645$ \\
\midrule
\multicolumn{7}{l}{\hlcustom{Yellow2}{\texttt{CDK9 (on) / CDK2 (off)}}} \\
\midrule
\multicolumn{7}{l}{\textit{Standard SBDD models}} \\
\midrule
DecompDiff & $0.44_{\scriptscriptstyle\pm0.17}$ & $0.57_{\scriptscriptstyle\pm0.09}$ & $2.31_{\scriptscriptstyle\pm1.83}$ & $4.67_{\scriptscriptstyle\pm0.53}$ & $394.51_{\scriptscriptstyle\pm13.59}$ & $0.675$ \\
DiffSBDD & $0.60_{\scriptscriptstyle\pm0.20}$ & $0.60_{\scriptscriptstyle\pm0.11}$ & $2.01_{\scriptscriptstyle\pm1.81}$ & $4.80_{\scriptscriptstyle\pm0.60}$ & $328.14_{\scriptscriptstyle\pm86.10}$ & $0.729$ \\
TargetDiff & $0.36_{\scriptscriptstyle\pm0.17}$ & $0.46_{\scriptscriptstyle\pm0.11}$ & $2.70_{\scriptscriptstyle\pm1.85}$ & $4.03_{\scriptscriptstyle\pm1.06}$ & $493.11_{\scriptscriptstyle\pm86.73}$ & $0.585$ \\
MolForm & $0.32_{\scriptscriptstyle\pm0.15}$ & $0.48_{\scriptscriptstyle\pm0.12}$ & $7.29_{\scriptscriptstyle\pm1.50}$ & $3.50_{\scriptscriptstyle\pm0.77}$ & $472.39_{\scriptscriptstyle\pm92.44}$ & $0.688$ \\
MolSnapper & $0.18_{\scriptscriptstyle\pm0.10}$ & $0.59_{\scriptscriptstyle\pm0.10}$ & $-1.25_{\scriptscriptstyle\pm2.19}$ & $3.69_{\scriptscriptstyle\pm0.93}$ & $346.08_{\scriptscriptstyle\pm85.55}$ & $0.761$ \\
Pocket2Mol & $0.62_{\scriptscriptstyle\pm0.08}$ & $0.81_{\scriptscriptstyle\pm0.10}$ & $1.19_{\scriptscriptstyle\pm0.93}$ & $5.00_{\scriptscriptstyle\pm0.00}$ & $166.49_{\scriptscriptstyle\pm23.28}$ & $0.855$ \\
PocketFlow & $0.50_{\scriptscriptstyle\pm0.20}$ & $0.80_{\scriptscriptstyle\pm0.10}$ & $4.98_{\scriptscriptstyle\pm2.47}$ & $4.33_{\scriptscriptstyle\pm0.72}$ & $317.77_{\scriptscriptstyle\pm143.83}$ & $0.806$ \\
\midrule
\multicolumn{7}{l}{\textit{Selectivity-trained models}} \\
\midrule
CMD-GEN & $0.56_{\scriptscriptstyle\pm0.09}$ & $0.69_{\scriptscriptstyle\pm0.03}$ & $4.14_{\scriptscriptstyle\pm0.45}$ & $4.98_{\scriptscriptstyle\pm0.14}$ & $401.72_{\scriptscriptstyle\pm8.12}$ & $0.582$ \\
SpecLig & $0.17_{\scriptscriptstyle\pm0.07}$ & $0.67_{\scriptscriptstyle\pm0.10}$ & $5.97_{\scriptscriptstyle\pm1.74}$ & $2.70_{\scriptscriptstyle\pm0.77}$ & $626.31_{\scriptscriptstyle\pm63.73}$ & $0.556$ \\
\midrule
\multicolumn{7}{l}{\textit{Inference-time guided methods}} \\
\midrule
DecompDiff + CFG & $0.51_{\scriptscriptstyle\pm0.16}$ & $0.48_{\scriptscriptstyle\pm0.07}$ & $2.45_{\scriptscriptstyle\pm2.18}$ & $4.70_{\scriptscriptstyle\pm0.46}$ & $406.09_{\scriptscriptstyle\pm11.93}$ & $0.589$ \\
DecompDiff + FKC & $0.64_{\scriptscriptstyle\pm0.10}$ & $0.51_{\scriptscriptstyle\pm0.08}$ & $2.29_{\scriptscriptstyle\pm1.53}$ & $4.92_{\scriptscriptstyle\pm0.27}$ & $415.06_{\scriptscriptstyle\pm8.25}$ & $0.556$ \\
DiffSBDD + CFG & $0.52_{\scriptscriptstyle\pm0.17}$ & $0.55_{\scriptscriptstyle\pm0.17}$ & $3.56_{\scriptscriptstyle\pm1.68}$ & $4.44_{\scriptscriptstyle\pm0.72}$ & $463.54_{\scriptscriptstyle\pm81.37}$ & $0.448$ \\
DiffSBDD + FKC & $0.54_{\scriptscriptstyle\pm0.16}$ & $0.63_{\scriptscriptstyle\pm0.12}$ & $4.14_{\scriptscriptstyle\pm1.60}$ & $4.42_{\scriptscriptstyle\pm0.68}$ & $438.01_{\scriptscriptstyle\pm64.66}$ & $0.508$ \\
TargetDiff + CFG & $0.49_{\scriptscriptstyle\pm0.19}$ & $0.46_{\scriptscriptstyle\pm0.10}$ & $3.62_{\scriptscriptstyle\pm1.68}$ & $4.32_{\scriptscriptstyle\pm0.83}$ & $472.77_{\scriptscriptstyle\pm105.26}$ & $0.631$ \\
TargetDiff + FKC & $0.43_{\scriptscriptstyle\pm0.15}$ & $0.58_{\scriptscriptstyle\pm0.11}$ & $2.75_{\scriptscriptstyle\pm1.60}$ & $4.34_{\scriptscriptstyle\pm0.82}$ & $436.75_{\scriptscriptstyle\pm62.15}$ & $0.618$ \\
\midrule
\multicolumn{7}{l}{\hlcustom{Purple1}{\texttt{ERK2 (on) / CDK2 (off)}}} \\
\midrule
\multicolumn{7}{l}{\textit{Standard SBDD models}} \\
\midrule
DecompDiff & $0.36_{\scriptscriptstyle\pm0.17}$ & $0.55_{\scriptscriptstyle\pm0.10}$ & $2.25_{\scriptscriptstyle\pm1.72}$ & $4.62_{\scriptscriptstyle\pm0.62}$ & $444.76_{\scriptscriptstyle\pm15.21}$ & $0.640$ \\
DiffSBDD & $0.46_{\scriptscriptstyle\pm0.20}$ & $0.58_{\scriptscriptstyle\pm0.12}$ & $1.73_{\scriptscriptstyle\pm1.63}$ & $4.56_{\scriptscriptstyle\pm0.91}$ & $399.62_{\scriptscriptstyle\pm109.63}$ & $0.670$ \\
TargetDiff & $0.30_{\scriptscriptstyle\pm0.15}$ & $0.48_{\scriptscriptstyle\pm0.13}$ & $2.44_{\scriptscriptstyle\pm2.04}$ & $3.87_{\scriptscriptstyle\pm1.05}$ & $487.96_{\scriptscriptstyle\pm107.60}$ & $0.609$ \\
MolForm & $0.36_{\scriptscriptstyle\pm0.16}$ & $0.50_{\scriptscriptstyle\pm0.11}$ & $7.06_{\scriptscriptstyle\pm1.75}$ & $3.66_{\scriptscriptstyle\pm0.74}$ & $454.18_{\scriptscriptstyle\pm92.92}$ & $0.666$ \\
MolSnapper & $0.21_{\scriptscriptstyle\pm0.11}$ & $0.62_{\scriptscriptstyle\pm0.10}$ & $-0.89_{\scriptscriptstyle\pm2.06}$ & $3.88_{\scriptscriptstyle\pm1.02}$ & $339.00_{\scriptscriptstyle\pm79.40}$ & $0.761$ \\
Pocket2Mol & $0.69_{\scriptscriptstyle\pm0.11}$ & $0.78_{\scriptscriptstyle\pm0.10}$ & $1.23_{\scriptscriptstyle\pm1.10}$ & $5.00_{\scriptscriptstyle\pm0.00}$ & $222.94_{\scriptscriptstyle\pm27.53}$ & $0.790$ \\
PocketFlow & $0.51_{\scriptscriptstyle\pm0.20}$ & $0.79_{\scriptscriptstyle\pm0.12}$ & $4.68_{\scriptscriptstyle\pm1.91}$ & $4.39_{\scriptscriptstyle\pm0.85}$ & $304.18_{\scriptscriptstyle\pm139.99}$ & $0.826$ \\
\midrule
\multicolumn{7}{l}{\textit{Selectivity-trained models}} \\
\midrule
CMD-GEN & $0.59_{\scriptscriptstyle\pm0.09}$ & $0.68_{\scriptscriptstyle\pm0.03}$ & $4.00_{\scriptscriptstyle\pm0.32}$ & $5.00_{\scriptscriptstyle\pm0.00}$ & $400.25_{\scriptscriptstyle\pm10.99}$ & $0.532$ \\
SpecLig & $0.30_{\scriptscriptstyle\pm0.14}$ & $0.68_{\scriptscriptstyle\pm0.11}$ & $4.04_{\scriptscriptstyle\pm2.86}$ & $3.35_{\scriptscriptstyle\pm1.30}$ & $521.66_{\scriptscriptstyle\pm67.86}$ & $0.558$ \\
\midrule
\multicolumn{7}{l}{\textit{Inference-time guided methods}} \\
\midrule
DecompDiff + CFG & $0.38_{\scriptscriptstyle\pm0.21}$ & $0.46_{\scriptscriptstyle\pm0.09}$ & $2.17_{\scriptscriptstyle\pm2.11}$ & $4.41_{\scriptscriptstyle\pm1.07}$ & $461.84_{\scriptscriptstyle\pm18.67}$ & $0.586$ \\
DecompDiff + FKC & $0.31_{\scriptscriptstyle\pm0.07}$ & $0.59_{\scriptscriptstyle\pm0.09}$ & $1.89_{\scriptscriptstyle\pm2.49}$ & $4.43_{\scriptscriptstyle\pm1.00}$ & $448.48_{\scriptscriptstyle\pm19.50}$ & $0.510$ \\
DiffSBDD + CFG & $0.49_{\scriptscriptstyle\pm0.17}$ & $0.67_{\scriptscriptstyle\pm0.10}$ & $2.99_{\scriptscriptstyle\pm1.53}$ & $4.67_{\scriptscriptstyle\pm0.59}$ & $418.07_{\scriptscriptstyle\pm100.79}$ & $0.612$ \\
DiffSBDD + FKC & $0.50_{\scriptscriptstyle\pm0.19}$ & $0.65_{\scriptscriptstyle\pm0.14}$ & $2.90_{\scriptscriptstyle\pm1.74}$ & $4.60_{\scriptscriptstyle\pm0.69}$ & $395.55_{\scriptscriptstyle\pm116.11}$ & $0.694$ \\
TargetDiff + CFG & $0.36_{\scriptscriptstyle\pm0.18}$ & $0.45_{\scriptscriptstyle\pm0.11}$ & $3.08_{\scriptscriptstyle\pm2.20}$ & $3.98_{\scriptscriptstyle\pm1.09}$ & $513.12_{\scriptscriptstyle\pm144.22}$ & $0.607$ \\
TargetDiff + FKC & $0.33_{\scriptscriptstyle\pm0.16}$ & $0.48_{\scriptscriptstyle\pm0.08}$ & $2.99_{\scriptscriptstyle\pm1.82}$ & $3.47_{\scriptscriptstyle\pm1.13}$ & $559.33_{\scriptscriptstyle\pm115.00}$ & $0.476$ \\
\midrule
\multicolumn{7}{l}{\hlcustom{Blue2}{\texttt{DRD2 (on) / HRH1 (off)}}} \\
\midrule
\multicolumn{7}{l}{\textit{Standard SBDD models}} \\
\midrule
DecompDiff & $0.44_{\scriptscriptstyle\pm0.16}$ & $0.63_{\scriptscriptstyle\pm0.07}$ & $3.31_{\scriptscriptstyle\pm1.64}$ & $4.53_{\scriptscriptstyle\pm0.50}$ & $312.58_{\scriptscriptstyle\pm16.08}$ & $0.848$ \\
DiffSBDD & $0.51_{\scriptscriptstyle\pm0.21}$ & $0.64_{\scriptscriptstyle\pm0.13}$ & $3.07_{\scriptscriptstyle\pm2.04}$ & $4.62_{\scriptscriptstyle\pm0.70}$ & $306.01_{\scriptscriptstyle\pm99.65}$ & $0.834$ \\
TargetDiff & $0.42_{\scriptscriptstyle\pm0.19}$ & $0.61_{\scriptscriptstyle\pm0.09}$ & $3.33_{\scriptscriptstyle\pm1.92}$ & $4.42_{\scriptscriptstyle\pm0.85}$ & $363.46_{\scriptscriptstyle\pm83.54}$ & $0.788$ \\
MolForm & $0.34_{\scriptscriptstyle\pm0.09}$ & $0.63_{\scriptscriptstyle\pm0.10}$ & $7.91_{\scriptscriptstyle\pm1.47}$ & $3.57_{\scriptscriptstyle\pm0.57}$ & $352.02_{\scriptscriptstyle\pm96.53}$ & $0.760$ \\
MolSnapper & $0.62_{\scriptscriptstyle\pm0.14}$ & $0.88_{\scriptscriptstyle\pm0.07}$ & $2.24_{\scriptscriptstyle\pm1.26}$ & $4.95_{\scriptscriptstyle\pm0.22}$ & $216.33_{\scriptscriptstyle\pm72.68}$ & $0.826$ \\
Pocket2Mol & $0.53_{\scriptscriptstyle\pm0.06}$ & $0.84_{\scriptscriptstyle\pm0.09}$ & $1.03_{\scriptscriptstyle\pm0.85}$ & $5.00_{\scriptscriptstyle\pm0.00}$ & $118.58_{\scriptscriptstyle\pm23.42}$ & $0.902$ \\
PocketFlow & $0.54_{\scriptscriptstyle\pm0.19}$ & $0.78_{\scriptscriptstyle\pm0.12}$ & $4.33_{\scriptscriptstyle\pm1.84}$ & $4.50_{\scriptscriptstyle\pm0.71}$ & $308.37_{\scriptscriptstyle\pm135.13}$ & $0.843$ \\
\midrule
\multicolumn{7}{l}{\textit{Selectivity-trained models}} \\
\midrule
CMD-GEN & $0.58_{\scriptscriptstyle\pm0.08}$ & $0.67_{\scriptscriptstyle\pm0.03}$ & $4.19_{\scriptscriptstyle\pm0.39}$ & $4.99_{\scriptscriptstyle\pm0.10}$ & $402.23_{\scriptscriptstyle\pm8.80}$ & $0.562$ \\
SpecLig & $0.43_{\scriptscriptstyle\pm0.16}$ & $0.62_{\scriptscriptstyle\pm0.12}$ & $4.84_{\scriptscriptstyle\pm1.64}$ & $4.04_{\scriptscriptstyle\pm0.84}$ & $441.72_{\scriptscriptstyle\pm45.13}$ & $0.705$ \\
\midrule
\multicolumn{7}{l}{\textit{Inference-time guided methods}} \\
\midrule
DecompDiff + CFG & $0.52_{\scriptscriptstyle\pm0.14}$ & $0.55_{\scriptscriptstyle\pm0.06}$ & $2.86_{\scriptscriptstyle\pm1.88}$ & $4.85_{\scriptscriptstyle\pm0.36}$ & $347.66_{\scriptscriptstyle\pm21.88}$ & $0.731$ \\
DecompDiff + FKC & $0.58_{\scriptscriptstyle\pm0.11}$ & $0.57_{\scriptscriptstyle\pm0.10}$ & $2.97_{\scriptscriptstyle\pm1.37}$ & $4.93_{\scriptscriptstyle\pm0.26}$ & $349.14_{\scriptscriptstyle\pm21.51}$ & $0.683$ \\
DiffSBDD + CFG & $0.59_{\scriptscriptstyle\pm0.15}$ & $0.72_{\scriptscriptstyle\pm0.11}$ & $4.09_{\scriptscriptstyle\pm1.73}$ & $4.64_{\scriptscriptstyle\pm0.52}$ & $345.26_{\scriptscriptstyle\pm75.66}$ & $0.772$ \\
DiffSBDD + FKC & $0.60_{\scriptscriptstyle\pm0.15}$ & $0.74_{\scriptscriptstyle\pm0.10}$ & $3.80_{\scriptscriptstyle\pm1.27}$ & $4.81_{\scriptscriptstyle\pm0.39}$ & $321.91_{\scriptscriptstyle\pm92.15}$ & $0.803$ \\
TargetDiff + CFG & $0.42_{\scriptscriptstyle\pm0.14}$ & $0.49_{\scriptscriptstyle\pm0.06}$ & $3.57_{\scriptscriptstyle\pm1.79}$ & $4.66_{\scriptscriptstyle\pm0.53}$ & $440.10_{\scriptscriptstyle\pm23.06}$ & $0.636$ \\
TargetDiff + FKC & $0.45_{\scriptscriptstyle\pm0.17}$ & $0.51_{\scriptscriptstyle\pm0.05}$ & $2.53_{\scriptscriptstyle\pm1.64}$ & $4.65_{\scriptscriptstyle\pm0.61}$ & $439.00_{\scriptscriptstyle\pm19.71}$ & $0.488$ \\
\end{longtable}}

\subsection{Molecular generation metrics on \taskB}
\label{app:xtra_metrics_safety}
{
\footnotesize
\renewcommand{\arraystretch}{1.00}
\begin{longtable}{lcccccc}
\mycaption{\taskB evaluation on additional molecular generation metrics.}{}
\label{tab:safety_xtra_metrics} \\

\toprule
Method & QED$\,(\uparrow)$ & SA$\,(\uparrow)$ & LogP & Lipinski$\,(\uparrow)$ & MW & Diversity$\,(\uparrow)$ \\
\midrule
\endfirsthead

\toprule
Method & QED$\,(\uparrow)$ & SA$\,(\uparrow)$ & LogP & Lipinski$\,(\uparrow)$ & MW & Diversity$\,(\uparrow)$ \\
\midrule
\endhead

\midrule
\multicolumn{7}{r}{Continued on next page}
\endfoot

\bottomrule
\endlastfoot

\multicolumn{7}{l}{\hlcustom{Blue1}{\texttt{POI: 5HT1B}}} \\
\midrule
\multicolumn{7}{l}{\textit{Standard SBDD models}} \\
\midrule
DecompDiff & $0.18_{\scriptscriptstyle\pm0.09}$ & $0.45_{\scriptscriptstyle\pm0.09}$ & $6.31_{\scriptscriptstyle\pm2.35}$ & $2.56_{\scriptscriptstyle\pm0.75}$ & $605.66_{\scriptscriptstyle\pm10.96}$ & $0.662$ \\
DiffSBDD & $0.43_{\scriptscriptstyle\pm0.21}$ & $0.58_{\scriptscriptstyle\pm0.14}$ & $2.87_{\scriptscriptstyle\pm1.72}$ & $4.44_{\scriptscriptstyle\pm0.86}$ & $405.97_{\scriptscriptstyle\pm156.40}$ & $0.721$ \\
TargetDiff & $0.36_{\scriptscriptstyle\pm0.16}$ & $0.50_{\scriptscriptstyle\pm0.13}$ & $5.98_{\scriptscriptstyle\pm2.04}$ & $3.51_{\scriptscriptstyle\pm0.93}$ & $495.19_{\scriptscriptstyle\pm100.06}$ & $0.690$ \\
MolForm & $0.21_{\scriptscriptstyle\pm0.08}$ & $0.52_{\scriptscriptstyle\pm0.14}$ & $11.19_{\scriptscriptstyle\pm1.55}$ & $2.81_{\scriptscriptstyle\pm0.57}$ & $522.51_{\scriptscriptstyle\pm117.95}$ & $0.726$ \\
MolSnapper & $0.25_{\scriptscriptstyle\pm0.15}$ & $0.53_{\scriptscriptstyle\pm0.11}$ & $-2.49_{\scriptscriptstyle\pm2.23}$ & $3.80_{\scriptscriptstyle\pm0.92}$ & $310.57_{\scriptscriptstyle\pm73.23}$ & $0.619$ \\
Pocket2Mol & $0.66_{\scriptscriptstyle\pm0.12}$ & $0.82_{\scriptscriptstyle\pm0.11}$ & $2.22_{\scriptscriptstyle\pm1.14}$ & $4.98_{\scriptscriptstyle\pm0.14}$ & $188.71_{\scriptscriptstyle\pm49.10}$ & $0.868$ \\
PocketFlow & $0.40_{\scriptscriptstyle\pm0.22}$ & $0.77_{\scriptscriptstyle\pm0.10}$ & $6.55_{\scriptscriptstyle\pm2.31}$ & $3.80_{\scriptscriptstyle\pm0.96}$ & $370.35_{\scriptscriptstyle\pm140.43}$ & $0.779$ \\
\midrule
\multicolumn{7}{l}{\textit{Selectivity-trained models}} \\
\midrule
CMD-GEN (base) & $0.60_{\scriptscriptstyle\pm0.09}$ & $0.67_{\scriptscriptstyle\pm0.04}$ & $4.00_{\scriptscriptstyle\pm0.39}$ & $5.00_{\scriptscriptstyle\pm0.00}$ & $392.41_{\scriptscriptstyle\pm9.77}$ & $0.507$ \\
SpecLig & $0.22_{\scriptscriptstyle\pm0.09}$ & $0.64_{\scriptscriptstyle\pm0.13}$ & $6.25_{\scriptscriptstyle\pm2.13}$ & $2.74_{\scriptscriptstyle\pm0.81}$ & $591.12_{\scriptscriptstyle\pm70.28}$ & $0.574$ \\
\midrule
\multicolumn{7}{l}{\textit{Inference-time guided methods}} \\
\midrule
DecompDiff + CFG & $0.60_{\scriptscriptstyle\pm0.13}$ & $0.51_{\scriptscriptstyle\pm0.10}$ & $3.87_{\scriptscriptstyle\pm1.58}$ & $4.66_{\scriptscriptstyle\pm0.51}$ & $391.87_{\scriptscriptstyle\pm12.39}$ & $0.671$ \\
DecompDiff + FKC & $0.35_{\scriptscriptstyle\pm0.10}$ & $0.60_{\scriptscriptstyle\pm0.06}$ & $5.22_{\scriptscriptstyle\pm1.20}$ & $3.52_{\scriptscriptstyle\pm0.60}$ & $408.85_{\scriptscriptstyle\pm12.91}$ & $0.711$ \\
DiffSBDD + CFG & $0.45_{\scriptscriptstyle\pm0.19}$ & $0.68_{\scriptscriptstyle\pm0.10}$ & $4.80_{\scriptscriptstyle\pm1.66}$ & $3.82_{\scriptscriptstyle\pm0.91}$ & $467.07_{\scriptscriptstyle\pm91.06}$ & $0.608$ \\
DiffSBDD + FKC & $0.55_{\scriptscriptstyle\pm0.15}$ & $0.73_{\scriptscriptstyle\pm0.08}$ & $4.45_{\scriptscriptstyle\pm1.37}$ & $4.39_{\scriptscriptstyle\pm0.65}$ & $366.48_{\scriptscriptstyle\pm71.72}$ & $0.755$ \\
TargetDiff + CFG & $0.24_{\scriptscriptstyle\pm0.12}$ & $0.42_{\scriptscriptstyle\pm0.09}$ & $3.71_{\scriptscriptstyle\pm3.51}$ & $3.13_{\scriptscriptstyle\pm1.07}$ & $541.87_{\scriptscriptstyle\pm60.77}$ & $0.666$ \\
TargetDiff + FKC & $0.29_{\scriptscriptstyle\pm0.14}$ & $0.51_{\scriptscriptstyle\pm0.09}$ & $3.54_{\scriptscriptstyle\pm1.83}$ & $3.97_{\scriptscriptstyle\pm0.95}$ & $481.68_{\scriptscriptstyle\pm45.13}$ & $0.679$ \\
\midrule
\multicolumn{7}{l}{\hlcustom{Pink1}{\texttt{POI: PARP1}}} \\
\midrule
\multicolumn{7}{l}{\textit{Standard SBDD models}} \\
\midrule
DecompDiff & $0.35_{\scriptscriptstyle\pm0.16}$ & $0.61_{\scriptscriptstyle\pm0.07}$ & $1.39_{\scriptscriptstyle\pm1.62}$ & $4.15_{\scriptscriptstyle\pm0.73}$ & $371.41_{\scriptscriptstyle\pm20.96}$ & $0.780$ \\
DiffSBDD & $0.40_{\scriptscriptstyle\pm0.20}$ & $0.58_{\scriptscriptstyle\pm0.12}$ & $1.33_{\scriptscriptstyle\pm2.01}$ & $4.23_{\scriptscriptstyle\pm1.12}$ & $409.59_{\scriptscriptstyle\pm126.79}$ & $0.677$ \\
TargetDiff & $0.30_{\scriptscriptstyle\pm0.17}$ & $0.52_{\scriptscriptstyle\pm0.09}$ & $2.73_{\scriptscriptstyle\pm2.22}$ & $3.82_{\scriptscriptstyle\pm1.17}$ & $487.29_{\scriptscriptstyle\pm104.70}$ & $0.648$ \\
MolForm & $0.26_{\scriptscriptstyle\pm0.13}$ & $0.55_{\scriptscriptstyle\pm0.11}$ & $8.65_{\scriptscriptstyle\pm1.56}$ & $3.15_{\scriptscriptstyle\pm0.75}$ & $491.33_{\scriptscriptstyle\pm103.56}$ & $0.703$ \\
MolSnapper & $0.43_{\scriptscriptstyle\pm0.15}$ & $0.66_{\scriptscriptstyle\pm0.11}$ & $0.67_{\scriptscriptstyle\pm1.57}$ & $4.82_{\scriptscriptstyle\pm0.48}$ & $332.93_{\scriptscriptstyle\pm70.75}$ & $0.691$ \\
Pocket2Mol & $0.56_{\scriptscriptstyle\pm0.09}$ & $0.86_{\scriptscriptstyle\pm0.10}$ & $1.21_{\scriptscriptstyle\pm0.94}$ & $5.00_{\scriptscriptstyle\pm0.00}$ & $147.84_{\scriptscriptstyle\pm13.42}$ & $0.861$ \\
PocketFlow & $0.51_{\scriptscriptstyle\pm0.21}$ & $0.81_{\scriptscriptstyle\pm0.12}$ & $4.48_{\scriptscriptstyle\pm2.30}$ & $4.38_{\scriptscriptstyle\pm0.87}$ & $319.34_{\scriptscriptstyle\pm150.06}$ & $0.822$ \\
\midrule
\multicolumn{7}{l}{\textit{Selectivity-trained models}} \\
\midrule
CMD-GEN (base) & $0.58_{\scriptscriptstyle\pm0.11}$ & $0.68_{\scriptscriptstyle\pm0.04}$ & $3.97_{\scriptscriptstyle\pm0.59}$ & $4.95_{\scriptscriptstyle\pm0.22}$ & $388.79_{\scriptscriptstyle\pm10.22}$ & $0.548$ \\
SpecLig & $0.35_{\scriptscriptstyle\pm0.14}$ & $0.67_{\scriptscriptstyle\pm0.10}$ & $3.98_{\scriptscriptstyle\pm2.46}$ & $3.71_{\scriptscriptstyle\pm0.95}$ & $524.30_{\scriptscriptstyle\pm50.51}$ & $0.554$ \\
\midrule
\multicolumn{7}{l}{\textit{Inference-time guided methods}} \\
\midrule
DecompDiff + CFG & $0.43_{\scriptscriptstyle\pm0.15}$ & $0.50_{\scriptscriptstyle\pm0.08}$ & $0.51_{\scriptscriptstyle\pm1.81}$ & $4.72_{\scriptscriptstyle\pm0.68}$ & $413.97_{\scriptscriptstyle\pm11.12}$ & $0.641$ \\
DecompDiff + FKC & $0.40_{\scriptscriptstyle\pm0.15}$ & $0.53_{\scriptscriptstyle\pm0.08}$ & $0.98_{\scriptscriptstyle\pm1.53}$ & $4.35_{\scriptscriptstyle\pm0.86}$ & $410.34_{\scriptscriptstyle\pm7.24}$ & $0.689$ \\
DiffSBDD + CFG & $0.48_{\scriptscriptstyle\pm0.22}$ & $0.58_{\scriptscriptstyle\pm0.17}$ & $3.29_{\scriptscriptstyle\pm1.53}$ & $4.13_{\scriptscriptstyle\pm0.88}$ & $484.07_{\scriptscriptstyle\pm129.95}$ & $0.544$ \\
DiffSBDD + FKC & $0.59_{\scriptscriptstyle\pm0.19}$ & $0.71_{\scriptscriptstyle\pm0.08}$ & $3.01_{\scriptscriptstyle\pm1.37}$ & $4.86_{\scriptscriptstyle\pm0.42}$ & $382.05_{\scriptscriptstyle\pm78.07}$ & $0.619$ \\
TargetDiff + CFG & $0.25_{\scriptscriptstyle\pm0.12}$ & $0.43_{\scriptscriptstyle\pm0.10}$ & $4.39_{\scriptscriptstyle\pm2.70}$ & $3.26_{\scriptscriptstyle\pm1.17}$ & $541.79_{\scriptscriptstyle\pm59.78}$ & $0.668$ \\
TargetDiff + FKC & $0.39_{\scriptscriptstyle\pm0.16}$ & $0.48_{\scriptscriptstyle\pm0.11}$ & $3.44_{\scriptscriptstyle\pm1.83}$ & $4.42_{\scriptscriptstyle\pm1.04}$ & $470.85_{\scriptscriptstyle\pm39.16}$ & $0.630$ \\
\midrule
\multicolumn{7}{l}{\hlcustom{Yellow2}{\texttt{POI: FA7}}} \\
\midrule
\multicolumn{7}{l}{\textit{Standard SBDD models}} \\
\midrule
DecompDiff & $0.11_{\scriptscriptstyle\pm0.06}$ & $0.51_{\scriptscriptstyle\pm0.03}$ & $2.72_{\scriptscriptstyle\pm1.90}$ & $2.30_{\scriptscriptstyle\pm0.81}$ & $575.43_{\scriptscriptstyle\pm16.23}$ & $0.644$ \\
DiffSBDD & $0.45_{\scriptscriptstyle\pm0.22}$ & $0.63_{\scriptscriptstyle\pm0.12}$ & $2.00_{\scriptscriptstyle\pm1.74}$ & $4.60_{\scriptscriptstyle\pm0.89}$ & $358.50_{\scriptscriptstyle\pm92.60}$ & $0.744$ \\
TargetDiff & $0.27_{\scriptscriptstyle\pm0.18}$ & $0.52_{\scriptscriptstyle\pm0.09}$ & $2.07_{\scriptscriptstyle\pm1.97}$ & $3.55_{\scriptscriptstyle\pm1.46}$ & $487.50_{\scriptscriptstyle\pm96.85}$ & $0.595$ \\
MolForm & $0.29_{\scriptscriptstyle\pm0.17}$ & $0.53_{\scriptscriptstyle\pm0.12}$ & $7.81_{\scriptscriptstyle\pm1.80}$ & $3.39_{\scriptscriptstyle\pm0.79}$ & $472.08_{\scriptscriptstyle\pm114.12}$ & $0.689$ \\
MolSnapper & $0.32_{\scriptscriptstyle\pm0.17}$ & $0.57_{\scriptscriptstyle\pm0.15}$ & $-1.32_{\scriptscriptstyle\pm2.17}$ & $4.11_{\scriptscriptstyle\pm1.03}$ & $338.98_{\scriptscriptstyle\pm69.66}$ & $0.622$ \\
Pocket2Mol & $0.65_{\scriptscriptstyle\pm0.10}$ & $0.85_{\scriptscriptstyle\pm0.10}$ & $1.39_{\scriptscriptstyle\pm1.15}$ & $5.00_{\scriptscriptstyle\pm0.00}$ & $178.95_{\scriptscriptstyle\pm28.12}$ & $0.873$ \\
PocketFlow & $0.51_{\scriptscriptstyle\pm0.21}$ & $0.81_{\scriptscriptstyle\pm0.11}$ & $3.11_{\scriptscriptstyle\pm2.22}$ & $4.60_{\scriptscriptstyle\pm0.79}$ & $294.40_{\scriptscriptstyle\pm145.17}$ & $0.837$ \\
\midrule
\multicolumn{7}{l}{\textit{Selectivity-trained models}} \\
\midrule
CMD-GEN (base) & $0.62_{\scriptscriptstyle\pm0.09}$ & $0.67_{\scriptscriptstyle\pm0.04}$ & $4.19_{\scriptscriptstyle\pm0.39}$ & $4.99_{\scriptscriptstyle\pm0.10}$ & $406.05_{\scriptscriptstyle\pm9.93}$ & $0.582$ \\
SpecLig & $0.21_{\scriptscriptstyle\pm0.09}$ & $0.70_{\scriptscriptstyle\pm0.09}$ & $3.94_{\scriptscriptstyle\pm1.85}$ & $3.71_{\scriptscriptstyle\pm1.20}$ & $514.97_{\scriptscriptstyle\pm65.04}$ & $0.570$ \\
\midrule
\multicolumn{7}{l}{\textit{Inference-time guided methods}} \\
\midrule
DecompDiff + CFG & $0.31_{\scriptscriptstyle\pm0.14}$ & $0.40_{\scriptscriptstyle\pm0.06}$ & $-0.31_{\scriptscriptstyle\pm1.54}$ & $4.40_{\scriptscriptstyle\pm1.02}$ & $465.26_{\scriptscriptstyle\pm12.91}$ & $0.532$ \\
DecompDiff + FKC & $0.20_{\scriptscriptstyle\pm0.10}$ & $0.55_{\scriptscriptstyle\pm0.06}$ & $-2.01_{\scriptscriptstyle\pm1.89}$ & $3.04_{\scriptscriptstyle\pm1.35}$ & $466.24_{\scriptscriptstyle\pm13.34}$ & $0.604$ \\
DiffSBDD + CFG & $0.30_{\scriptscriptstyle\pm0.15}$ & $0.65_{\scriptscriptstyle\pm0.07}$ & $4.09_{\scriptscriptstyle\pm1.80}$ & $3.68_{\scriptscriptstyle\pm0.99}$ & $501.78_{\scriptscriptstyle\pm66.87}$ & $0.571$ \\
DiffSBDD + FKC & $0.37_{\scriptscriptstyle\pm0.15}$ & $0.72_{\scriptscriptstyle\pm0.10}$ & $3.19_{\scriptscriptstyle\pm2.18}$ & $4.37_{\scriptscriptstyle\pm0.81}$ & $365.72_{\scriptscriptstyle\pm163.51}$ & $0.732$ \\
TargetDiff + CFG & $0.28_{\scriptscriptstyle\pm0.14}$ & $0.42_{\scriptscriptstyle\pm0.10}$ & $3.82_{\scriptscriptstyle\pm2.59}$ & $3.51_{\scriptscriptstyle\pm0.98}$ & $533.38_{\scriptscriptstyle\pm61.34}$ & $0.665$ \\
TargetDiff + FKC & $0.33_{\scriptscriptstyle\pm0.14}$ & $0.49_{\scriptscriptstyle\pm0.06}$ & $3.19_{\scriptscriptstyle\pm2.58}$ & $4.17_{\scriptscriptstyle\pm0.95}$ & $481.02_{\scriptscriptstyle\pm39.15}$ & $0.650$ \\
\midrule
\multicolumn{7}{l}{\hlcustom{Purple1}{\texttt{POI: JAK2}}} \\
\midrule
\multicolumn{7}{l}{\textit{Standard SBDD models}} \\
\midrule
DecompDiff & $0.32_{\scriptscriptstyle\pm0.13}$ & $0.53_{\scriptscriptstyle\pm0.10}$ & $2.27_{\scriptscriptstyle\pm1.87}$ & $4.35_{\scriptscriptstyle\pm0.74}$ & $470.77_{\scriptscriptstyle\pm15.33}$ & $0.596$ \\
DiffSBDD & $0.55_{\scriptscriptstyle\pm0.23}$ & $0.59_{\scriptscriptstyle\pm0.12}$ & $2.41_{\scriptscriptstyle\pm1.69}$ & $4.62_{\scriptscriptstyle\pm0.77}$ & $382.42_{\scriptscriptstyle\pm98.35}$ & $0.634$ \\
TargetDiff & $0.54_{\scriptscriptstyle\pm0.19}$ & $0.55_{\scriptscriptstyle\pm0.09}$ & $1.88_{\scriptscriptstyle\pm1.75}$ & $4.74_{\scriptscriptstyle\pm0.48}$ & $360.84_{\scriptscriptstyle\pm71.29}$ & $0.651$ \\
MolForm & $0.49_{\scriptscriptstyle\pm0.15}$ & $0.57_{\scriptscriptstyle\pm0.16}$ & $6.05_{\scriptscriptstyle\pm1.23}$ & $4.07_{\scriptscriptstyle\pm0.45}$ & $363.07_{\scriptscriptstyle\pm70.12}$ & $0.715$ \\
MolSnapper & $0.33_{\scriptscriptstyle\pm0.15}$ & $0.67_{\scriptscriptstyle\pm0.12}$ & $0.67_{\scriptscriptstyle\pm1.50}$ & $4.89_{\scriptscriptstyle\pm0.40}$ & $305.53_{\scriptscriptstyle\pm72.72}$ & $0.784$ \\
Pocket2Mol & $0.64_{\scriptscriptstyle\pm0.10}$ & $0.80_{\scriptscriptstyle\pm0.09}$ & $1.29_{\scriptscriptstyle\pm0.87}$ & $5.00_{\scriptscriptstyle\pm0.00}$ & $192.82_{\scriptscriptstyle\pm22.97}$ & $0.806$ \\
PocketFlow & $0.50_{\scriptscriptstyle\pm0.20}$ & $0.80_{\scriptscriptstyle\pm0.13}$ & $4.22_{\scriptscriptstyle\pm2.21}$ & $4.40_{\scriptscriptstyle\pm1.02}$ & $296.26_{\scriptscriptstyle\pm153.82}$ & $0.824$ \\
\midrule
\multicolumn{7}{l}{\textit{Selectivity-trained models}} \\
\midrule
CMD-GEN (base) & $0.62_{\scriptscriptstyle\pm0.08}$ & $0.67_{\scriptscriptstyle\pm0.05}$ & $4.12_{\scriptscriptstyle\pm0.40}$ & $4.98_{\scriptscriptstyle\pm0.14}$ & $400.35_{\scriptscriptstyle\pm10.77}$ & $0.586$ \\
SpecLig & $0.61_{\scriptscriptstyle\pm0.15}$ & $0.71_{\scriptscriptstyle\pm0.08}$ & $2.93_{\scriptscriptstyle\pm1.55}$ & $4.77_{\scriptscriptstyle\pm0.61}$ & $407.41_{\scriptscriptstyle\pm53.41}$ & $0.601$ \\
\midrule
\multicolumn{7}{l}{\textit{Inference-time guided methods}} \\
\midrule
DecompDiff + CFG & $0.63_{\scriptscriptstyle\pm0.16}$ & $0.62_{\scriptscriptstyle\pm0.10}$ & $0.67_{\scriptscriptstyle\pm1.56}$ & $4.93_{\scriptscriptstyle\pm0.26}$ & $306.99_{\scriptscriptstyle\pm9.79}$ & $0.712$ \\
DecompDiff + FKC & $0.53_{\scriptscriptstyle\pm0.19}$ & $0.65_{\scriptscriptstyle\pm0.08}$ & $0.02_{\scriptscriptstyle\pm1.11}$ & $4.91_{\scriptscriptstyle\pm0.29}$ & $309.51_{\scriptscriptstyle\pm7.27}$ & $0.695$ \\
DiffSBDD + CFG & $0.43_{\scriptscriptstyle\pm0.17}$ & $0.54_{\scriptscriptstyle\pm0.14}$ & $4.56_{\scriptscriptstyle\pm1.66}$ & $4.16_{\scriptscriptstyle\pm0.81}$ & $500.63_{\scriptscriptstyle\pm92.77}$ & $0.384$ \\
DiffSBDD + FKC & $0.52_{\scriptscriptstyle\pm0.16}$ & $0.65_{\scriptscriptstyle\pm0.11}$ & $3.99_{\scriptscriptstyle\pm1.66}$ & $4.55_{\scriptscriptstyle\pm0.64}$ & $420.09_{\scriptscriptstyle\pm73.21}$ & $0.488$ \\
TargetDiff + CFG & $0.28_{\scriptscriptstyle\pm0.12}$ & $0.42_{\scriptscriptstyle\pm0.10}$ & $3.61_{\scriptscriptstyle\pm3.09}$ & $3.57_{\scriptscriptstyle\pm1.18}$ & $512.92_{\scriptscriptstyle\pm64.02}$ & $0.683$ \\
TargetDiff + FKC & $0.38_{\scriptscriptstyle\pm0.15}$ & $0.48_{\scriptscriptstyle\pm0.08}$ & $3.63_{\scriptscriptstyle\pm1.88}$ & $4.46_{\scriptscriptstyle\pm0.67}$ & $457.87_{\scriptscriptstyle\pm10.07}$ & $0.660$ \\
\midrule
\multicolumn{7}{l}{\hlcustom{Blue2}{\texttt{POI: BRAF}}} \\
\midrule
\multicolumn{7}{l}{\textit{Standard SBDD models}} \\
\midrule
DecompDiff & $0.46_{\scriptscriptstyle\pm0.18}$ & $0.57_{\scriptscriptstyle\pm0.11}$ & $2.11_{\scriptscriptstyle\pm1.82}$ & $4.69_{\scriptscriptstyle\pm0.59}$ & $394.01_{\scriptscriptstyle\pm11.53}$ & $0.700$ \\
DiffSBDD & $0.52_{\scriptscriptstyle\pm0.20}$ & $0.59_{\scriptscriptstyle\pm0.12}$ & $1.80_{\scriptscriptstyle\pm1.65}$ & $4.77_{\scriptscriptstyle\pm0.51}$ & $372.20_{\scriptscriptstyle\pm103.42}$ & $0.645$ \\
TargetDiff & $0.29_{\scriptscriptstyle\pm0.17}$ & $0.52_{\scriptscriptstyle\pm0.09}$ & $2.42_{\scriptscriptstyle\pm1.80}$ & $3.87_{\scriptscriptstyle\pm1.22}$ & $489.39_{\scriptscriptstyle\pm97.92}$ & $0.584$ \\
MolForm & $0.35_{\scriptscriptstyle\pm0.15}$ & $0.47_{\scriptscriptstyle\pm0.12}$ & $6.62_{\scriptscriptstyle\pm1.53}$ & $3.74_{\scriptscriptstyle\pm0.70}$ & $471.47_{\scriptscriptstyle\pm98.39}$ & $0.659$ \\
MolSnapper & $0.39_{\scriptscriptstyle\pm0.16}$ & $0.63_{\scriptscriptstyle\pm0.12}$ & $0.54_{\scriptscriptstyle\pm1.54}$ & $4.78_{\scriptscriptstyle\pm0.44}$ & $306.29_{\scriptscriptstyle\pm75.39}$ & $0.775$ \\
Pocket2Mol & $0.68_{\scriptscriptstyle\pm0.14}$ & $0.78_{\scriptscriptstyle\pm0.10}$ & $1.88_{\scriptscriptstyle\pm1.39}$ & $4.97_{\scriptscriptstyle\pm0.17}$ & $265.24_{\scriptscriptstyle\pm48.52}$ & $0.731$ \\
PocketFlow & $0.49_{\scriptscriptstyle\pm0.22}$ & $0.80_{\scriptscriptstyle\pm0.11}$ & $4.82_{\scriptscriptstyle\pm2.46}$ & $4.31_{\scriptscriptstyle\pm0.89}$ & $327.30_{\scriptscriptstyle\pm151.12}$ & $0.822$ \\
\midrule
\multicolumn{7}{l}{\textit{Selectivity-trained models}} \\
\midrule
CMD-GEN (base) & $0.60_{\scriptscriptstyle\pm0.10}$ & $0.68_{\scriptscriptstyle\pm0.03}$ & $4.17_{\scriptscriptstyle\pm0.42}$ & $4.99_{\scriptscriptstyle\pm0.10}$ & $399.31_{\scriptscriptstyle\pm10.98}$ & $0.535$ \\
SpecLig & $0.37_{\scriptscriptstyle\pm0.11}$ & $0.70_{\scriptscriptstyle\pm0.11}$ & $4.61_{\scriptscriptstyle\pm1.66}$ & $3.96_{\scriptscriptstyle\pm0.92}$ & $495.00_{\scriptscriptstyle\pm51.76}$ & $0.523$ \\
\midrule
\multicolumn{7}{l}{\textit{Inference-time guided methods}} \\
\midrule
DecompDiff + CFG & $0.54_{\scriptscriptstyle\pm0.17}$ & $0.46_{\scriptscriptstyle\pm0.08}$ & $1.88_{\scriptscriptstyle\pm1.72}$ & $4.89_{\scriptscriptstyle\pm0.37}$ & $399.28_{\scriptscriptstyle\pm14.47}$ & $0.634$ \\
DecompDiff + FKC & $0.56_{\scriptscriptstyle\pm0.12}$ & $0.57_{\scriptscriptstyle\pm0.06}$ & $1.49_{\scriptscriptstyle\pm1.05}$ & $4.97_{\scriptscriptstyle\pm0.17}$ & $398.88_{\scriptscriptstyle\pm8.31}$ & $0.564$ \\
DiffSBDD + CFG & $0.50_{\scriptscriptstyle\pm0.23}$ & $0.64_{\scriptscriptstyle\pm0.16}$ & $3.22_{\scriptscriptstyle\pm1.58}$ & $4.28_{\scriptscriptstyle\pm0.84}$ & $447.27_{\scriptscriptstyle\pm148.07}$ & $0.610$ \\
DiffSBDD + FKC & $0.55_{\scriptscriptstyle\pm0.20}$ & $0.73_{\scriptscriptstyle\pm0.12}$ & $2.42_{\scriptscriptstyle\pm1.81}$ & $4.64_{\scriptscriptstyle\pm0.69}$ & $330.81_{\scriptscriptstyle\pm139.63}$ & $0.750$ \\
TargetDiff + CFG & $0.23_{\scriptscriptstyle\pm0.12}$ & $0.42_{\scriptscriptstyle\pm0.09}$ & $3.92_{\scriptscriptstyle\pm3.42}$ & $3.16_{\scriptscriptstyle\pm1.10}$ & $542.00_{\scriptscriptstyle\pm61.26}$ & $0.669$ \\
TargetDiff + FKC & $0.21_{\scriptscriptstyle\pm0.13}$ & $0.48_{\scriptscriptstyle\pm0.09}$ & $2.80_{\scriptscriptstyle\pm2.08}$ & $3.15_{\scriptscriptstyle\pm1.49}$ & $538.43_{\scriptscriptstyle\pm82.04}$ & $0.673$ \\
\end{longtable}}

\subsection{Molecular generation metrics on \taskC}
\label{app:xtra_metrics_xl}
{
\footnotesize
\renewcommand{\arraystretch}{1.00}
\begin{longtable}{lcccccc}
\mycaption{\taskB evaluation on additional molecular generation metrics.}{}
\label{tab:safety_xtra_metrics} \\

\toprule
Method & QED$\,(\uparrow)$ & SA$\,(\uparrow)$ & LogP & Lipinski$\,(\uparrow)$ & MW & Diversity$\,(\uparrow)$ \\
\midrule
\endfirsthead

\toprule
Method & QED$\,(\uparrow)$ & SA$\,(\uparrow)$ & LogP & Lipinski$\,(\uparrow)$ & MW & Diversity$\,(\uparrow)$ \\
\midrule
\endhead

\midrule
\multicolumn{7}{r}{Continued on next page}
\endfoot

\bottomrule
\endlastfoot

\multicolumn{7}{l}{\hlcustom{Blue1}{\texttt{Hard}}} \\
\midrule
\multicolumn{7}{l}{\textit{Standard SBDD models}} \\
\midrule
DecompDiff & $0.44_{\scriptscriptstyle\pm0.12}$ & $0.68_{\scriptscriptstyle\pm0.10}$ & $1.92_{\scriptscriptstyle\pm2.24}$ & $4.44_{\scriptscriptstyle\pm0.77}$ & $270.66_{\scriptscriptstyle\pm163.66}$ & $0.80_{\scriptscriptstyle\pm0.08}$ \\
DiffSBDD & $0.50_{\scriptscriptstyle\pm0.05}$ & $0.64_{\scriptscriptstyle\pm0.04}$ & $0.83_{\scriptscriptstyle\pm0.82}$ & $4.77_{\scriptscriptstyle\pm0.17}$ & $270.73_{\scriptscriptstyle\pm67.72}$ & $0.82_{\scriptscriptstyle\pm0.05}$ \\
TargetDiff & $0.47_{\scriptscriptstyle\pm0.07}$ & $0.58_{\scriptscriptstyle\pm0.06}$ & $1.44_{\scriptscriptstyle\pm1.74}$ & $4.53_{\scriptscriptstyle\pm0.32}$ & $337.44_{\scriptscriptstyle\pm110.22}$ & $0.72_{\scriptscriptstyle\pm0.09}$ \\
MolForm & $0.45_{\scriptscriptstyle\pm0.07}$ & $0.67_{\scriptscriptstyle\pm0.11}$ & $5.62_{\scriptscriptstyle\pm1.93}$ & $4.17_{\scriptscriptstyle\pm0.56}$ & $309.89_{\scriptscriptstyle\pm117.62}$ & $0.75_{\scriptscriptstyle\pm0.08}$ \\
MolSnapper & $0.52_{\scriptscriptstyle\pm0.11}$ & $0.73_{\scriptscriptstyle\pm0.09}$ & $1.82_{\scriptscriptstyle\pm1.46}$ & $4.83_{\scriptscriptstyle\pm0.46}$ & $322.69_{\scriptscriptstyle\pm22.52}$ & $0.73_{\scriptscriptstyle\pm0.05}$ \\
Pocket2Mol & $0.58_{\scriptscriptstyle\pm0.11}$ & $0.82_{\scriptscriptstyle\pm0.04}$ & $0.94_{\scriptscriptstyle\pm1.04}$ & $4.99_{\scriptscriptstyle\pm0.01}$ & $158.63_{\scriptscriptstyle\pm45.46}$ & $0.87_{\scriptscriptstyle\pm0.04}$ \\
PocketFlow & $0.53_{\scriptscriptstyle\pm0.04}$ & $0.82_{\scriptscriptstyle\pm0.03}$ & $3.72_{\scriptscriptstyle\pm1.14}$ & $4.56_{\scriptscriptstyle\pm0.21}$ & $267.95_{\scriptscriptstyle\pm36.15}$ & $0.85_{\scriptscriptstyle\pm0.02}$ \\
\midrule
\multicolumn{7}{l}{\textit{Selectivity-trained models}} \\
\midrule
CMD-GEN & $0.61_{\scriptscriptstyle\pm0.03}$ & $0.67_{\scriptscriptstyle\pm0.01}$ & $4.08_{\scriptscriptstyle\pm0.08}$ & $4.99_{\scriptscriptstyle\pm0.02}$ & $404.10_{\scriptscriptstyle\pm5.56}$ & $0.61_{\scriptscriptstyle\pm0.05}$ \\
SpecLig & $0.48_{\scriptscriptstyle\pm0.09}$ & $0.70_{\scriptscriptstyle\pm0.06}$ & $1.07_{\scriptscriptstyle\pm1.89}$ & $4.55_{\scriptscriptstyle\pm0.38}$ & $304.00_{\scriptscriptstyle\pm129.90}$ & $0.76_{\scriptscriptstyle\pm0.11}$ \\
\midrule
\multicolumn{7}{l}{\textit{Inference-time guided methods}} \\
\midrule
DecompDiff + CFG & $0.51_{\scriptscriptstyle\pm0.11}$ & $0.61_{\scriptscriptstyle\pm0.13}$ & $0.81_{\scriptscriptstyle\pm1.56}$ & $4.63_{\scriptscriptstyle\pm0.39}$ & $291.67_{\scriptscriptstyle\pm142.68}$ & $0.73_{\scriptscriptstyle\pm0.12}$ \\
DecompDiff + FKC & $0.50_{\scriptscriptstyle\pm0.13}$ & $0.66_{\scriptscriptstyle\pm0.16}$ & $0.78_{\scriptscriptstyle\pm1.71}$ & $4.62_{\scriptscriptstyle\pm0.52}$ & $291.85_{\scriptscriptstyle\pm144.48}$ & $0.61_{\scriptscriptstyle\pm0.14}$ \\
DiffSBDD + CFG & $0.58_{\scriptscriptstyle\pm0.07}$ & $0.73_{\scriptscriptstyle\pm0.06}$ & $2.78_{\scriptscriptstyle\pm1.46}$ & $4.71_{\scriptscriptstyle\pm0.25}$ & $308.10_{\scriptscriptstyle\pm82.97}$ & $0.73_{\scriptscriptstyle\pm0.09}$ \\
DiffSBDD + FKC & $0.59_{\scriptscriptstyle\pm0.07}$ & $0.74_{\scriptscriptstyle\pm0.05}$ & $3.04_{\scriptscriptstyle\pm1.44}$ & $4.72_{\scriptscriptstyle\pm0.20}$ & $312.45_{\scriptscriptstyle\pm69.60}$ & $0.72_{\scriptscriptstyle\pm0.07}$ \\
TargetDiff + CFG & $0.46_{\scriptscriptstyle\pm0.11}$ & $0.49_{\scriptscriptstyle\pm0.05}$ & $3.76_{\scriptscriptstyle\pm1.13}$ & $4.39_{\scriptscriptstyle\pm0.42}$ & $442.85_{\scriptscriptstyle\pm33.98}$ & $0.60_{\scriptscriptstyle\pm0.14}$ \\
TargetDiff + FKC & $0.39_{\scriptscriptstyle\pm0.10}$ & $0.54_{\scriptscriptstyle\pm0.05}$ & $4.01_{\scriptscriptstyle\pm2.03}$ & $4.15_{\scriptscriptstyle\pm0.45}$ & $441.45_{\scriptscriptstyle\pm31.22}$ & $0.64_{\scriptscriptstyle\pm0.07}$ \\
\midrule
\multicolumn{7}{l}{\hlcustom{Pink1}{\texttt{Medium}}} \\
\midrule
\multicolumn{7}{l}{\textit{Standard SBDD models}} \\
\midrule
DecompDiff & $0.47_{\scriptscriptstyle\pm0.16}$ & $0.61_{\scriptscriptstyle\pm0.07}$ & $2.42_{\scriptscriptstyle\pm1.49}$ & $4.50_{\scriptscriptstyle\pm0.70}$ & $365.99_{\scriptscriptstyle\pm123.45}$ & $0.72_{\scriptscriptstyle\pm0.08}$ \\
DiffSBDD & $0.52_{\scriptscriptstyle\pm0.06}$ & $0.61_{\scriptscriptstyle\pm0.04}$ & $1.41_{\scriptscriptstyle\pm0.68}$ & $4.71_{\scriptscriptstyle\pm0.22}$ & $324.03_{\scriptscriptstyle\pm62.20}$ & $0.75_{\scriptscriptstyle\pm0.07}$ \\
TargetDiff & $0.52_{\scriptscriptstyle\pm0.08}$ & $0.55_{\scriptscriptstyle\pm0.05}$ & $2.16_{\scriptscriptstyle\pm0.96}$ & $4.61_{\scriptscriptstyle\pm0.32}$ & $371.73_{\scriptscriptstyle\pm75.28}$ & $0.66_{\scriptscriptstyle\pm0.08}$ \\
MolForm & $0.45_{\scriptscriptstyle\pm0.09}$ & $0.59_{\scriptscriptstyle\pm0.09}$ & $6.23_{\scriptscriptstyle\pm1.39}$ & $4.00_{\scriptscriptstyle\pm0.44}$ & $355.92_{\scriptscriptstyle\pm82.96}$ & $0.74_{\scriptscriptstyle\pm0.06}$ \\
MolSnapper & $0.42_{\scriptscriptstyle\pm0.18}$ & $0.66_{\scriptscriptstyle\pm0.12}$ & $-0.17_{\scriptscriptstyle\pm3.24}$ & $4.21_{\scriptscriptstyle\pm1.16}$ & $330.28_{\scriptscriptstyle\pm21.64}$ & $0.72_{\scriptscriptstyle\pm0.05}$ \\
Pocket2Mol & $0.62_{\scriptscriptstyle\pm0.06}$ & $0.80_{\scriptscriptstyle\pm0.05}$ & $1.12_{\scriptscriptstyle\pm0.65}$ & $4.99_{\scriptscriptstyle\pm0.02}$ & $196.90_{\scriptscriptstyle\pm48.29}$ & $0.81_{\scriptscriptstyle\pm0.07}$ \\
PocketFlow & $0.51_{\scriptscriptstyle\pm0.04}$ & $0.81_{\scriptscriptstyle\pm0.04}$ & $4.33_{\scriptscriptstyle\pm0.48}$ & $4.46_{\scriptscriptstyle\pm0.17}$ & $292.98_{\scriptscriptstyle\pm35.84}$ & $0.82_{\scriptscriptstyle\pm0.02}$ \\
\midrule
\multicolumn{7}{l}{\textit{Selectivity-trained models}} \\
\midrule
CMD-GEN & $0.62_{\scriptscriptstyle\pm0.03}$ & $0.67_{\scriptscriptstyle\pm0.01}$ & $4.09_{\scriptscriptstyle\pm0.09}$ & $4.99_{\scriptscriptstyle\pm0.01}$ & $404.85_{\scriptscriptstyle\pm3.61}$ & $0.61_{\scriptscriptstyle\pm0.05}$ \\
SpecLig & $0.52_{\scriptscriptstyle\pm0.12}$ & $0.71_{\scriptscriptstyle\pm0.05}$ & $2.22_{\scriptscriptstyle\pm1.26}$ & $4.46_{\scriptscriptstyle\pm0.40}$ & $368.81_{\scriptscriptstyle\pm94.43}$ & $0.69_{\scriptscriptstyle\pm0.09}$ \\
\midrule
\multicolumn{7}{l}{\textit{Inference-time guided methods}} \\
\midrule
DecompDiff + CFG & $0.55_{\scriptscriptstyle\pm0.13}$ & $0.59_{\scriptscriptstyle\pm0.12}$ & $1.03_{\scriptscriptstyle\pm1.16}$ & $4.71_{\scriptscriptstyle\pm0.45}$ & $322.30_{\scriptscriptstyle\pm112.50}$ & $0.68_{\scriptscriptstyle\pm0.13}$ \\
DecompDiff + FKC & $0.57_{\scriptscriptstyle\pm0.17}$ & $0.66_{\scriptscriptstyle\pm0.16}$ & $1.14_{\scriptscriptstyle\pm1.62}$ & $4.59_{\scriptscriptstyle\pm0.76}$ & $323.62_{\scriptscriptstyle\pm116.93}$ & $0.58_{\scriptscriptstyle\pm0.16}$ \\
DiffSBDD + CFG & $0.57_{\scriptscriptstyle\pm0.08}$ & $0.71_{\scriptscriptstyle\pm0.07}$ & $3.31_{\scriptscriptstyle\pm0.85}$ & $4.73_{\scriptscriptstyle\pm0.18}$ & $349.94_{\scriptscriptstyle\pm60.08}$ & $0.68_{\scriptscriptstyle\pm0.07}$ \\
DiffSBDD + FKC & $0.56_{\scriptscriptstyle\pm0.08}$ & $0.71_{\scriptscriptstyle\pm0.08}$ & $3.32_{\scriptscriptstyle\pm0.91}$ & $4.72_{\scriptscriptstyle\pm0.18}$ & $345.03_{\scriptscriptstyle\pm65.55}$ & $0.69_{\scriptscriptstyle\pm0.07}$ \\
TargetDiff + CFG & $0.46_{\scriptscriptstyle\pm0.09}$ & $0.48_{\scriptscriptstyle\pm0.04}$ & $3.96_{\scriptscriptstyle\pm1.19}$ & $4.34_{\scriptscriptstyle\pm0.43}$ & $444.68_{\scriptscriptstyle\pm37.19}$ & $0.61_{\scriptscriptstyle\pm0.09}$ \\
TargetDiff + FKC & $0.38_{\scriptscriptstyle\pm0.14}$ & $0.49_{\scriptscriptstyle\pm0.07}$ & $3.75_{\scriptscriptstyle\pm1.58}$ & $4.09_{\scriptscriptstyle\pm0.78}$ & $471.42_{\scriptscriptstyle\pm65.61}$ & $0.57_{\scriptscriptstyle\pm0.15}$ \\
\midrule
\multicolumn{7}{l}{\hlcustom{Yellow2}{\texttt{Easy}}} \\
\midrule
\multicolumn{7}{l}{\textit{Standard SBDD models}} \\
\midrule
DecompDiff & $0.56_{\scriptscriptstyle\pm0.13}$ & $0.70_{\scriptscriptstyle\pm0.09}$ & $1.61_{\scriptscriptstyle\pm1.43}$ & $4.76_{\scriptscriptstyle\pm0.38}$ & $255.53_{\scriptscriptstyle\pm89.58}$ & $0.80_{\scriptscriptstyle\pm0.06}$ \\
DiffSBDD & $0.51_{\scriptscriptstyle\pm0.07}$ & $0.63_{\scriptscriptstyle\pm0.07}$ & $1.09_{\scriptscriptstyle\pm1.02}$ & $4.76_{\scriptscriptstyle\pm0.19}$ & $281.23_{\scriptscriptstyle\pm70.63}$ & $0.80_{\scriptscriptstyle\pm0.07}$ \\
TargetDiff & $0.50_{\scriptscriptstyle\pm0.10}$ & $0.61_{\scriptscriptstyle\pm0.08}$ & $1.31_{\scriptscriptstyle\pm1.77}$ & $4.62_{\scriptscriptstyle\pm0.38}$ & $310.60_{\scriptscriptstyle\pm93.39}$ & $0.74_{\scriptscriptstyle\pm0.10}$ \\
MolForm & $0.50_{\scriptscriptstyle\pm0.06}$ & $0.67_{\scriptscriptstyle\pm0.09}$ & $5.10_{\scriptscriptstyle\pm1.63}$ & $4.41_{\scriptscriptstyle\pm0.41}$ & $285.77_{\scriptscriptstyle\pm92.72}$ & $0.79_{\scriptscriptstyle\pm0.06}$ \\
MolSnapper & $0.53_{\scriptscriptstyle\pm0.12}$ & $0.69_{\scriptscriptstyle\pm0.11}$ & $1.57_{\scriptscriptstyle\pm1.68}$ & $4.77_{\scriptscriptstyle\pm0.53}$ & $320.86_{\scriptscriptstyle\pm17.82}$ & $0.71_{\scriptscriptstyle\pm0.07}$ \\
Pocket2Mol & $0.56_{\scriptscriptstyle\pm0.08}$ & $0.83_{\scriptscriptstyle\pm0.04}$ & $0.94_{\scriptscriptstyle\pm0.98}$ & $4.98_{\scriptscriptstyle\pm0.04}$ & $154.44_{\scriptscriptstyle\pm37.37}$ & $0.86_{\scriptscriptstyle\pm0.04}$ \\
PocketFlow & $0.53_{\scriptscriptstyle\pm0.04}$ & $0.81_{\scriptscriptstyle\pm0.05}$ & $3.45_{\scriptscriptstyle\pm1.19}$ & $4.64_{\scriptscriptstyle\pm0.15}$ & $263.05_{\scriptscriptstyle\pm29.91}$ & $0.85_{\scriptscriptstyle\pm0.02}$ \\
\midrule
\multicolumn{7}{l}{\textit{Selectivity-trained models}} \\
\midrule
CMD-GEN & $0.62_{\scriptscriptstyle\pm0.02}$ & $0.67_{\scriptscriptstyle\pm0.01}$ & $4.05_{\scriptscriptstyle\pm0.08}$ & $4.99_{\scriptscriptstyle\pm0.01}$ & $403.22_{\scriptscriptstyle\pm5.05}$ & $0.62_{\scriptscriptstyle\pm0.04}$ \\
SpecLig & $0.51_{\scriptscriptstyle\pm0.14}$ & $0.70_{\scriptscriptstyle\pm0.09}$ & $0.73_{\scriptscriptstyle\pm1.65}$ & $4.58_{\scriptscriptstyle\pm0.32}$ & $289.66_{\scriptscriptstyle\pm87.24}$ & $0.77_{\scriptscriptstyle\pm0.08}$ \\
\midrule
\multicolumn{7}{l}{\textit{Inference-time guided methods}} \\
\midrule
DecompDiff + CFG & $0.55_{\scriptscriptstyle\pm0.12}$ & $0.68_{\scriptscriptstyle\pm0.13}$ & $0.99_{\scriptscriptstyle\pm1.29}$ & $4.81_{\scriptscriptstyle\pm0.39}$ & $235.35_{\scriptscriptstyle\pm112.65}$ & $0.76_{\scriptscriptstyle\pm0.09}$ \\
DecompDiff + FKC & $0.56_{\scriptscriptstyle\pm0.14}$ & $0.74_{\scriptscriptstyle\pm0.13}$ & $0.94_{\scriptscriptstyle\pm1.50}$ & $4.85_{\scriptscriptstyle\pm0.33}$ & $239.06_{\scriptscriptstyle\pm114.62}$ & $0.63_{\scriptscriptstyle\pm0.17}$ \\
DiffSBDD + CFG & $0.56_{\scriptscriptstyle\pm0.07}$ & $0.73_{\scriptscriptstyle\pm0.06}$ & $2.68_{\scriptscriptstyle\pm1.48}$ & $4.70_{\scriptscriptstyle\pm0.25}$ & $309.31_{\scriptscriptstyle\pm68.60}$ & $0.74_{\scriptscriptstyle\pm0.07}$ \\
DiffSBDD + FKC & $0.56_{\scriptscriptstyle\pm0.08}$ & $0.75_{\scriptscriptstyle\pm0.09}$ & $2.61_{\scriptscriptstyle\pm1.62}$ & $4.72_{\scriptscriptstyle\pm0.30}$ & $292.61_{\scriptscriptstyle\pm78.74}$ & $0.75_{\scriptscriptstyle\pm0.06}$ \\
TargetDiff + CFG & $0.37_{\scriptscriptstyle\pm0.10}$ & $0.53_{\scriptscriptstyle\pm0.04}$ & $2.48_{\scriptscriptstyle\pm1.74}$ & $4.24_{\scriptscriptstyle\pm0.53}$ & $438.35_{\scriptscriptstyle\pm24.49}$ & $0.65_{\scriptscriptstyle\pm0.05}$ \\
TargetDiff + FKC & $0.28_{\scriptscriptstyle\pm0.14}$ & $0.55_{\scriptscriptstyle\pm0.04}$ & $2.35_{\scriptscriptstyle\pm1.62}$ & $3.84_{\scriptscriptstyle\pm0.87}$ & $450.70_{\scriptscriptstyle\pm40.10}$ & $0.69_{\scriptscriptstyle\pm0.07}$ \\
\end{longtable}}

\clearpage
\section{Visualizing pair similarity}
\subsection{Probis Z-Score similarity for each \name task}
\label{app:probis-sim}

\begin{figure}[h]
    \centering
    \includegraphics[width=0.7\linewidth]{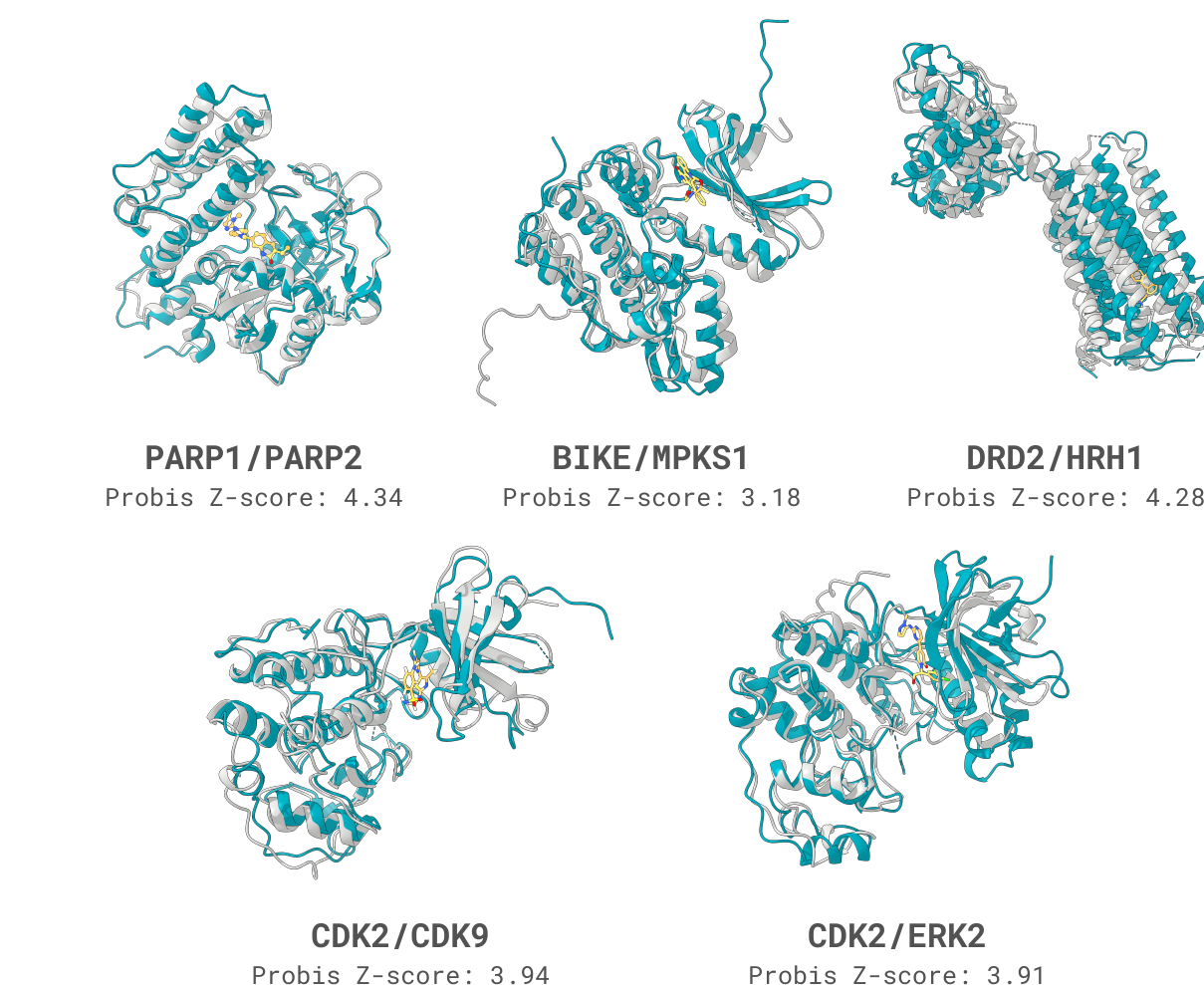}
    \mycaption{\taskA on-/off-target pairs protein structures overlay and their associated similarity.}{This visually demonstrates the high structural similarity of pairs in \taskA. Co-crystallized ligands are displayed in yellow sticks. Reported ProBiS Z-scores quantify global structural similarity of the structures, with higher values indicating stronger similarity.}
    \label{fig:probis_sb_case}
\end{figure}

\begin{figure}[h]
    \centering
    \includegraphics[width=\linewidth]{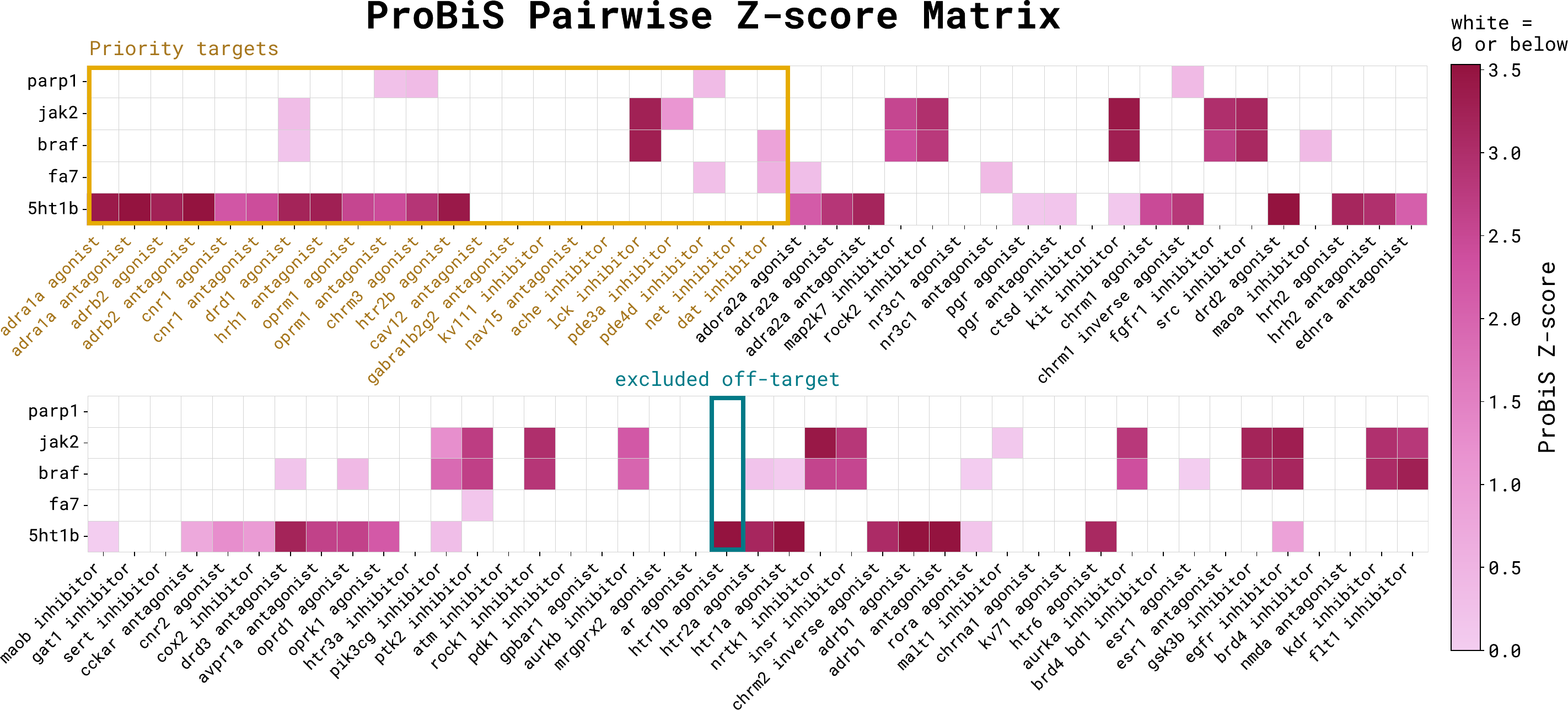}
    \mycaption{Heatmaps of structural similarity between POIs and the off-targets panel in \taskB.}{Each row corresponds to an on-target, each column corresponds to an off-target, and color intensity reflects the ProBiS Z-score, with darker shades indicating higher local structural similarity. ProBiS Z-score of zero or below are set to white. Priority targets of \taskBsmall are highlighted by the yellow box, the HTR1B off-target structure excluded due to overlap with POI 5HT1B is highlighted by the blue box.}
    \label{fig:probis_sb_safety}
\end{figure}

\begin{figure}[h]
    \centering
    \includegraphics[width=\linewidth]{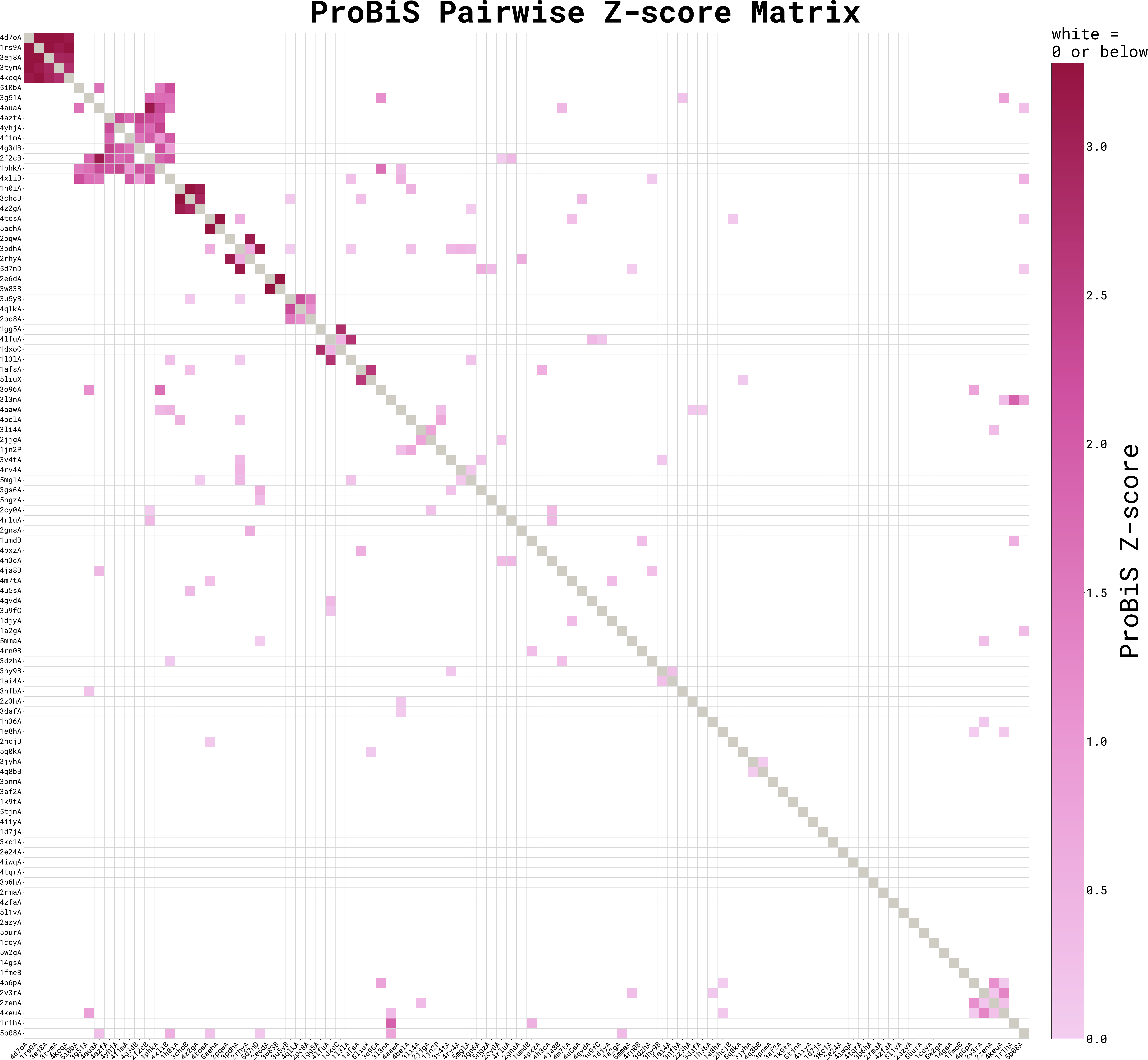}
    \mycaption{Heatmaps of structural similarity between on-/off-target pairs in \taskC.}{Each row corresponds to an on-target, each column corresponds to an off-target, and color intensity reflects the ProBiS Z-score, with darker shades indicating higher local structural similarity. ProBiS Z-score of zero or below are set to white. Grey shade indicates self-pairs, which are excluded from \taskC.}
    \label{fig:probis_sb_xl}
\end{figure}




\clearpage
\pagestyle{plain}
\section*{NeurIPS Paper Checklist}

\begin{enumerate}

\item {\bf Claims}
    \item[] Question: Do the main claims made in the abstract and introduction accurately reflect the paper's contributions and scope?
    \item[] Answer: \answerYes{}
    \item[] Justification: They are supported by our evaluation task, evaluation pipeline, and results section. 

\item {\bf Limitations}
    \item[] Question: Does the paper discuss the limitations of the work performed by the authors?
    \item[] Answer: \answerYes{} 
    \item[] Justification: Limitations of our work are discussed in the limitations section.

\item {\bf Theory assumptions and proofs}
    \item[] Question: For each theoretical result, does the paper provide the full set of assumptions and a complete (and correct) proof?
    \item[] Answer: \answerNA{} 
    \item[] Justification: The paper does not include theoretical results.

    \item {\bf Experimental result reproducibility}
    \item[] Question: Does the paper fully disclose all the information needed to reproduce the main experimental results of the paper to the extent that it affects the main claims and/or conclusions of the paper (regardless of whether the code and data are provided or not)?
    \item[] Answer: \answerYes{} 
    \item[] Justification: Scripts to process data and run models are provided on GitHub and detailed in the paper and appendix.

\item {\bf Open access to data and code}
    \item[] Question: Does the paper provide open access to the data and code, with sufficient instructions to faithfully reproduce the main experimental results, as described in supplemental material?
    \item[] Answer: \answerYes{} 
    \item[] Justification: All scripts alongside detailed \texttt{README.md} to reproduce the main experimental results are provided on the project’s GitHub. A \texttt{CLI} is also provided.

\item {\bf Experimental setting/details}
    \item[] Question: Does the paper specify all the training and test details (e.g., data splits, hyperparameters, how they were chosen, type of optimizer) necessary to understand the results?
    \item[] Answer: \answerYes{}
    \item[] Justification: There is no additional training done since our work is primarily evaluation and benchmarking with our proposed new dataset and metrics. As mentioned in our appendix, we follow each baseline's default inference setting and also detail how to run each baseline in \texttt{README.md} in our GitHub. We include the choice of hyperparameters for inference-time guidance in the paper.

\item {\bf Experiment statistical significance}
    \item[] Question: Does the paper report error bars suitably and correctly defined or other appropriate information about the statistical significance of the experiments?
    \item[] Answer: \answerYes{}
    \item[] Justification: Our results table in appendix reports mean and standard deviation for relevant evaluation metrics.

\item {\bf Experiments compute resources}
    \item[] Question: For each experiment, does the paper provide sufficient information on the computer resources (type of compute workers, memory, time of execution) needed to reproduce the experiments?
    \item[] Answer: \answerYes{}
    \item[] Justification: They are included in compute details in appendix.
    
\item {\bf Code of ethics}
    \item[] Question: Does the research conducted in the paper conform, in every respect, with the NeurIPS Code of Ethics \url{https://neurips.cc/public/EthicsGuidelines}?
    \item[] Answer: \answerYes{} 
    \item[] Justification: The authors have reviewed the NeurIPS Code of Ethics and consider this work is conform to it.

\item {\bf Broader impacts}
    \item[] Question: Does the paper discuss both potential positive societal impacts and negative societal impacts of the work performed?
    \item[] Answer: \answerNA{}
    \item[] Justification: We do not discuss them explicitly in our work.
    
\item {\bf Safeguards}
    \item[] Question: Does the paper describe safeguards that have been put in place for responsible release of data or models that have a high risk for misuse (e.g., pre-trained language models, image generators, or scraped datasets)?
    \item[] Answer: \answerNA{} 
    \item[] Justification: The data and models explored are not high risk for misuse.

\item {\bf Licenses for existing assets}
    \item[] Question: Are the creators or original owners of assets (e.g., code, data, models), used in the paper, properly credited and are the license and terms of use explicitly mentioned and properly respected?
    \item[] Answer: \answerYes{} 
    \item[] Justification: The code to process data is open source. For each dataset, the original paper is cited in the main publication and the license name is provided in the supplementary material.

\item {\bf New assets}
    \item[] Question: Are new assets introduced in the paper well documented and is the documentation provided alongside the assets?
    \item[] Answer: \answerYes{} 
    \item[] Justification: For new datasets, code for processing is provided on GitHub and the curation is described in the main text and Appendix. For the new benchmark, all results are described in the main text and Appendix and code is provided on GitHub to reproduce the results.

\item {\bf Crowdsourcing and research with human subjects}
    \item[] Question: For crowdsourcing experiments and research with human subjects, does the paper include the full text of instructions given to participants and screenshots, if applicable, as well as details about compensation (if any)? 
    \item[] Answer: \answerNA{} 
    \item[] Justification: The paper does not involve crowdsourcing or research with human subjects.

\item {\bf Institutional review board (IRB) approvals or equivalent for research with human subjects}
    \item[] Question: Does the paper describe potential risks incurred by study participants, whether such risks were disclosed to the subjects, and whether Institutional Review Board (IRB) approvals (or an equivalent approval/review based on the requirements of your country or institution) were obtained?
    \item[] Answer: \answerNA{} 
    \item[] Justification: The paper does not involve crowdsourcing or research with human subjects.

\item {\bf Declaration of LLM usage}
    \item[] Question: Does the paper describe the usage of LLMs if it is an important, original, or non-standard component of the core methods in this research? Note that if the LLM is used only for writing, editing, or formatting purposes and does \emph{not} impact the core methodology, scientific rigor, or originality of the research, declaration is not required.
    \item[] Answer: \answerYes{}
    \item[] Justification: We used LLM to edit our codes. But LLMs are not evaluated in this benchmark and do not constitute any important, original, or non-standard components of this work.

\end{enumerate}

\end{document}